\documentclass{article}
\PassOptionsToPackage{numbers, compress, sort}{natbib}
\usepackage[preprint]{neurips_2024}
\usepackage[utf8]{inputenc} 
\usepackage[T1]{fontenc}    
\usepackage{url}            
\usepackage{booktabs}       
\usepackage{amsfonts}       
\usepackage{nicefrac}       
\usepackage{microtype}      
\usepackage{xcolor}         
\usepackage{amsmath}
\usepackage{amssymb}
\usepackage{mathtools}
\usepackage{adjustbox}
\usepackage{amsthm}
\usepackage{subcaption}
\usepackage{dsfont}
\usepackage{xspace}
\usepackage{enumitem}
\usepackage{algorithm}
\usepackage{algpseudocode}
\usepackage{graphicx} 
\usepackage{multirow} 
\usepackage{color,colortbl}
\definecolor{tablegray}{gray}{0.9} 
\usepackage{nicefrac} 
\usepackage{pifont}
\definecolor{ao}{rgb}{0.01, 0.75, 0.24}

\usepackage[hidelinks,colorlinks=true,citecolor=blue!50!black,linkcolor=red!50!black,urlcolor=green!50!black]{hyperref}
\usepackage{cleveref}

\usepackage{acronym}

\newcommand{\tool}{\texttt{ALPBench}\xspace}
\newcommand{\cmark}{\ding{51}}

\newcommand{\xmark}{\ding{55}}%

\newcommand{\cD}{\mathcal{D}}

\newcommand{\cX}{\mathcal{X}}
\newcommand{\cY}{\mathcal{Y}}

\renewcommand{\P}{\mathbb{P}}
\newcommand{\R}{\mathbb{R}}

\newcommand{\DU}{\cD_{U}}
\newcommand{\DL}{\cD_{L}}

\newcommand{\Dtrain}{\cD_{\text{train}}}
\newcommand{\Dtest}{\cD_{\text{test}}}
\renewcommand\vec{\mathbf}

\title{ALPBench: A Benchmark for Active Learning Pipelines on Tabular Data}

\author{%
  Valentin Margraf\\
  MCML, LMU Munich\\
  \texttt{valentin.margraf@ifi.lmu.de}
  \And
  Marcel Wever\\
  MCML, LMU Munich\\
  \texttt{marcel.wever@ifi.lmu.de}
  \And
  Sandra Gilhuber\\
  MCML, LMU Munich\\
  \texttt{gilhuber@dbs.ifi.lmu.de}
  \And
  Gabriel Marques Tavares\\
  MCML, LMU Munich\\
  \texttt{tavares@dbs.ifi.lmu.de}
  \And
  Thomas Seidl\\
  MCML, LMU Munich\\
  \texttt{seidl@dbs.ifi.lmu.de}
  \And
  Eyke H{\"u}llermeier\\
  MCML, LMU Munich\\
  \texttt{eyke@ifi.lmu.de}
}

\newcommand{\numdatasets}{86\xspace}
\newcommand{\numquerystrategies}{9\xspace}
\newcommand{\numlearningalgorithms}{8\xspace}
\newcommand{\numalps}{72}
\newcommand{\numsettings}{2\xspace}

\begin{document}

\maketitle

\begin{abstract}
In settings where only a budgeted amount of labeled data can be afforded, active learning seeks to devise query strategies for selecting the most informative data points to be labeled, aiming to enhance learning algorithms' efficiency and performance. Numerous such query strategies have been proposed and compared in the active learning literature.
However, the community still lacks standardized benchmarks for comparing the performance of different query strategies.
This particularly holds for the combination of query strategies with different learning algorithms into active learning pipelines and examining the impact of the learning algorithm choice. To close this gap, we propose \tool, which facilitates the specification, execution, and performance monitoring of active learning pipelines. It has built-in measures to ensure evaluations are done reproducibly, saving exact dataset splits and hyperparameter settings of used algorithms. In total, \tool consists of \numdatasets real-world tabular classification datasets and 5 active learning settings, yielding 430 active learning problems. To demonstrate its usefulness and broad compatibility with various learning algorithms and query strategies, we conduct an exemplary study evaluating \numquerystrategies query strategies paired with \numlearningalgorithms learning algorithms in \numsettings different settings. We provide \tool here: \url{https://github.com/ValentinMargraf/ActiveLearningPipelines}.

\end{abstract}

\section{Introduction}

Tabular data is one of the most commonly used data types across various sectors, including medicine~\cite{przystalski2023medical_tabular}, finance, insurance~\cite{finance_insurance_HussainP16} or manufacturing~\cite{app13031903_manufacturing}. Therefore, it is also one of the most frequently utilized data types for real-world \ac{ml} applications~\cite{Chui2018_ai_frontier}. However, generally speaking, data must first be labeled before it can be used in supervised \ac{ml} tasks. Labeling data is a cumbersome and costly process, as it is time-consuming and typically accomplished by human domain experts.

For situations where only a limited budget is available for labeling datasets, the field of \ac{al} \cite{settles2009active,al_survey_recent} has emerged to develop methods for actively selecting the most suitable data points from unlabeled data to be labeled by a so-called \textit{oracle}.
The term ''most suitable`` here refers to data points that help achieve the best possible generalization performance for a given learning algorithm.
A diverse array of \acp{qs} have been developed that somehow quantify the suitability of a data point being labeled.
Which \ac{qs} works best heavily depends on various factors, including the dataset, the budget constraints, the learning algorithm, among other things~\cite{comparison_different_al,combi_learner_al,when_does_al_work}. 

However, while several empirical evaluations have already been conducted \cite{rebenchmarking_al,margin_all_you_need,ijcai_poolbased_al_benchmark,benchmark_log_reg,rebenchmarking_al}, the community still lacks a standardized benchmark for comparing the performance of different query strategies, providing an open-source implementation that facilitates their comparison. Moreover, the existing evaluations are often limited in the number of datasets, considering only binary classification datasets or already outdated query strategies \cite{benchmark_log_reg,rebenchmarking_al}. Further studies only consider one particular learning algorithm \cite{rebenchmarking_al,margin_all_you_need,ijcai_poolbased_al_benchmark}, which might lead to biased results, as the algorithm also influences the performance of a query strategy. Lastly, these learning algorithms often do not properly represent state-of-the-art methods. While \ac{gbdt} ensembles, such as XGBoost \cite{xgboost} or Catboost \cite{catboost}, as well as deep learning architectures \cite{tabnet,tabpfn} have proven particularly successful for tabular data, they are not included in these studies. 

\begin{figure}[t]
    \centering
    \includegraphics[width=\textwidth]{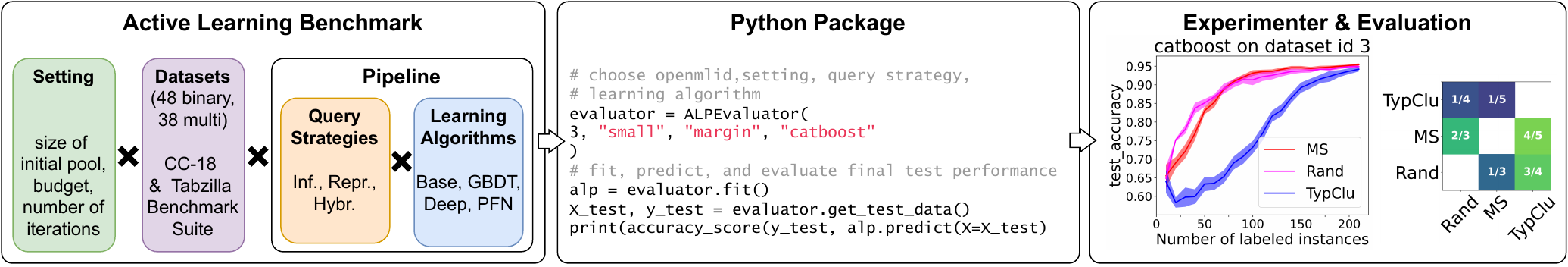}
    \caption{The contributions of our paper are threefold: (i) the first active learning benchmark considering pipelines of query strategies and learning algorithms, (ii) an extensible Python package for applying and benchmarking active learning pipelines, and (iii) an extensive empirical evaluation of active learning pipelines.}
    \label{fig:overview}
\end{figure}

\paragraph{Contributions.} Thus far, a comprehensive benchmark to investigate the benefits of different query strategies for active learning, especially in combination with the choice of learning algorithms, is still lacking.
In this work, we address this gap by proposing \tool, a comprehensive benchmark for active learning pipelines, comprising query strategies and learning algorithms, in the domain of tabular data classification tasks.
The benchmark includes a total of \numdatasets different multi-class and binary-class tabular datasets and 5 different active learning problem settings, specifying the amount of initially labeled data points, the budget per iteration, and the number of iterations. Overall, as visualized in Figure~\ref{fig:overview}, our contributions are threefold, and can be summarized as follows:
\begin{enumerate}[leftmargin=*]
    \item[\textbf{1.}] 
    We propose \tool, the first active learning benchmark for tabular data that allows for combining different learning algorithms and query strategies into active learning pipelines to execute them, monitor their performance, and benchmark against other pipelines.
    
    \item[\textbf{2.}] 
    We provide an open-source implementation of \tool as an extensible Python package for applying and benchmarking active learning pipelines. In an experimental study we showcase its usefulness by evaluating 72 different active learning pipelines on 86 real-world classification datasets across 2 settings. 
    
    \item[\textbf{3.}] 
    We find that strong active learning pipelines are often combining state-of-the-art classification methods, such as TabPFN \cite{tabpfn}, Catboost \cite{catboost} or Random Forest \cite{leobreiman_rf}, as learning algorithms and information-based query strategies. While the learning algorithm turns out to be the most crucial choice, the suitability of query strategies also varies for different datasets. This underlines the need of also considering different learning algorithms in future empirical studies in active learning.
\end{enumerate}

\section{Related Work}

There exists some literature evaluating active learning pipelines on tabular data classification tasks.
An early benchmark of \ac{al} demonstrated that \ac{ms} often outperforms other \acp{qs}~\cite{al_for_logreg} in combination with \ac{lr} as a learning algorithm. The performance of combining varying learning algorithms and \acp{qs} was investigated in \cite{when_does_al_work, combi_learner_al, comparison_different_al}. However, the studies are outdated since the chosen algorithms are pretty simplistic, i.e., there are stronger \ac{ml} algorithms nowadays~\cite{why_tree_still_outperform_nn,when_nn_outperform_trees_on_tab}, and many of the used datasets from the UCI repository~\cite{uci_datasets} are rather old.

More recent \acp{qs} were investigated in~\cite{benchmark_log_reg,ijcai_poolbased_al_benchmark,rebenchmarking_al}. Although varying the strategy for instance selection, the learning algorithm is fixed, precisely a  \ac{svm}~\cite{ijcai_poolbased_al_benchmark,rebenchmarking_al} or \ac{lr}~\cite{benchmark_log_reg}. However, as the employed learner is crucial to the overall performance of \ac{al}~\cite{comparison_different_al}, such design choice raises the question of whether the findings generalize to other learners as well. Further, their scope is limited to binary or only a handful of multi-class datasets. 

All mentioned papers so far only considered one specific \ac{al} setting, i.e., the size of the initially labeled pool and the budget. \cite{benchmark_log_reg} initially provided only one labeled instance for each class, compared to, e.g.,~\cite{rebenchmarking_al}, which randomly sampled 20 instances for the labeled pool. These misalignments across different benchmarks complicate comparisons and hinder the ability to draw general conclusions. \cite{margin_all_you_need} are the first ones addressing this issue by investigating three different \ac{al} settings. They also considered very recent \acp{qs} and datasets from the OpenML-CC18 Benchmark Suite~\cite{cc18}. However, again, the authors chose one specific learner, in this case, a deep neural network.

Motivated by recent papers such as~\cite{when_nn_outperform_trees_on_tab, why_tree_still_outperform_nn}, we believe that an up-to-date benchmark has to include multiple \ac{sota} learning algorithms for tabular data such as \acp{gbdt} (e.g., Catboost) and \acp{pfn} (e.g., TabPFN~\cite{tabpfn}) as well as recent \acp{qs}, e.g., power margin sampling and power BALD~\cite{power_versions}. To the best of our knowledge, we are the first to combine various \ac{sota} learning algorithms with \acp{qs} and evaluate their performance on a large amount of binary and multi-class real-world classification tasks for tabular data. Furthermore, we provide our implementation as a Python package with a unified API.

\section{Pool-based Active Learning}
In pool-based \ac{al} instances from the pool of unlabeled data are selected to be labeled by an oracle, which is done in an iterative procedure. Three different scenarios are commonly considered in \ac{al}, namely, the membership-query synthesis, stream-based, and pool-based scenario~\cite{settles2009active}. We focus on the pool-based scenario, being the preferred one in real-world applications~\cite{al_survey_recent}. We first describe this scenario in Section~\ref{sec:problem} before elaborating on the implemented learning algorithms and \acp{qs} within \tool (Sections~\ref{sec:learning_algorithm} and~\ref{sec:query_strategies}), and their combination into \acp{alp} (Section~\ref{sec:alps}).

\subsection{Problem Definition}\label{sec:problem}
In a classification setting, we are given a $d$-dimensional feature space $\cX \in \R^d$ and a label set $\cY = \{1, ..., C\}$. A \ac{ds} is denoted as $\cD = \{(\vec{x}_i,y_i)\}_{i=1}^n \subset \mathcal{X} \times \mathcal{Y}$, where each instance $\vec{x}_i = (x_i^1, \ldots , x_i^d) \in \cX$ is associated with an underlying true label $y_i \in \cY$.
In \ac{al}, however, only a small \ac{ds} $\DL^0 = \{(\vec{x}_i,y_i)\}_{i=1}^l$ is initially labeled, whereas a considerably larger pool of instances $\DU = \{(\vec{x}_i)\}_{i=l+1}^n$ is unlabeled. From this unlabeled pool, instances to be labeled by the oracle~$\mathcal{O}$ are selected by a \ac{qs}. More specifically, the goal is to strategically select instances such that the predictive (probabilistic) model $h:\cX \rightarrow \P(\cY)$ induced by the learning algorithm on the labeled data minimizes the generalization error (risk) with respect to a given loss function $\ell: \cY \times \P(\cY) \rightarrow \R^+$. Here, $\P(\cY)$ denotes the space of probability distributions over $\cY$. A given budget of $B$ can be spent for labeling, meaning that $B$ instances from $\DU$ can be chosen and queried to $\mathcal{O}$. In the pool-based scenario, a predefined amount of $R$ instances is queried per iteration ($R \leq B$) and added to the current labeled \ac{ds} $\DL^i$, on which the learning algorithm is run to induce an updated model $h$.

\subsection{Learning Algorithms}\label{sec:learning_algorithm}
The choice of the learning algorithm is quite important for the overall success of \ac{al}~\cite{caravalho_meta_features}. However, existing benchmarks typically fix a single learning algorithm, such as a \ac{dnn}~\cite{margin_all_you_need} or an \ac{svm}~\cite{rebenchmarking_al}, and recommend suitable \acp{qs} for this choice. To reveal insights for suitable \acp{qs} based on different learning algorithms, we investigate a variety of models. In particular, we choose the following models, which cover a wide range of model types and include \ac{sota} learning algorithms for tabular data~\cite{when_nn_outperform_trees_on_tab}: \ac{svm}, \ac{knn}, \ac{rf}, \ac{etc}, \ac{lr}, and \ac{nb} represent the group of base learners. For each of them, we implement multiple instantiations with different parameter settings. Further, we choose two \acp{gbdt}, namely \ac{xgb} and Catboost. Finally, we include a \ac{mlp} and TabNet~\cite{tabnet} as representatives of \acp{dnn}, and TabPFN~\cite{tabpfn} representing \acp{pfn} in \tool.

\subsection{Query Strategies}\label{sec:query_strategies}

Query strategies (\acsp{qs}) can be classified into information-based (Info.), representation-based (Repr.), and hybrid strategies (Hybr.)~\cite{settles2009active,al_survey_recent}. Info.-based strategies leverage the predictions of the learning algorithm to select instances where the learner exhibits uncertainty, as from these instances we expect the most informative insights. Repr.-based strategies rely solely on the structure of the data to identify the most representative instances. Hybr. strategies combine both of the aforementioned strategies.

Formally, let $z_i \in \mathcal{Z}$ either be a raw input instance, or its embedding of a neural network, $p_i \in \mathcal{P}$ the predicted class probabilities of a learning algorithm for that instance and $\{(\vec{x}_i)\}_{i=1}^\mathcal{R} \subseteq \DU$ the pool of instances that is queried by the \ac{qs} in each iteration. Loosely speaking, info.-based approaches select instances based on some uncertainty measure $u(\cdot)$ on the probability scores, repr.-based compute representativeness $r(\cdot)$ leveraging the structure of $\mathcal{Z}$; hybr. approaches combine both:

\[
\begin{array}{c@{\hspace{1cm}}c@{\hspace{1cm}}c}
\{(\vec{x}_i)\}_{i=l+1}^\mathcal{R} \sim u(p_i)  & \{(\vec{x}_i)\}_{i=l+1}^\mathcal{R} \sim r(z_i) & \{(\vec{x}_i)\}_{i=l+1}^\mathcal{R} \sim u(p_i) + r(z_i) \\
&&\\
\text{Information-based} & \text{Representation-based} & \text{Hybrid}
\end{array}
\]

\textit{Information-based.} Information or uncertainty-based approaches calculate the uncertainty for each instance in the unlabeled pool, leveraging probability scores of the learning algorithm to subsequently select the most uncertain instances. These approaches are quite fast, as the calculations are performed in the (lower-dimensional) space of probabilities. However, they bear the risk of leading to a strong distribution shift in the data. In this work, we implement various approaches, most of which were also considered by~\cite{margin_all_you_need}. 
Amongst them are the well-known \acf{ms}~\cite{margin_sampling}, \ac{es}~\cite{shannon} and \ac{lc}~\cite{least_confident}, sampling instances which have the lowest margin, highest entropy, or where the learning algorithm is the least-confident about, respectively. \ac{vr}~\cite{variance_reduction} and \ac{eer}~\cite{expected_error_reduction} select instances that are expected to reduce the prediction error or output variance, respectively. \ac{eu}~\cite{epistemic_uncertainty_sampling_eyke} samples instances that have the highest epistemic uncertainty. Further, we also considered methods that compute uncertainty based on predicted probabilities of an ensemble such as \ac{qbc}~\cite{qbc} (disagreement of the ensemble members), \ac{maxent}~\cite{max_entropy} (entropy of the averaged predictions) and \acs{bald}~\cite{BALD} (difference between \ac{maxent} and the averaged entropy of the members' predictions). \acs{powms} and \acs{powbald}~\cite{power_versions} build on \ac{ms} and \acs{bald} but add a noise term to the uncertainty scores to enforce diversity within the queried instances.

\textit{Representation-based.} \acp{qs} compute the representativeness of each instance in the raw input space or in some feature space. Both can potentially be high-dimensional, leading to high computational costs. \Ac{kmeans}~\cite{kmeans} performs clustering of the instances in $\DU$ and selects those that are nearest to the cluster centers. \Ac{typclu}~\cite{typical_clustering} clusters all instances in $\DL$ and $\DU$ and then selects instances that lie in clusters in which no instance of $\DL$ is located. \acs{coreset}~\cite{core_set} queries those instances from $\DU$ for which the closest neighbor in $\DL$ is the most distant. 

\textit{Hybrid.} Hybrid approaches combine uncertainty and representativeness. \Ac{clums} ~\cite{cluster_margin} selects instances by first performing clustering on $\DU$ and then taking into account the margin scores as well. \Ac{clue}~\cite{clue} performs weighted k-means clustering on $\DU$ with the entropy of the learning algorithm as sample weight. \acs{falcun}~\cite{FALCUN_gilhuber} computes a relevance score per instance, consisting of the margin scores of the learning algorithm and a diversity score, which is computed from distances between the predicted probabilities.

\subsection{Active Learning Pipelines}\label{sec:alps}
We call the combination of a learning algorithm and a \ac{qs} an \acf{alp}.
Within an \ac{alp} the learning algorithm and \ac{qs} are used in alternating order to (re-)fit a model for the labeled data points and determine data points to be queried to the oracle. In \tool, we explicitly account for this interplay and therefore allow for constructing \acp{alp} out of every possible combination of learning algorithms and \ac{qs} as long as they work with certain interfaces. 

\section{Active Learning Pipeline Benchmark}\label{sec:alpb}

\tool is meant to provide an easy-to-use and easy-to-extend platform for investigating \acp{alp}, considering different combinations of learning algorithms and \acp{qs}, and evaluating new query strategies to be tested and compared against already known strategies. To this end, in \tool, we aim for high modularity with simple interfaces for the individual parts of an \ac{alp}, as well as for applying the composed pipelines to different datasets and experiment setups.

To facilitate the usage of \tool, we subsequently explain how \ac{al} problems are specified (Section~\ref{sec:alpbench-active-learning-problems}), how \acp{alp} are specified (Section~\ref{sec:alpbench-active-learning-pipelines}), and what measures are taken for ensuring reproducibility and therewith high-quality experimental studies (Section~\ref{sec:alpbench-reproducibility}).

\subsection{Specification of Active Learning Problems}\label{sec:alpbench-active-learning-problems}
\textit{Setting.}
A setting describes the basic parameters of an \ac{al} benchmark problem. This includes the size of the test data, the size of the initially labeled dataset, the number of \ac{al} iterations, and how many data points may be queried in one such iteration.

\textit{Scenario.} A scenario combines the fundamental parameters of a setting with a concrete classification task, i.e., an OpenML dataset ID, seeds for splitting the dataset into initially labeled, unlabeled, and test data, and a seed for pseudo-random execution of the active learning pipeline.

In \tool, we predefine 5 different settings of varying sizes of initially labeled data and additional budget for labeling, with and without defining the budget depending on the number of classes.
Therefore, by specifying a scenario, we can describe a single task. However, to conduct broader empirical studies, we need to have suites of benchmarks, which can also be specified in \tool:

\textit{Benchmark Suite} Benchmark Suites in \tool are essentially collections of datasets that can be combined with scenarios. \tool allows for specifying custom benchmark suites, with OpenML \cite{openml} serving as the backbone for datasets. To define new benchmark suites, it suffices to either give a benchmark ID from OpenML or specify a list of OpenML dataset IDs.

In our benchmark implementation, we provide five scenarios and two benchmark suites: OpenML-CC18 \cite{cc18} and TabZilla \cite{when_nn_outperform_trees_on_tab}. Both benchmark suites together comprise a total of \numdatasets datasets.

\subsection{Specification of Active Learning Pipelines}\label{sec:alpbench-active-learning-pipelines}

To apply \ac{al} methods to \ac{al} problems, \acfp{alp} are specified by a learner and a \acf{qs}, as has been outlined in Section~\ref{sec:alps}

\textit{Active Learning Pipeline.} An \ac{alp} is mainly composed of two components, precisely a \ac{qs} and a learner, plus one optional component in the form of an initializer, which can be used to select an initial set of data points. It implements the main logic for the interplay between the learner and \ac{qs} and takes care of the communication with the oracle.

\textit{Learner.} The learner is a learning algorithm that implements the scikit-learn classifier interface and is responsible for model induction. There are no restrictions on the type of learner as long as its interface matches that of a \texttt{scikit-learn} classifier. It is only provided with labeled data points.

\textit{Query Strategy.} The \ac{qs} selects unlabeled data points to be queried by the oracle. To perform its task, it is provided with the learner, the already labeled and unlabeled data points. While some \acp{qs} are wrapped and included from the \texttt{scikit-activeml} library~\cite{skactiveml2021}, the remaining \acp{qs} are original implementations in \tool.

In our benchmark implementation, we include 23 \acp{qs} and a broad spectrum of 23 different learners that can be combined into more than 500 \acp{alp}. 

\subsection{Reproducibility and Experimentation}\label{sec:alpbench-reproducibility}

As we would like to ensure a high-quality standard for experiments conducted with \tool, we provide support for logging and facilitate the execution of experiments.

\textit{Benchmark Connector.} The benchmark connector stores meta-information relevant for reproducibility. This includes storing the indices of data points that are labeled initially and used for testing. Furthermore, the settings of hyperparameters of learners and query strategies are stored so that the same configurations can be maintained for future studies. We provide two facades of the Benchmark Connector, one using a database as data storage and one that works locally with a filesystem.
    
\textit{Experimenter.} Building on pyExperimenter~\cite{pyExperimenter}, \tool comes with some convenience functionalities to foster large-scale experimental studies. A cross-product experiment grid is specified for some default setup and can be easily extended by more alternatives. Furthermore, we provide logging facilities to observe the active learning process, recording labeling statistics and learner performances.

\section{Experiments}\label{sec:exp}
To demonstrate the usefulness of \tool, we conduct an empirical study comparing various active learning pipelines composed of different combinations of \acp{qs} and learning algorithms. Concretely, we investigate the effectiveness of \numquerystrategies \acp{qs} and pair them with \numlearningalgorithms learning algorithms, thereby constituting the most extensive study on active learning pipelines. The experimental setup is explained in Section~\ref{sec:experimental-setup} before the evaluation methods and results are described in Sections~\ref{sec:eval_methods} and~\ref{sec:results}, respectively.
More details regarding the experiments are given in Appendix~\ref{sec:apx-experiments}. 

\subsection{Experimental Setup}\label{sec:experimental-setup}
In our experimental study, we consider a total of \numdatasets real-world datasets from the OpenML-CC18~\cite{cc18} and the TabZilla~\cite{when_nn_outperform_trees_on_tab} benchmark suites. Furthermore, of the 23 \acp{qs} \tool provides, we select a set of \numquerystrategies representatives covering the different types of query strategies, as well as a subset of \numlearningalgorithms learning algorithms from different ends of the bias-variance spectrum. Our selection of learning algorithms ranges from linear models to highly non-linear models, including various decision tree ensembles and SOTA deep learning methods for tabular data. More precisely, we include \ac{es}~\cite{shannon}, \ac{ms}~\cite{margin_sampling}, \acs{powms} and \acs{powbald}~\cite{power_versions}, \acs{coreset}~\cite{core_set}, \acs{falcun}~\cite{FALCUN_gilhuber}, \ac{clums}~\cite{cluster_margin} and \ac{typclu}~\cite{typical_clustering}, and \ac{rand} as \acp{qs} and \ac{svm}, \ac{knn}, \ac{mlp}, \ac{rf}, \ac{xgb}, Catboost, TabNet, and TabPFN as learning algorithms.

In Table~\ref{tab:comparison_benchmarks_small}, we compare the scope of our experimental evaluation to previous studies on active learning for tabular data~\cite{benchmark_log_reg,ijcai_poolbased_al_benchmark,margin_all_you_need,rebenchmarking_al}. 
We would like to note that despite considering only a subset of methods provided in \tool, our work provides the most comprehensive empirical comparison so far.
To further contain computational costs, we limit the training time of learning algorithms to 180 seconds per iteration.
This of course may decrease the performance of deep learning algorithms, such as TabNet, and poses a limitation to the generalizability of our empirical study.

\textit{Active Learning Setting.} Acknowledging the effect of different sizes of the initially labeled pool $\DL$ and the budget $\mathcal{B}$, we consider\,--\,similarly to~\cite{margin_all_you_need}\,--\,different \ac{al} settings. However, we firstly choose a \textit{small} and a \textit{large} setting, excluding a \textit{medium} setting, for which~\cite{margin_all_you_need} found quite similar results as for the \textit{small} one. Secondly, we do not choose a \textit{task-independent} size of $|\DL^0|$ and the per-iteration budget $R$, but rather a \textit{task-dependent} size dependent on the difficulty of the task.

\begin{table}[t]
\caption{Comparison of the scopes of our \textbf{evaluation study} and previous ones.}
\label{tab:comparison_benchmarks_small}
\centering
\resizebox{ .9\textwidth}{!}{

\begin{tabular}
{crp{2.2cm}p{2.2cm}p{2.2cm}p{2.2cm}p{2.4cm}}
\toprule
                             &             &\citet{benchmark_log_reg} &
                             \citet{ijcai_poolbased_al_benchmark}& 
                             \citet{rebenchmarking_al}&
                             \citet{margin_all_you_need} &
                              \textbf{Ours} \\
\midrule
                    
\multirow{4}{*}{\rotatebox{90}{\ac{qs}}} & Info.         &  \cite{shannon,expected_error_reduction,max_error_reduction,combined_error_reduction,al_for_logreg,fisher_for_al,minimum_loss_increase,max_model_change}  &  \cite{shannon,least_confident,margin_sampling,qbc,expected_model_change,expected_error_reduction,variance_reduction}  &  \cite{shannon,least_confident,margin_sampling,qbc,variance_reduction,expected_error_reduction} &  \cite{shannon,least_confident,margin_sampling,min_margin,max_entropy,BALD,power_versions,qbc_VR} &  \cite{shannon,margin_sampling,power_versions} \\ 
                             & Repr.         &  -  &  \cite{kmeans}  & \cite{hier,core_set}  & \cite{core_set} &  \cite{core_set, typical_clustering}   \\ 
                             & Hybr.      &  \cite{adaptive_active_learning}  &  \cite{density_weighted,QUIRE,GRAPH,al_by_learning}  &  \cite{density_weighted,GRAPH,QUIRE,al_by_learning} &  \cite{margin_density,cluster_margin}  &  \cite{cluster_margin,typical_clustering,FALCUN_gilhuber}  \\  
\midrule
\multirow{4}{*}{\rotatebox{90}{Learner}}    & Base        &  \ac{lr}  &  \ac{svm}  & \ac{svm}  &  -  &   \ac{knn}, \ac{svm}, \ac{rf}  \\
                             & \ac{gbdt}        &  -  &  -  &  - &  -  &    CatBoost, \ac{xgb}  \\
                             & \ac{dnn}        &  -  &  -  &  - &  \ac{mlp}  &    \ac{mlp}, TabNet~\cite{tabnet}   \\
                             & \ac{pfn} & -&-&-&-&TabPFN~\cite{tabpfn}\\

\midrule
\ac{alp} & $\sum$ & 10 & 14 & 13 & 13 & \textbf{\numalps} \\
\midrule
\multirow{3}{*}{\rotatebox{90}{\ac{ds}}}    & Binary      &  44  &  35  &  26 & 35   &  48 \\
                             & Multi &  -  &  9  & -  &  34  &   38  \\
                             & $\sum$ & 44 & 44 & 26 & 69 & \textbf{\numdatasets} \\
\midrule
\multicolumn{2}{l}{\ac{al} Setting}                       &   1 &   1 & 1  &  3 (s,m,l)  &  \numsettings (s,l) \\     

\bottomrule

\end{tabular}
}
\end{table}
\textit{Datasets.}
We evaluate each \ac{alp} on all \acp{ds} from the OpenML-CC18~\cite{cc18} and the TabZilla Benchmark Suite~\cite{when_nn_outperform_trees_on_tab}, except for 4 quite large datasets, with OpenML IDs 1567, 1169, 41147, and 1493, leaving us with 48 binary and 38 multi-class datasets. 

\subsection{Evaluation Methods}\label{sec:eval_methods}
We first want to investigate whether strong learning algorithms for tabular data found by~\cite{when_nn_outperform_trees_on_tab} still perform well in the low-label regime, especially when combined with \acp{qs} into \acp{alp}. 

\textit{Heatmaps.} Assuming $N$ learning algorithms and $M$ \acp{qs}, we compute so-called heatmaps $H$ of size $N \times M$. Let learner $i$ and \ac{qs} $j$ form the combined \ac{alp}$_{(i,j)}$, \ac{alp}$_d$ be the winning \ac{alp} for the dataset $d$ and $D$ the number of available datasets. Then, the entry of $H$ at position $(i,j)$ is defined as
\begin{equation*}
H_{(i,j)} = \sum_{d=1}^{D} \mathds{1}[\text{\ac{alp}$_{(i,j)}$ is not statistically significant from \ac{alp}$_d$ on dataset d}]
\end{equation*}
To determine statistical significance, we use Welch's t-test with $p = 0.05$. The indicator function hence evaluates to one for \ac{alp}$_d$ and all \acp{alp} for which the null hypothesis cannot be rejected.

We further aim to investigate the interplay of the \ac{qs} with different learners, to reveal which \acp{qs} are particularly effective for each learner. In agreement with common evaluation procedures for comparing \acp{qs}~\cite{scarf,margin_all_you_need,rebenchmarking_al}, we also computed budget curves and win-matrices.

\begin{figure}[t]
    \centering
    \begin{subfigure}[b]{0.24\textwidth}
        \centering
        \includegraphics[width=\textwidth]{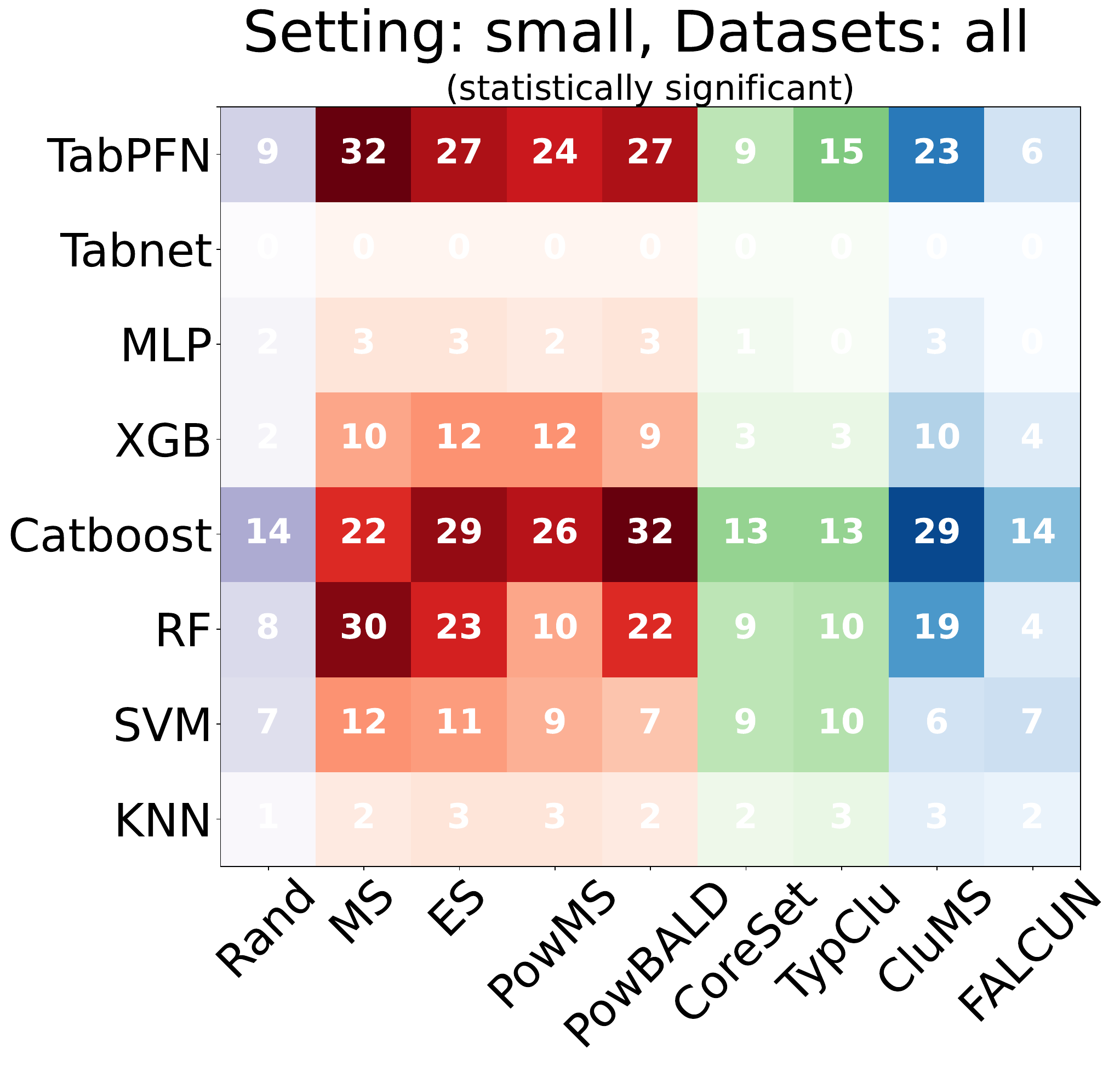}
    \end{subfigure}
    \begin{subfigure}[b]{0.24\textwidth}
        \centering
        \includegraphics[width=\textwidth]{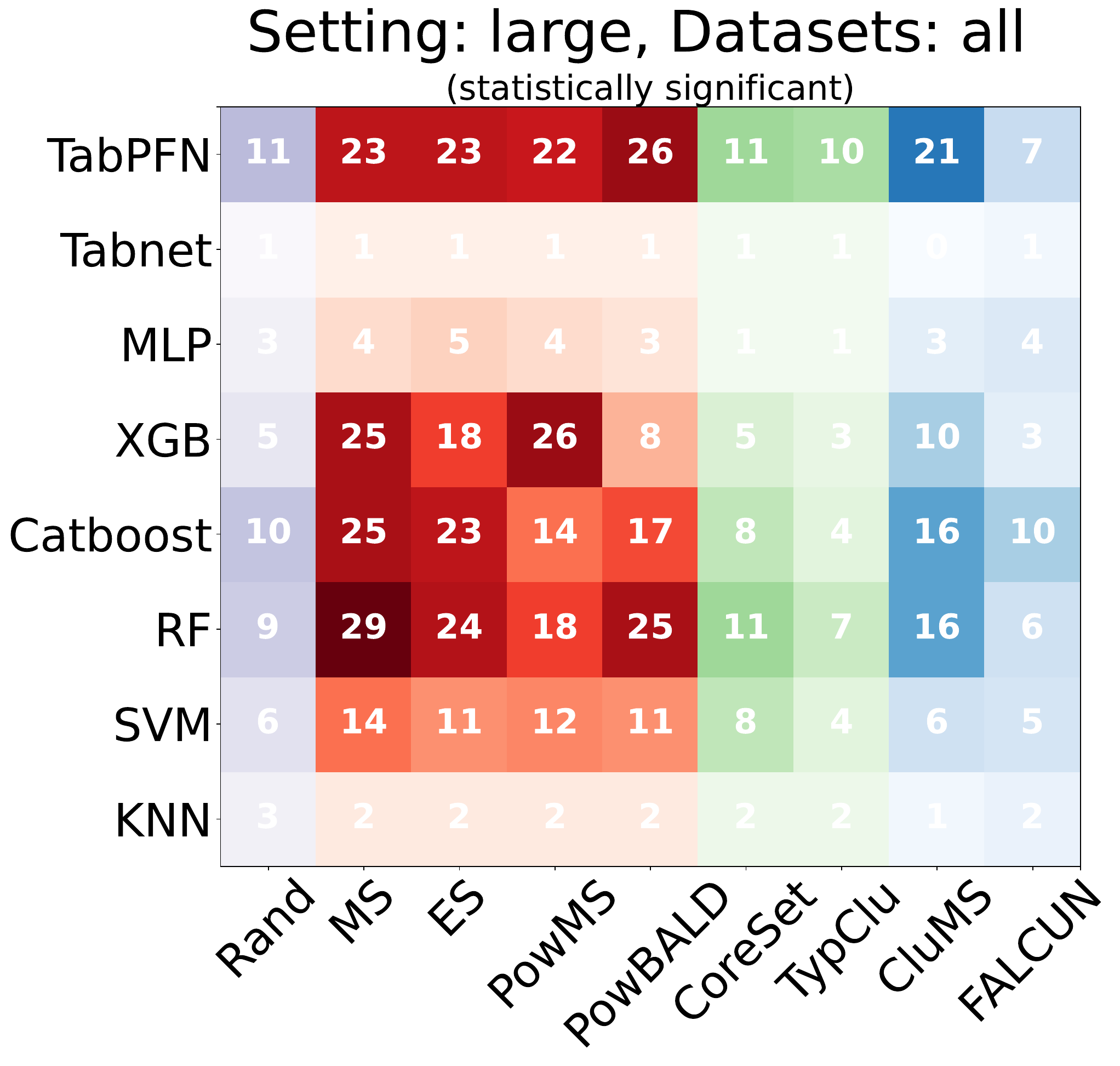}
    \end{subfigure}
    \begin{subfigure}[b]{0.24\textwidth}
        \centering
        \includegraphics[width=\textwidth]{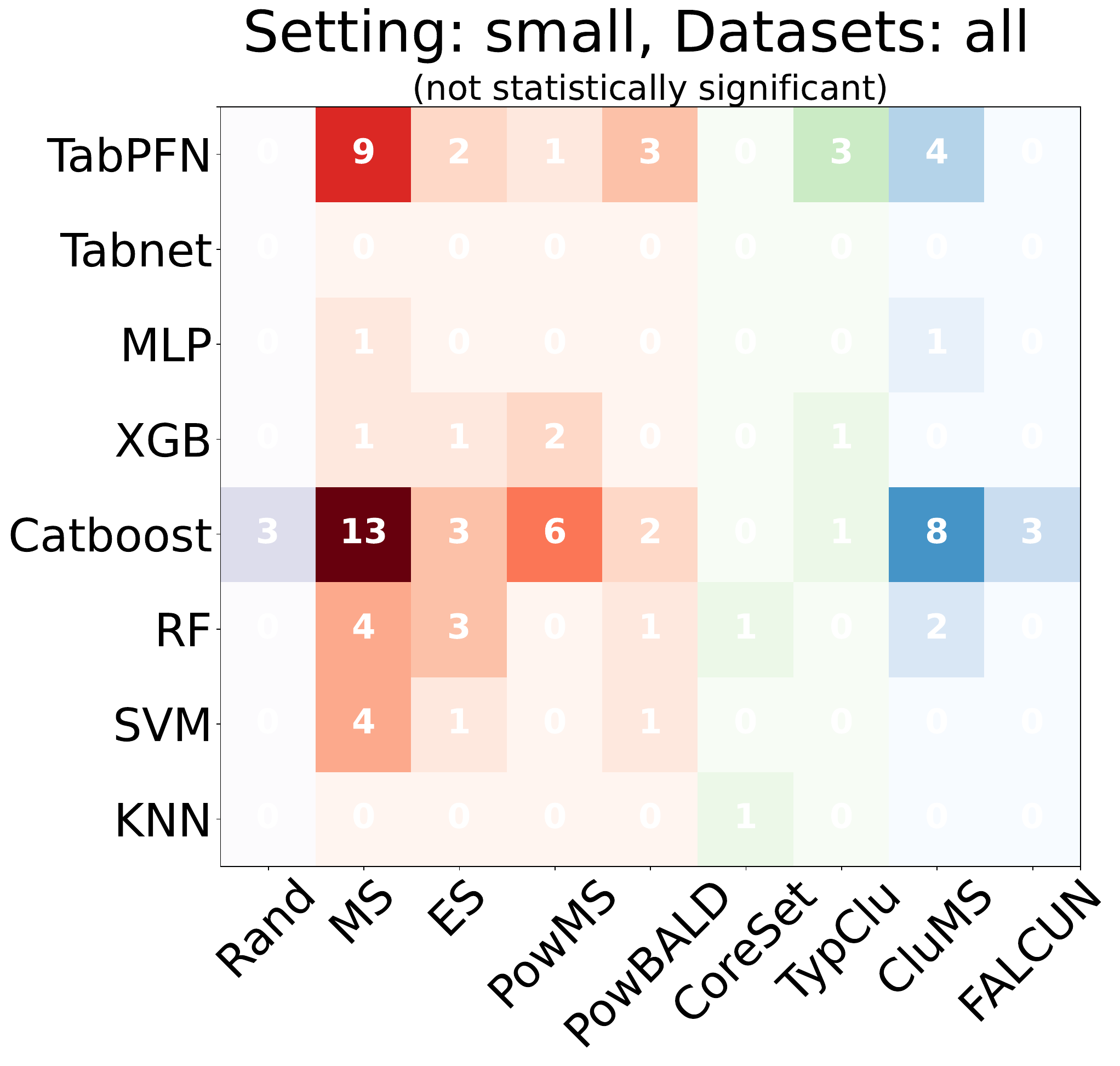}
    \end{subfigure}
    \begin{subfigure}[b]{0.24\textwidth}
        \centering
        \includegraphics[width=\textwidth]{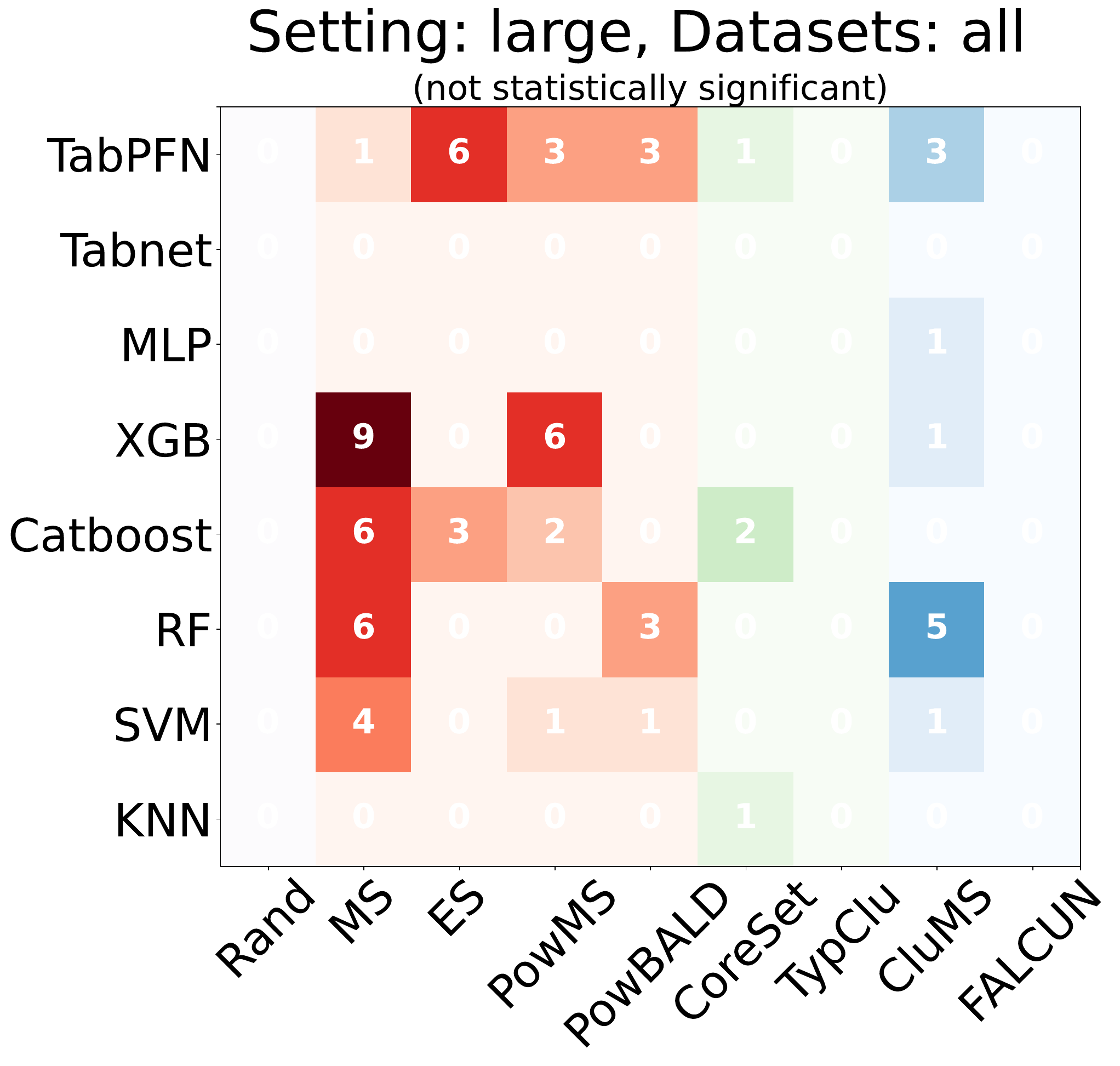}
    \end{subfigure}
    \caption{Heatmaps for all \acp{alp} within our evaluation study \textbf{with} statistical significance (first and second subfigure) and \textbf{without} (third and fourth). Information-based, representation-based, and hybrid \acp{qs} are colored in red, green, and blue, respectively, and random sampling is in purple.}
    \label{fig:heatmaps}
\end{figure}

\begin{figure}[t]
    \centering
    \begin{subfigure}[b]{0.24\textwidth}
        \centering
        \includegraphics[width=\textwidth]{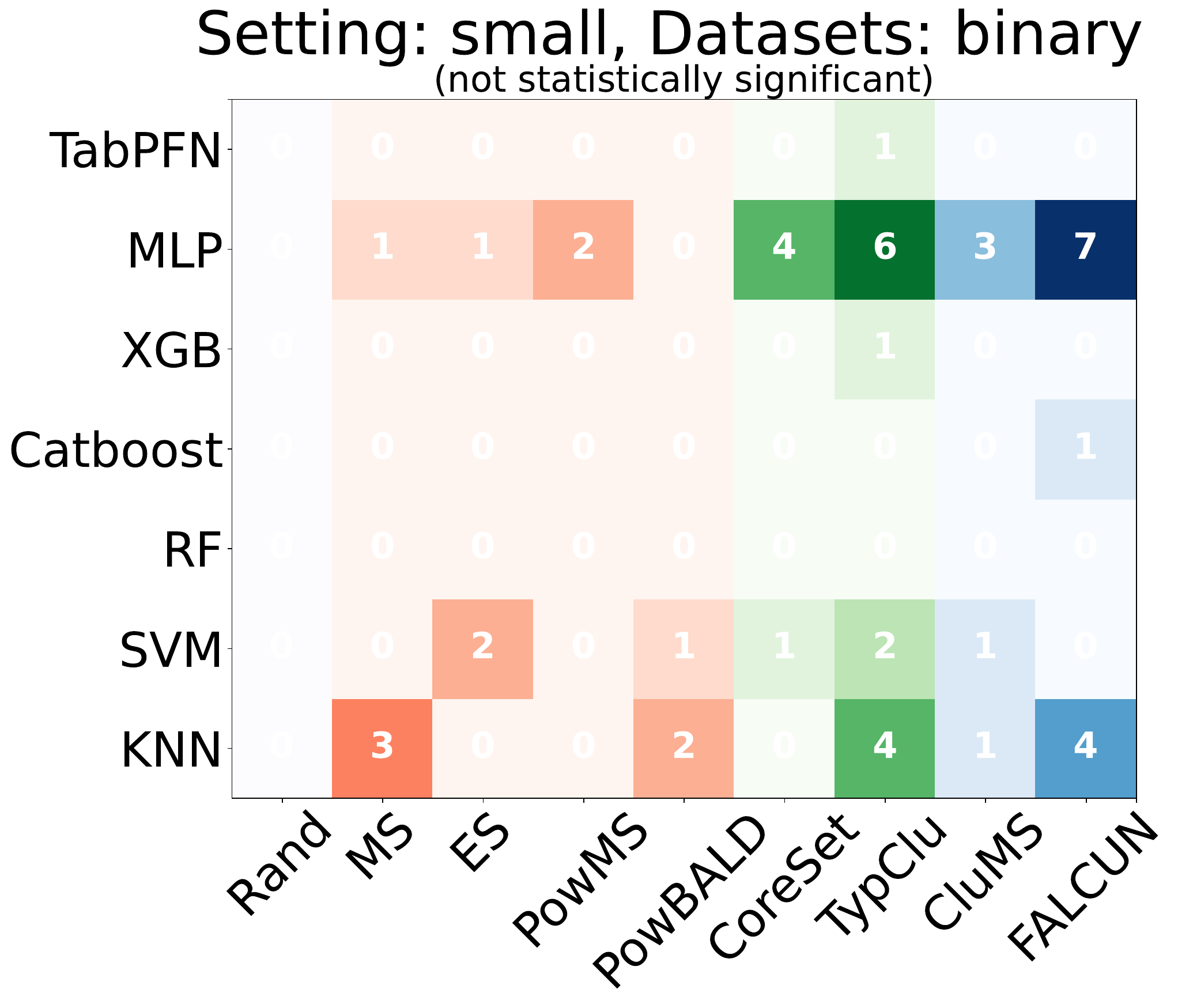}
    \end{subfigure}
    \begin{subfigure}[b]{0.24\textwidth}
        \centering
        \includegraphics[width=\textwidth]{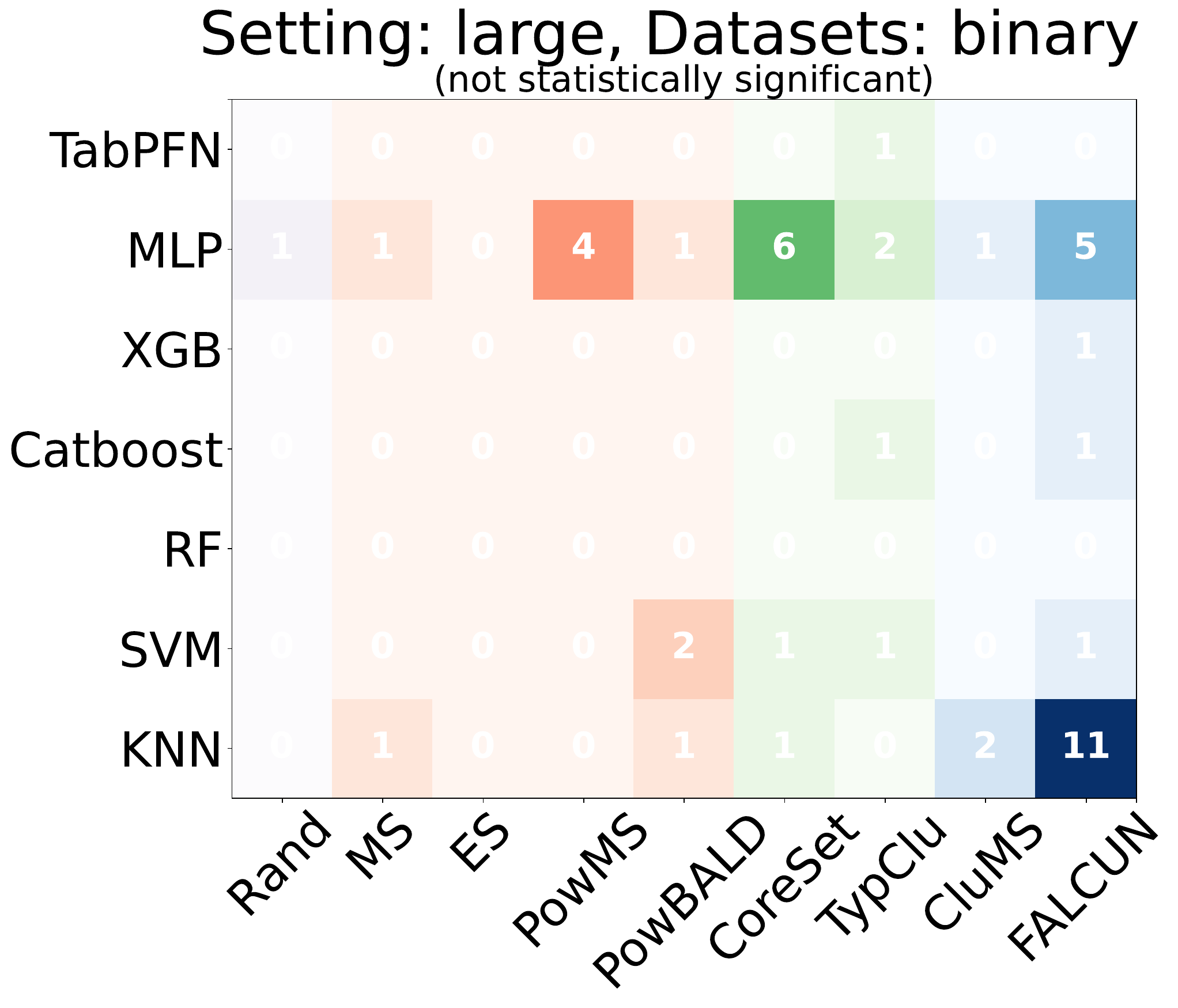}
    \end{subfigure}
    \begin{subfigure}[b]{0.24\textwidth}
        \centering
        \includegraphics[width=\textwidth]{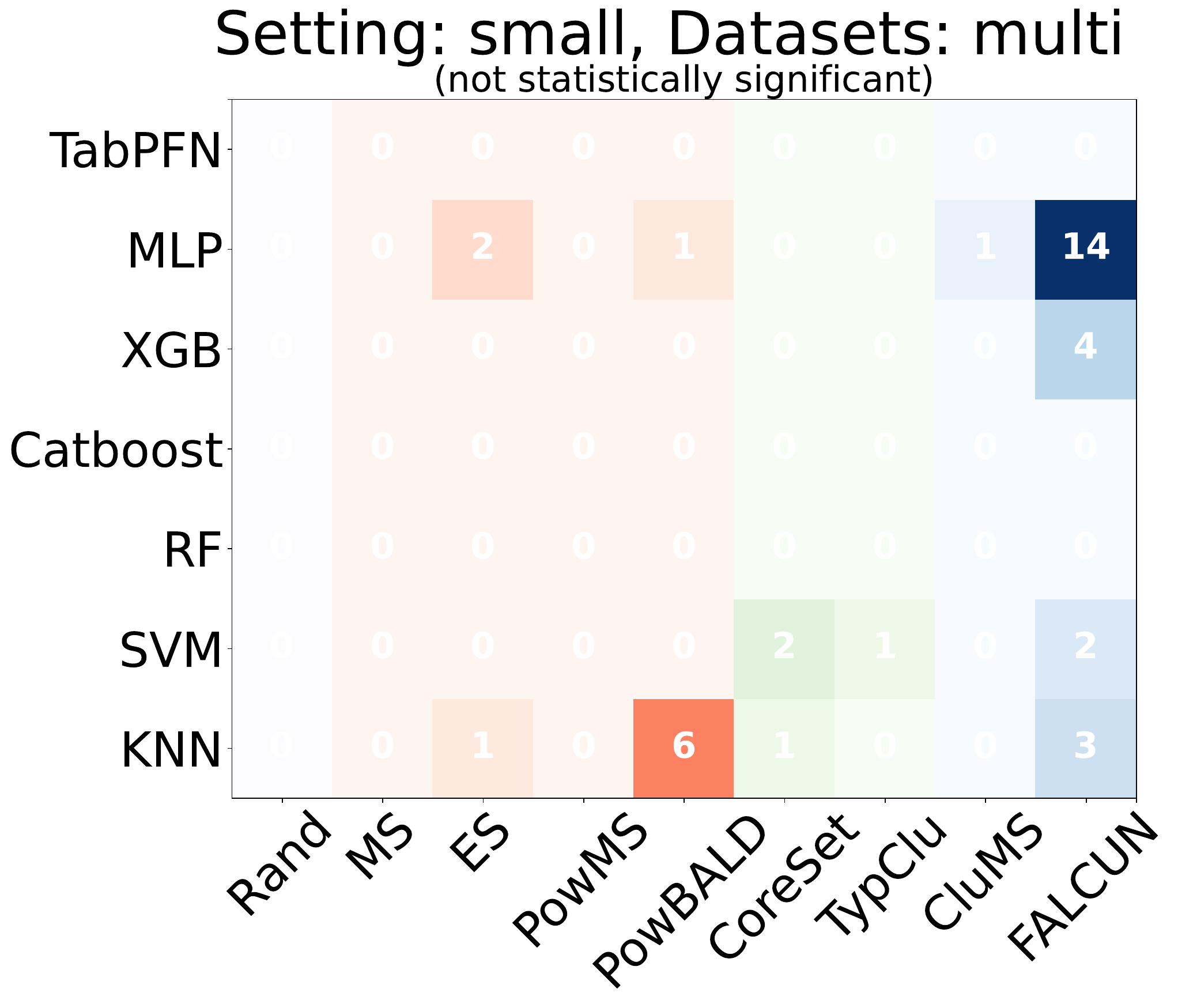}
    \end{subfigure}
    \begin{subfigure}[b]{0.24\textwidth}
        \centering
        \includegraphics[width=\textwidth]{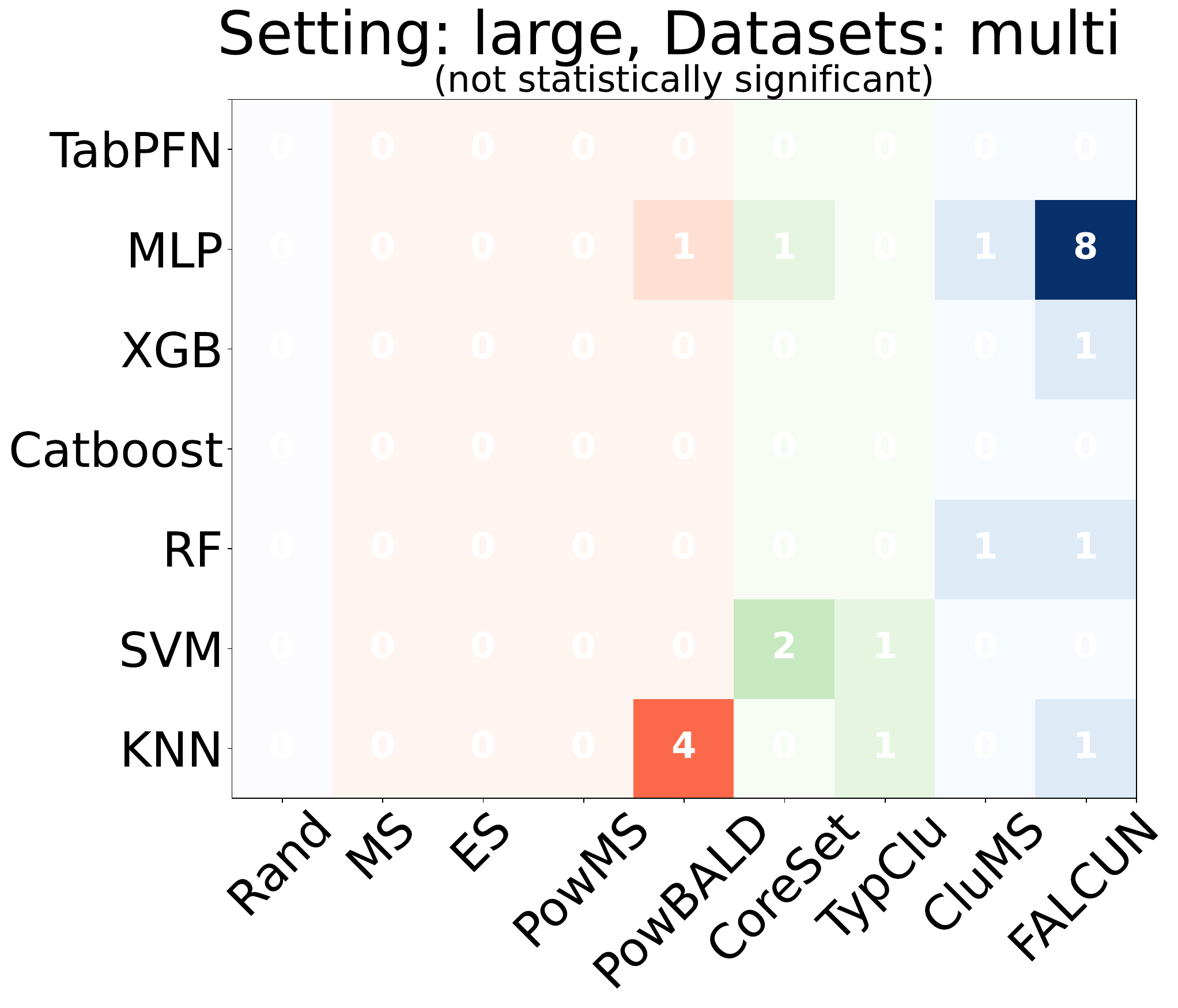}
    \end{subfigure}
    \caption{Lose-Heatmaps for all \acp{alp} excluding TabNet \textbf{without} statistical significance, considering \textbf{binary} and \textbf{multiclass} datasets. The color-coding is consistent with Figure~\ref{fig:heatmaps}.}
    \label{fig:loose_heatmaps}
\end{figure}
\textit{Budget curves.} Budget curves quantify the (test) performance of an \ac{alp} at each round of the \ac{al} procedure. The \ac{aubc} then offers a robust metric to compare different \acp{alp}. Tailored to our benchmark, we generate budget curves for each dataset and learning algorithm.

\textit{Win-matrices.} Further, we compute a win-matrix $W$ for each learning algorithm, comparing the performances of query strategies for that learner. Sticking to the notation above, this results in an $M \times M$ matrix $W$ for each learner. To make the plots of the win-matrices visually more appealing, we slightly modify the definition of the entry at position $(i,j)$ compared to~\cite{scarf,margin_all_you_need} as 
\begin{equation*}
W_{(i,j)} = \frac{\sum_{d=1}^{D} \mathds{1}[\text{\ac{qs} i beats \ac{qs} j on dataset d}]} 
{D}
\end{equation*}
Again, wins and losses are only defined in case of statistical significance. To determine a win or a loss, respectively, we compare the \ac{aubc} of two \acp{qs} after the total amount of iterations. This provides us with a robust measure since the overall performance across all iterations is captured.

\subsection{Results}\label{sec:results}
In this section, we present our main insights, aiming to answer the following research questions (RQ): \textit{RQ1:} Which \acp{alp} perform best and worst? \textit{RQ2:} Given a specific learning algorithm, which \ac{qs} is particularly well suited? \textit{RQ3:} Are there datasets and/or settings where \ac{al} leads to a decrease in performance? 

\begin{figure}[t]
    \centering
    \begin{subfigure}[b]{0.32\textwidth}
        \centering
        \includegraphics[width=\textwidth]{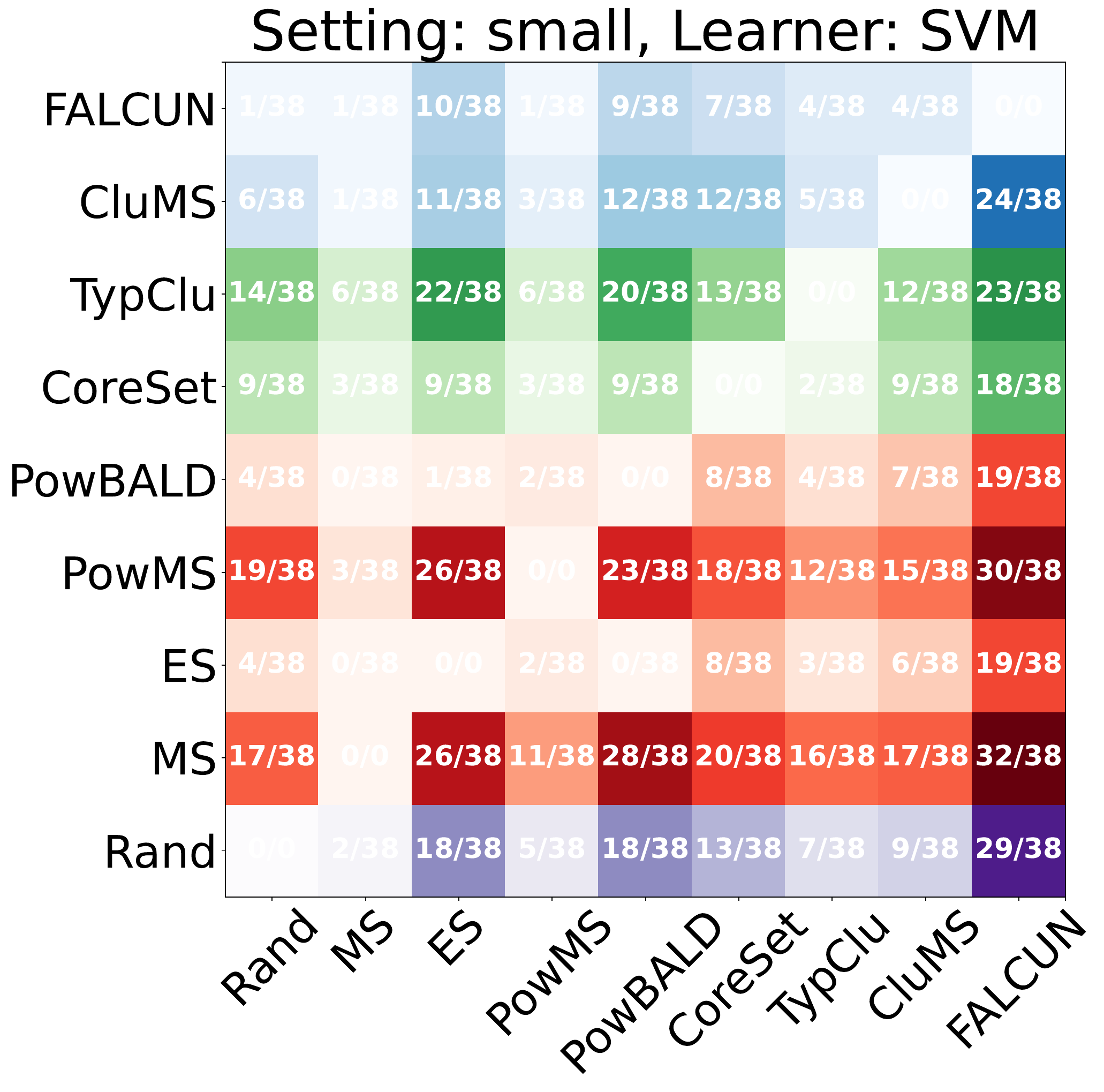}
    \end{subfigure}
    \begin{subfigure}[b]{0.32\textwidth}
        \centering
        \includegraphics[width=\textwidth]{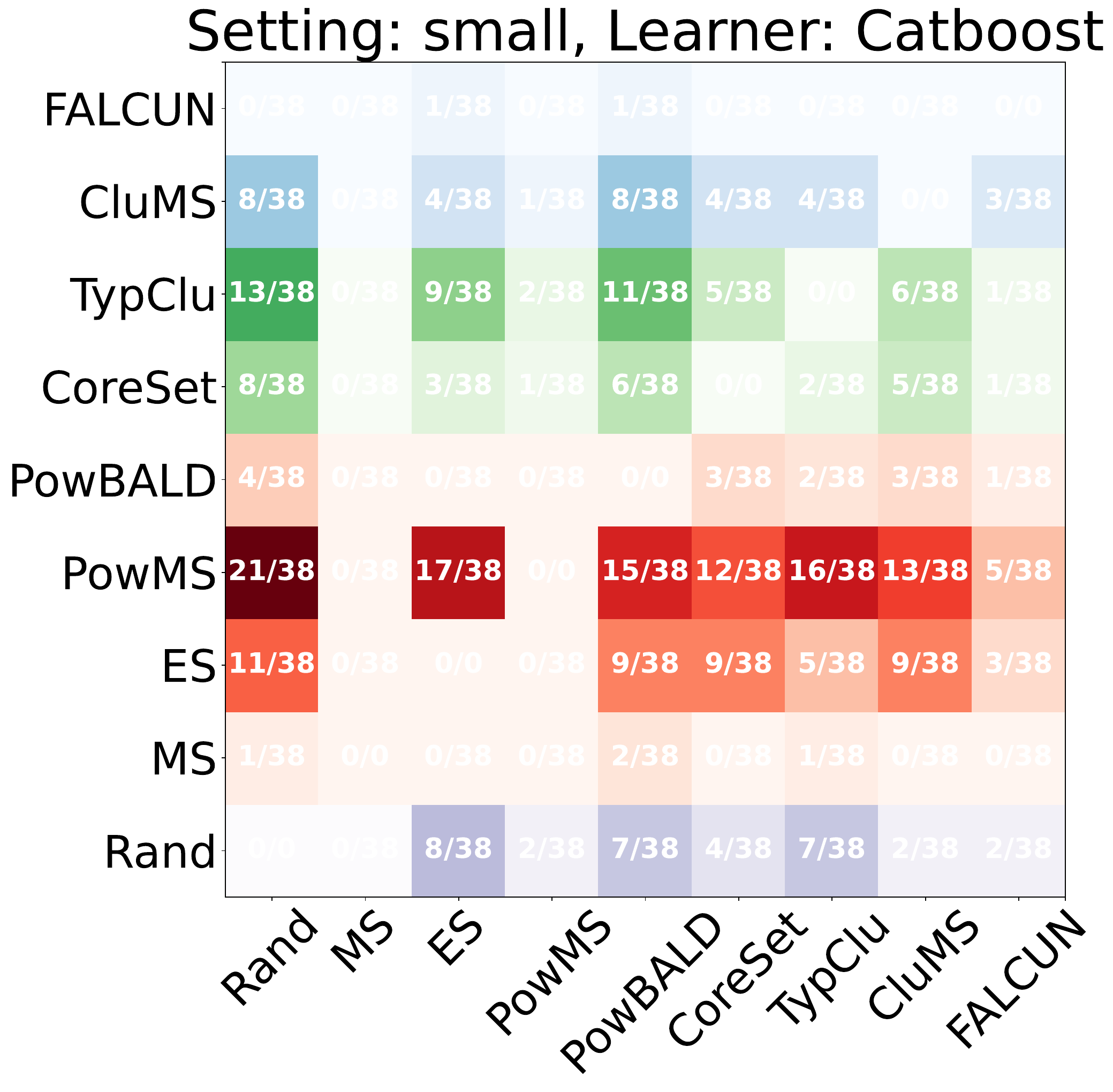}
    \end{subfigure}
    \begin{subfigure}[b]{0.32\textwidth}
        \centering
        \includegraphics[width=\textwidth]{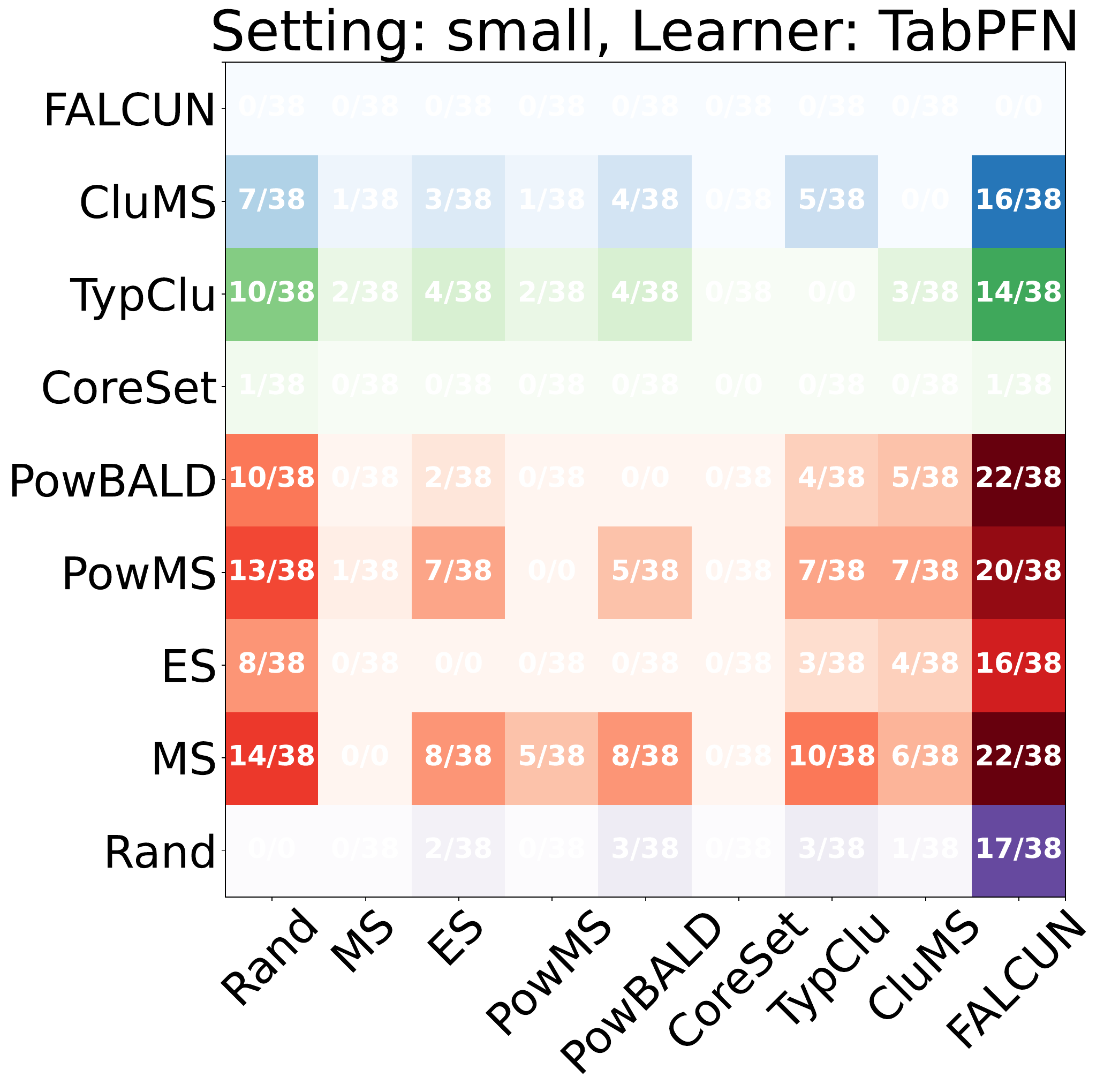}
    \end{subfigure}
    \caption{Win-Matrices for different learners (SVM, Catboost and TabPFN) for the \textbf{small} setting considering \textbf{multi-class} datasets \textbf{with} statistical significance.}
    \label{fig:win_win_matrices}
\end{figure}
\textit{RQ1.} In Figure~\ref{fig:heatmaps}, we show heatmaps as described in Section~\ref{sec:eval_methods}. \Ac{rf}, Catboost, and TabPFN are quite dominant, as they constitute to many winning \acp{alp}. \Ac{xgb} also performs well overall, however, showing a preference for large over small settings. TabNet, \ac{mlp}, and \ac{knn} are performing inferior, which, in the case of TabNet, might be due to limited training time. In the third and fourth subfigure, we present heatmaps without considering statistical significance, which may seem an unusual perspective. However, it reveals insights into which \acp{alp} reach peak performance for each dataset, showcasing that almost every combination of learning algorithm and \ac{qs} at least once constitutes a winning pipeline. While Catboost and TabPFN are still quite dominant, we find that repr.-based and hybr. \acp{qs} perform better in the small setting. This is reasonable, since the number of labeled instances might not suffice to learn the data distribution in this setting, which initially might favor querying diverse instances. In contrast, in the large setting, the learner is initially given enough instances to roughly learn the data distribution and then gains more when instances close to the decision boundary are queried.

To investigate which \acp{alp} perform particularly poorly, we present \textit{Lose-Heatmaps} in Figure~\ref{fig:loose_heatmaps}, where the losing pipeline replaces \ac{alp}$_d$. We again neglect statistical significance and consider binary and multi-class datasets separately, to investigate class-dependent differences. In this figure, we exclude TabNet as it did not perform at all in our investigated setting. We find that it is more important to choose a strong learner than selecting a suitable \ac{qs} and that repr. and hybr. \acp{qs} are better suited for multi-class than for binary datasets. Concretely, one should avoid using \ac{mlp} or \ac{knn}, and \acp{alp} combining \ac{knn} with \acs{powbald} or \ac{mlp} with \acs{falcun} proved disadvantageous.

\textit{RQ2.} In Figure~\ref{fig:win_win_matrices}, we present win-matrices for different learning algorithms. We choose \ac{svm} since it has been chosen as a learning algorithm in other \ac{al} studies, such as~\cite{ijcai_poolbased_al_benchmark,rebenchmarking_al}. Further, we choose Catboost and TabPFN, as they have shown strong performance when combined into \acp{alp}. The win matrices clearly show that the suitability of different \acp{qs} varies, depending on the given learning algorithm. Specifically for \ac{svm}, \ac{rand} outperforms most other \acp{qs} on roughly a third of the datasets, which does not hold for Catboost nor TabPFN. While the dependence of the suitability of different \acp{qs} on the chosen learning algorithm may not be surprising, it highlights that findings from \ac{al} benchmarks considering only one learning algorithm~\cite{margin_all_you_need,ijcai_poolbased_al_benchmark,rebenchmarking_al,benchmark_log_reg} are biased towards that particular learning algorithm. This reinforces the need for a joint benchmark as \tool.

\begin{figure}[t]
    \centering
    \begin{subfigure}[b]{0.24\textwidth}
        \centering
        \includegraphics[width=\textwidth]{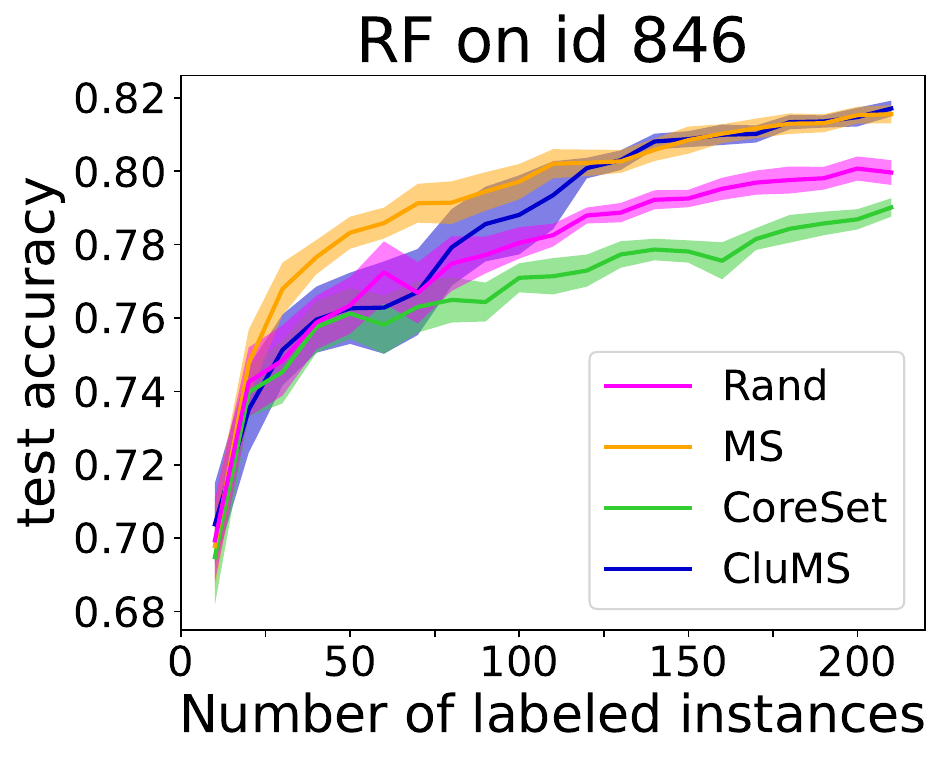}
    \end{subfigure}
    \begin{subfigure}[b]{0.24\textwidth}
        \centering
        \includegraphics[width=\textwidth]{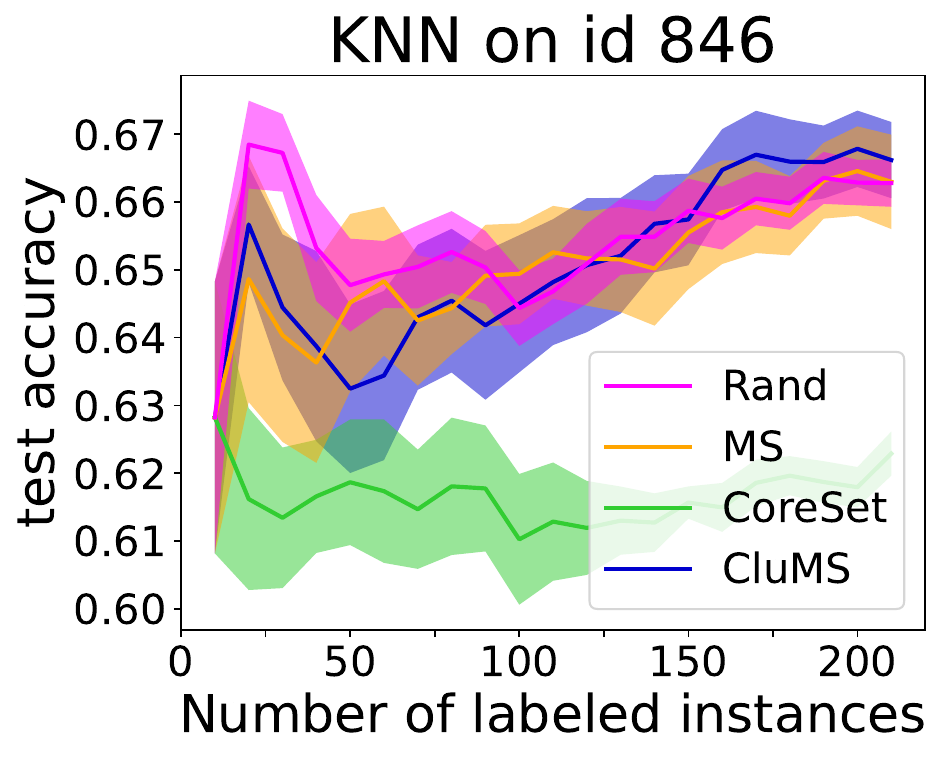}
    \end{subfigure}
    \begin{subfigure}[b]{0.24\textwidth}
        \centering
        \includegraphics[width=\textwidth]{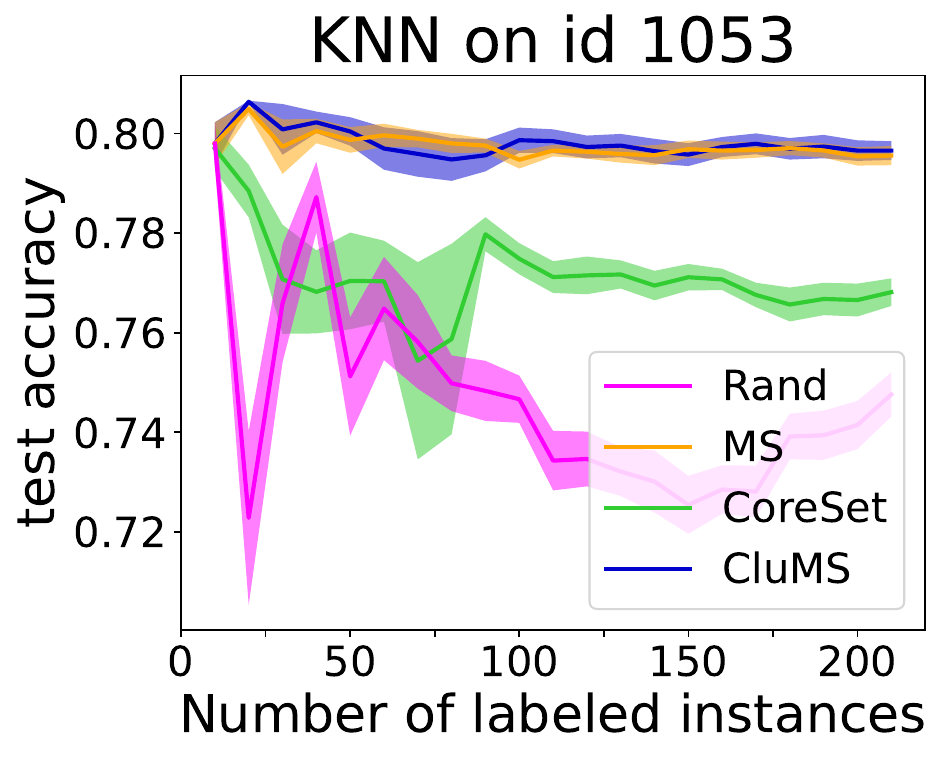}
    \end{subfigure}
    \begin{subfigure}[b]{0.24\textwidth}
        \centering
        \includegraphics[width=\textwidth]{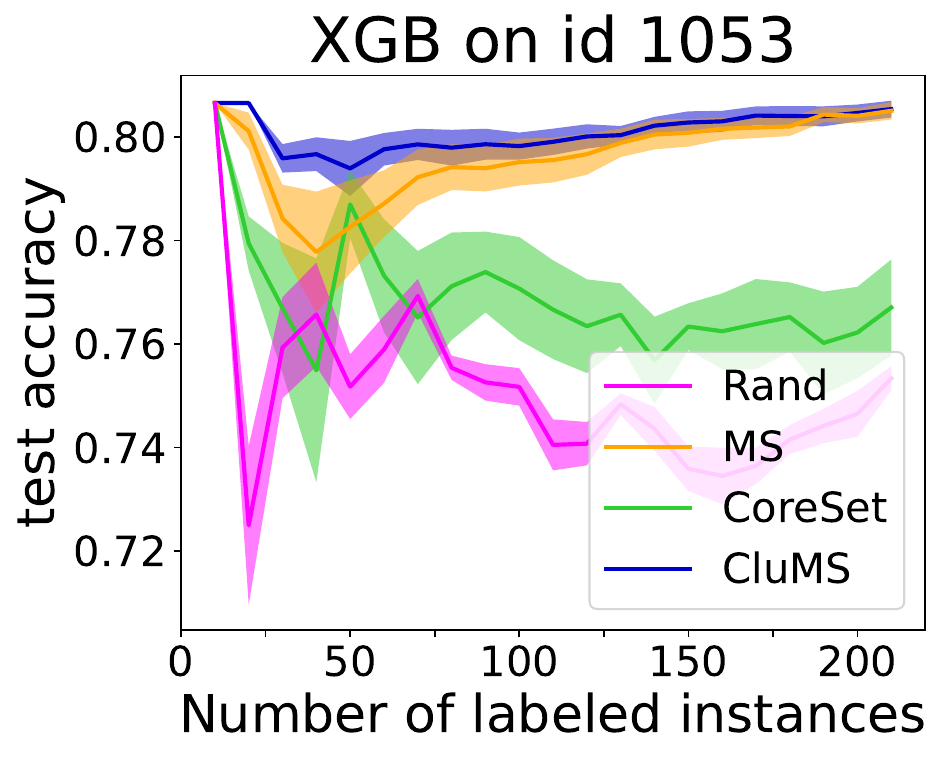}
    \end{subfigure}
    \caption{Budget curves for different \acp{alp} combined out of \ac{rf}, \ac{knn}, \ac{xgb} and \ac{rand}, \ac{ms}, \acs{coreset} and \ac{clums} on different datasets, considering the \textbf{small} setting.}
    \label{fig:budget_curves}
\end{figure}

\textit{RQ3.} In Figure~\ref{fig:budget_curves}, we present budget curves for \ac{rf}, \ac{knn}, and \ac{xgb} on two datasets (OpenML ID 846 and 1053) in the small setting. For better visual clarity, we only combine the learners with \ac{rand}, \ac{ms}, \acs{coreset}, and \ac{clums}, each representing a different family of \acp{qs}.

Budget curves in \ac{al} are generally expected to show an upward trend, indicating improved performance with an increasing budget, as visual in the first subfigure. However, this pattern is not consistent across all learners, as for the combination of \ac{knn} and \acs{coreset}, the performance decreases. On a different dataset (third subfigure), this also holds for random sampling and slightly for \ac{ms} and \ac{clums}. Even for a learning algorithm that demonstrates a strong overall performance, the picture can look quite similar, as shown in the fourth subfigure. To conclude, \ac{al} can deteriorate performance, as has also been shown in~\cite{schuurmans_07,gasperin-2009}. We also want to emphasize again the strong dependence of the performance of \acp{alp} on the chosen learning algorithm, dataset, setting, and potentially other properties, which still need to be understood. We hope that \tool will serve as a tool to gain more insights into this.

\section{Conclusion and Future Work}\label{sec:disc_future_work}
We proposed \tool, a benchmark for \acfp{alp} on tabular data. \tool allows for easily combining \acp{qs} and learning algorithms into \acp{alp} and provides a unified API to evaluate and benchmark them against each other. The open-source implementation of our benchmark is available as a Python package. 

In the benchmark so far, we focused on tabular classification tasks. A natural extension to \tool would thus be to add more benchmark families such as regression tasks or different types of data, such as image data. We predefined five different settings, which were partly inspired by~\cite{margin_all_you_need}. However, the exploration of more settings having different requirements for suitable pairs of \acp{qs} and learning algorithms outlines an interesting avenue for future work. Further, it might be appealing to incorporate other more recent trends, such as label noise, multiple annotators, etc.

In our experimental evaluation, we find that most of the time, strong pipelines consist of learners such as \ac{rf}, Catboost, or TabPFN and information-based query strategies. However, there are also datasets for which representation-based and hybrid approaches achieve better results. Also, the suitability of a \ac{qs} heavily depends on the chosen learner and the number of classes of the dataset.

With this benchmark and library, we hope to foster further research to fairly evaluate new \ac{qs} considering different datasets, settings, and learners. Moreover, it might be appealing to specifically develop new \acp{qs} for certain settings and/or learners. Lastly, we would also like to study whether it might be advantageous to devise hyperheuristics switching between different \acp{qs} within one active learning procedure. Based on the findings in Figure~\ref{fig:heatmaps}, it might be a good idea to first focus more on diversity and later on uncertainty.

\section{Limitations and Broader Impact Statement}\label{sec:lim}
Both the benchmark and the evaluation study are limited to tabular classification problems and consider a specific set of active learning settings. Furthermore, in the empirical study, we restricted the training time to 180 seconds per iteration, which might limit generalizability for the large settings. Nevertheless, we observe complementary performance for both learning algorithms and query strategies, which underpins the need for a benchmark as \tool.

This is also one of the reasons why we believe that \tool has the clear potential to foster more research on active learning and may incentivize the development of new query strategies that are adaptive to the learning algorithm's behavior.
As this paper presents work that aims to advance the field of machine learning, there are many potential societal consequences of our work. However, we feel that none of these needs to be specifically highlighted here.

{
\small

\clearpage

\begin{ack}
The authors gratefully acknowledge the computing time provided to them on the high-performance computers Noctua 2 at the NHR Center PC$^2$, granted under the AProSys project funded by Bundesministerium für Wirtschaft und Klimaschutz (BMWK), Förderkennzeichen: 03EI6090E. These are funded by the German Federal Ministry of Education and Research and the state governments participating on the basis of the resolutions of the GWK for the national high-performance computing at universities (www.nhr-verein.de/unsere-partner).
\end{ack}

\bibliographystyle{apalike}
\bibliography{references}


\newpage
\section*{NeurIPS Paper Checklist}

\begin{enumerate}

\item {\bf Claims}
    \item[] Question: Do the main claims made in the abstract and introduction accurately reflect the paper's contributions and scope?
    \item[] Answer: \answerYes{} 
    \item[] Justification: Yes, the claims presented in abstract and introduction of the paper are carefully tailored based on the experimental design and results obtained from it. We show that other benchmarks ignore crucial aspects of pipeline synthesis, often neglecting one or multiple steps of pipeline creation. We provide a standardized setting for active learning pipeline benchmarking (see Sections \ref{sec:alpb} and \ref{sec:exp}) and extract insights from the performance of different configurations (see Sections \ref{sec:results}).

\item {\bf Limitations}
    \item[] Question: Does the paper discuss the limitations of the work performed by the authors?
    \item[] Answer: \answerYes{} 
    \item[] Justification: Limitations of the paper are discussed in Section \ref{sec:lim}, showing possible additional perspectives that should be considered when investigating active learning pipelines.

\item {\bf Theory Assumptions and Proofs}
    \item[] Question: For each theoretical result, does the paper provide the full set of assumptions and a complete (and correct) proof?
    \item[] Answer: \answerNA{} 
    \item[] Justification: The paper does not contain theoretical proofs and results.

    \item {\bf Experimental Result Reproducibility}
    \item[] Question: Does the paper fully disclose all the information needed to reproduce the main experimental results of the paper to the extent that it affects the main claims and/or conclusions of the paper (regardless of whether the code and data are provided or not)?
    \item[] Answer: \answerYes{} 
    \item[] Justification: Yes, the paper provides detailed descriptions on experiment formulations, datasets, configurations and measurements. Section \ref{sec:alpb} describes how the benchmark was designed and demonstrates an example on how to reproduce results. Section \ref{sec:exp} details the used materials and evaluation methods. Moreover, we provide a library to facilitate reproducibility that also enables possible extensions for practitioners. 

\item {\bf Open access to data and code}
    \item[] Question: Does the paper provide open access to the data and code, with sufficient instructions to faithfully reproduce the main experimental results, as described in supplemental material?
    \item[] Answer: \answerYes{} 
    \item[] Justification: Yes, all the used datasets and materials are public. Moreover, we provide the code with clear instructions to enable readers to run the same experiments reported in the paper.

\item {\bf Experimental Setting/Details}
    \item[] Question: Does the paper specify all the training and test details (e.g., data splits, hyperparameters, how they were chosen, type of optimizer, etc.) necessary to understand the results?
    \item[] Answer: \answerYes{} 
    \item[] Justification: Yes, the experimental setting is carefully described in Sections \ref{sec:alpb} and \ref{sec:exp}. The description includes dataset manipulation, experimental setup (dataset splits and active learning settings), learning models and evaluation methods. 

\item {\bf Experiment Statistical Significance}
    \item[] Question: Does the paper report error bars suitably and correctly defined or other appropriate information about the statistical significance of the experiments?
    \item[] Answer: \answerYes{} 
    \item[] Justification: In Section \ref{sec:eval_methods} we provide information about statistical significance of the experiments. As mentioned there, we use Welch's t-test with $p = 0.05$ to determine statistical sinificance.

\item {\bf Experiments Compute Resources}
    \item[] Question: For each experiment, does the paper provide sufficient information on the computer resources (type of compute workers, memory, time of execution) needed to reproduce the experiments?
    \item[] Answer: \answerYes{} 
    \item[] Justification: Yes, computational resources are reported in Appendix \ref{sec:apx-experiments}. All experiments were conducted with 2 CPU cores and 8GiB RAM or 16GiB for the small
and large settings, respectively, to resemble end-user environments. The HPC nodes for the computations are
equipped with two AMD Milan 7763 and 256GiB main memory in total. Runs exceeding these limits have been
canceled by the workload manager.
    
\item {\bf Code Of Ethics}
    \item[] Question: Does the research conducted in the paper conform, in every respect, with the NeurIPS Code of Ethics \url{https://neurips.cc/public/EthicsGuidelines}?
    \item[] Answer: \answerYes{} 
    \item[] Justification: Yes, the paper follows NeurIPS Code of Ethics with no harms caused by the research process.

\item {\bf Broader Impacts}
    \item[] Question: Does the paper discuss both potential positive societal impacts and negative societal impacts of the work performed?
    \item[] Answer: \answerYes{} 
    \item[] Justification: Yes, the paper thoroughly discusses the societal impacts in the broader impact statement in Section \ref{sec:lim}.
    
\item {\bf Safeguards}
    \item[] Question: Does the paper describe safeguards that have been put in place for responsible release of data or models that have a high risk for misuse (e.g., pretrained language models, image generators, or scraped datasets)?
    \item[] Answer: \answerNA{} 
    \item[] Justification: The paper does not present such risks.

\item {\bf Licenses for existing assets}
    \item[] Question: Are the creators or original owners of assets (e.g., code, data, models), used in the paper, properly credited and are the license and terms of use explicitly mentioned and properly respected?
    \item[] Answer: \answerYes{} 
    \item[] Justification: Yes, all data and materials used are publicly available under the MIT License.

\item {\bf New Assets}
    \item[] Question: Are new assets introduced in the paper well documented and is the documentation provided alongside the assets?
    \item[] Answer: \answerYes{} 
    \item[] Justification: Yes, we provide a new library for active learning benchmarking under the MIT license.

\item {\bf Crowdsourcing and Research with Human Subjects}
    \item[] Question: For crowdsourcing experiments and research with human subjects, does the paper include the full text of instructions given to participants and screenshots, if applicable, as well as details about compensation (if any)? 
    \item[] Answer: \answerNA{} 
    \item[] Justification: The paper does not involve crowdsourcing nor research with human subjects.

\item {\bf Institutional Review Board (IRB) Approvals or Equivalent for Research with Human Subjects}
    \item[] Question: Does the paper describe potential risks incurred by study participants, whether such risks were disclosed to the subjects, and whether Institutional Review Board (IRB) approvals (or an equivalent approval/review based on the requirements of your country or institution) were obtained?
    \item[] Answer: \answerNA{} 
    \item[] Justification: The paper does not involve crowdsourcing nor research with human subjects.

\end{enumerate}

\clearpage

\appendix
\section{Appendix}

\subsection{Glossary of Acronyms}

\begin{acronym}\itemsep0pt
\acro{al}[AL]{active learning}
\acro{alp}[ALP]{active learning pipeline}
\acro{aubc}[AUBC]{area under the budget curve}
\acro{ds}[DS]{dataset}
\acro{ml}[ML]{machine learning}
\acro{qs}[QS]{query strategy}
\acrodefplural{qs}{query strategies}
\acro{sota}[SOTA]{state-of-the-art}
\acro{dnn}[DNN]{deep neural network}
\acro{etc}[ETC]{extremely randomized trees}
\acro{gbdt}[GBDT]{gradient-boosted decision tree}
\acro{knn}[k-NN]{k-nearest neighbor}
\acro{lr}[LR]{logistic regression}
\acro{mlp}[MLP]{multi-layer perceptron}
\acro{nb}[NB]{na{\"i}ve Bayes}
\acro{pfn}[PFN]{prior-fitted network}
\acro{rf}[RF]{random forest}
\acro{svm}[SVM]{support vector machine}
\acro{xgb}[XGB]{XGBoost}
\acro{aal}[AAL]{adaptive active learning}
\acro{albl}[ALBL]{active learning by learning}
\acro{bald}[BALD]{Bayesian active learning by disagreement}
\acro{cer}[CER]{combined error reduction}
\acro{clue}[CLUE]{clustering uncertainty-weighted embeddings}
\acro{clums}[CluMS]{cluster margin}
\acro{coreset}[CoreSet]{CoreSet}
\acro{dwus}[DWUS]{density weighted uncertainty sampling}
\acro{eer}[EER]{expected error reduction}
\acro{emc}[EMC]{expected model change}
\acro{es}[ES]{entropy sampling}
\acro{eu}[EU]{epistemic uncertainty sampling}
\acro{evr}[EVR]{expected variance reduction}
\acro{falcun}[FALCUN]{fast active learning by contrastive uncertainty}
\acro{fivr}[FIVR]{Fisher information variance reduction}
\acro{graph}[GRAPH]{graph density}
\acro{hier}[HIER]{hierarchical sampling}
\acro{lc}[LC]{least-confident sampling}
\acro{kmeans}[k-means]{k-means sampling}
\acro{margdens}[MarginDensity]{pre-clustering and margin sampling}
\acro{maxent}[MaxEnt]{maximum entropy}
\acro{maxer}[MaxER]{maximum error reduction}
\acro{minmarg}[Min Margin]{minimum margin sampling}
\acro{mli}[MLI]{minimum loss increase}
\acro{mmc}[MMC]{maximum model change}
\acro{ms}[MS]{margin sampling}
\acro{powbald}[PowBALD]{power-set \acs{bald}}
\acro{powms}[PowMS]{power-set margin sampling}
\acro{qbc}[QBC]{query-by-committee}
\acro{qbcvr}[QBC VR]{\acs{qbc} \acs{vr}}
\acro{quire}[QUIRE]{querying informative and representative examples}
\acro{rand}[Rand]{random sampling}
\acro{typclu}[TypClu]{typical clustering}
\acro{vr}[VR]{variance reduction}

\end{acronym}

\subsection{Comparison to existing benchmarks}
In the following, we present an extensive tabular which compares \tool with existing active learning benchmarks. The \ac{qs} and learning algorithms are ordered by their year of appearance.

\newcommand{\gcmark}{\textcolor{ao}{\cmark}}
\newcommand{\rxmark}{\textcolor{gray}{\xmark}}
\resizebox{\textwidth}{!}{
\begin{tabular}{l|c|c|c|c|c|c}
\toprule
    Query Strategy & Year & \cite{benchmark_log_reg} (\citeyear{benchmark_log_reg}) & \cite{ijcai_poolbased_al_benchmark} (\citeyear{ijcai_poolbased_al_benchmark}) & \cite{margin_all_you_need} (\citeyear{margin_all_you_need}) & \cite{rebenchmarking_al} (\citeyear{rebenchmarking_al}) & \tool (2024)\\
    \midrule
     \ac{es} \cite{shannon} & \citeyear{shannon} & \gcmark & \gcmark & \gcmark & \gcmark & \gcmark \\
     \ac{qbc} \cite{qbc} & \citeyear{qbc}     & \rxmark      & \gcmark & \rxmark      & \gcmark & \gcmark\\
     \ac{vr} \cite{variance_reduction} & \citeyear{variance_reduction}      & \rxmark      & \gcmark & \rxmark      & \gcmark & \gcmark \\
     \ac{lc} \cite{least_confident} & \citeyear{least_confident}      & \rxmark      & \gcmark & \gcmark & \gcmark & \gcmark \\
     \acs{fivr} \cite{fisher_for_al} & \citeyear{fisher_for_al}    & \gcmark & \rxmark      & \rxmark      & \rxmark & \rxmark \\
     \ac{ms} \cite{margin_sampling} & \citeyear{margin_sampling}      & \rxmark      & \gcmark & \gcmark & \gcmark & \gcmark \\
     \ac{eer} \cite{expected_error_reduction} & \citeyear{expected_error_reduction}     & \gcmark & \gcmark & \rxmark      & \gcmark & \gcmark \\
     \acs{maxer} \cite{max_error_reduction} & \citeyear{max_error_reduction}   & \gcmark & \rxmark      & \rxmark      & \rxmark & \rxmark\\
     \acs{cer} \cite{combined_error_reduction} & \citeyear{combined_error_reduction}     & \gcmark & \rxmark      & \rxmark      & \rxmark & \rxmark\\
     \acs{evr} \cite{al_for_logreg} & \citeyear{al_for_logreg}     & \gcmark & \rxmark      & \rxmark      & \rxmark & \rxmark\\
     \acs{emc} \cite{expected_model_change} & \citeyear{expected_model_change}     & \rxmark      & \gcmark & \rxmark      & \rxmark & \rxmark\\
     \acs{mli} \cite{minimum_loss_increase} & \citeyear{minimum_loss_increase}     & \gcmark & \rxmark      & \rxmark      & \rxmark & \rxmark\\
     \acs{bald} \cite{BALD} & \citeyear{BALD}    & \rxmark      & \rxmark      & \gcmark & \rxmark & \gcmark\\
     \acs{mmc} \cite{max_model_change} & \citeyear{max_model_change}     & \gcmark & \rxmark      & \rxmark      & \rxmark & \rxmark \\
     \ac{maxent} \cite{max_entropy} & \citeyear{max_entropy} & \rxmark  & \rxmark      & \gcmark & \rxmark & \gcmark \\
     \acs{qbcvr} \cite{qbc_VR} & \citeyear{qbc_VR}  & \rxmark      & \rxmark      & \gcmark & \rxmark & \gcmark \\
     \ac{eu} \cite{epistemic_uncertainty_sampling_eyke} & \citeyear{epistemic_uncertainty_sampling_eyke} & \rxmark & \rxmark & \rxmark & \rxmark & \gcmark \\
     \acs{powms} \cite{power_versions} & \citeyear{power_versions}   & \rxmark      & \rxmark      & \gcmark & \rxmark & \gcmark \\
     \acs{minmarg} \cite{min_margin} & \citeyear{min_margin} & \rxmark   & \rxmark      & \gcmark & \rxmark & \gcmark\\
     \midrule
     \ac{kmeans} \cite{kmeans} & \citeyear{kmeans} & \rxmark & \gcmark & \rxmark & \rxmark & \gcmark \\
     \acs{hier} \cite{hier} & \citeyear{hier} & \rxmark & \gcmark & \rxmark & \gcmark & \rxmark \\
     \acs{coreset} \cite{core_set} & \citeyear{core_set} & \rxmark & \rxmark & \gcmark & \gcmark & \gcmark \\
     \ac{typclu} \cite{typical_clustering} & \citeyear{typical_clustering} & \rxmark & \rxmark & \gcmark & \rxmark & \gcmark \\
     \acs{falcun} \cite{FALCUN_gilhuber} & \citeyear{FALCUN_gilhuber} & \rxmark & \rxmark & \rxmark & \rxmark & \gcmark \\
     \midrule
     \acs{margdens} \cite{margin_density} & \citeyear{margin_density} & \rxmark & \rxmark & \gcmark & \rxmark & \rxmark \\
     \acs{dwus} \cite{density_weighted} & \citeyear{density_weighted} & \rxmark & \gcmark & \rxmark & \gcmark & \rxmark \\
     \acs{quire} \cite{QUIRE} & \citeyear{QUIRE} & \rxmark & \gcmark & \rxmark & \gcmark & \rxmark \\
     \acs{graph} \cite{GRAPH} & \citeyear{GRAPH} & \rxmark & \gcmark & \rxmark & \gcmark & \rxmark \\
     \acs{aal} \cite{adaptive_active_learning} & \citeyear{adaptive_active_learning} & \gcmark & \rxmark & \rxmark & \rxmark & \rxmark \\
     \acs{albl} \cite{al_by_learning} & \citeyear{al_by_learning} & \rxmark & \gcmark & \rxmark & \gcmark & \rxmark \\
     \ac{clums} \cite{cluster_margin} & \citeyear{cluster_margin} & \rxmark & \rxmark & \gcmark & \rxmark & \gcmark \\
     \ac{clue} \cite{clue} & \citeyear{clue} & \rxmark & \rxmark & \rxmark & \rxmark & \gcmark \\
     \bottomrule
\end{tabular}}

\resizebox{\textwidth}{!}{
\begin{tabular}{l|c|c|c|c|c|c}
\toprule
    Learning Algorithm & Year & \cite{benchmark_log_reg} (\citeyear{benchmark_log_reg}) & \cite{ijcai_poolbased_al_benchmark} (\citeyear{ijcai_poolbased_al_benchmark}) & \cite{margin_all_you_need} (\citeyear{margin_all_you_need}) & \cite{rebenchmarking_al} (\citeyear{rebenchmarking_al}) & \tool (2024)\\
    \midrule
     \ac{lr} \cite{logisticregression} & \citeyear{logisticregression} & \gcmark & \rxmark & \rxmark & \rxmark & \gcmark \\
     \ac{knn} \cite{knn} & \citeyear{knn} & \rxmark & \rxmark & \rxmark & \rxmark & \gcmark \\
     \ac{mlp} \cite{mlp} & \citeyear{mlp} & \rxmark & \rxmark & \gcmark & \rxmark & \gcmark \\
     \ac{nb} \cite{naivebayes} & \citeyear{naivebayes} & \rxmark & \rxmark & \rxmark & \rxmark & \gcmark \\
     \ac{svm} \cite{svm} & \citeyear{svm} & \rxmark & \gcmark & \rxmark & \gcmark & \gcmark \\
     \ac{rf} \cite{leobreiman_rf} & \citeyear{leobreiman_rf} & \rxmark & \rxmark & \rxmark & \rxmark & \gcmark \\
     \ac{etc} \cite{extratrees} & \citeyear{extratrees} & \rxmark & \rxmark & \rxmark & \rxmark & \gcmark \\
     \ac{xgb} \cite{xgboost} & \citeyear{xgboost} & \rxmark & \rxmark & \rxmark & \rxmark & \gcmark \\
     Catboost \cite{catboost} & \citeyear{catboost} & \rxmark & \rxmark & \rxmark & \rxmark & \gcmark \\
     TabNet \cite{tabnet} & \citeyear{tabnet} & \rxmark & \rxmark & \rxmark & \rxmark & \gcmark \\
     TabPFN \cite{tabpfn} & \citeyear{tabpfn} & \rxmark & \rxmark & \rxmark & \rxmark & \gcmark \\
     \bottomrule
\end{tabular}}

\subsection{Predefined Settings}\label{sec:apx-settings}
In Table~\ref{tab:predefined-settings}, we list the different settings for configuring different active learning scenarios that are predefined and implemented in \tool.
\begin{table}[h]
    \centering
    \caption{Predefined active learning settings in \tool.}
    \label{tab:predefined-settings}
    \begin{tabular}{cccc}
        \toprule
         Name & $|D^0_L|$ & $B$ & $R$ \\
         \midrule
         small & 30 & 200 & 10 \\
         medium& 100 & 1,000 & 50 \\
         large& 300 & 10,000 & 200\\
         \midrule
         small (dynamic) & 10 & $100 \cdot |\mathcal{Y}|$ & $5 \cdot |\mathcal{Y}|$\\
         large (dynamic) & 10 & $400 \cdot |\mathcal{Y}|$ & $20 \cdot |\mathcal{Y}|$\\
         \bottomrule
    \end{tabular}
\end{table}

\subsection{Experiments}\label{sec:apx-experiments}

In this section, we elaborate in more detail on the experiments that were conducted within our evaluation study.

\textit{Datasets.} From the 90 datasets from the OpenML-CC18~\cite{cc18} and the TabZilla Benchmark Suite~\cite{when_nn_outperform_trees_on_tab} we filtered and excluded the datasets with OpenML IDs 1567, 1169, 41147, and 1493. The first three were filtered out for all settings because they consist of more than 300,000 data points, which would result in a large amount of computing time for the non-info. based \acp{qs}. The last dataset with OpenML ID 1493 was filtered out since it consists of 100 classes, which would result in a huge amount of $\mathcal{R}$, limiting the number of iterations to a high degree. Further, for the large setting, we wanted to guarantee that at least 10 iterations can be performed until all instances from $\DU$ are queried. This led to the removal of OpenML IDs 11, 12, 14, 16, 18, 22, 25, 51, 54, 188, 307, 458, 469, 1468, 1501, 40966, and 40979 for this setting. For the preprocessing steps, we proceed as follows. Categorical features are one-hot encoded and missing values are imputed by the mean or mode of the corresponding feature.

\textit{Active Learning Setting.} As mentioned, we investigate a small and a large setting. Explicitly, the small and large settings are specified by $|\DL^0| = R = 5 \cdot |\mathcal{Y}|$ and $|\DL^0| = R = 20 \cdot |\mathcal{Y}|$, respectively, for the given dataset and a total amount of 20 iterations or until all instances from the unlabeled pool $\DU$ are queried. We choose the factor 5 for the small setting, since then $\mathcal{R}$ matches the one in the (static) small setting in~\cite{margin_all_you_need}. For the large setting, we should have chosen a factor of 100 to be again consistent with~\cite{margin_all_you_need}. However, this seemed unrealistic to us for real-world applications. For some (imbalanced) datasets, it may happen that not every class is at least once represented in $\DL^0$. In these cases, we additionally randomly sample one instance from $\DU$ per missing class and add them with their corresponding label to $\DL^0$. 
We run each \ac{alp} ten times with different seeds, where the seed defines the $\frac{2}{3} / \frac{1}{3}$-split of the total dataset $\mathcal{D}$ into $\Dtrain$ and $\Dtest$ as well as the split of $\Dtrain$ into $\DL$ and $\DU$. Needless to say, the datasets we consider are originally (fully-)labeled datasets. Tailored to the AL setting, we discard the labels for the instances in $\DU$ and assure that only the oracle $\mathcal{O}$ can access them. 

\textit{Configuration of Learning Algorithms.}
In general, we do not perform any hyperparameter optimization (HPO) but rather stick to the default parameters. For \ac{xgb} and Catboost, we reduce the training time by setting the tree method to \textit{hist} and limiting the amount of iterations, respectively. For Catboost, we implement a timeout of three minutes for the same purpose. TabPFN~\cite{tabpfn} can so far only be fitted on a maximum amount of 1,000 instances. Therefore, we uniformly sample 1,000 instances from the current dataset to be fitted on, in case this constraint is violated, similar to~\cite{when_nn_outperform_trees_on_tab}. For TabPFN and TabNet we modify the implementation for the representation-based and hybrid approaches. Concretely, we extract the output of the encoder from the TabPFN and the activations of the penultimate layer from TabNet to compute the representativeness of each instance based on its embedding. The exact details can obviously be looked up in our implementation.

\textit{Implementation.}
All experiments were conducted with 2 CPU cores and 8GiB RAM or 16GiB for the small and large settings, respectively, to resemble end-user environments. The HPC nodes for the computations are equipped with two AMD Milan 7763 and 256GiB main memory in total. Runs exceeding these limits have been canceled by the workload manager.

\subsection{Results}\label{sec:apx-results}

This section contains more experimental results, comprising more heatmaps and win-matrices distinguishing between binary and multi-class datasets and more budget curves for other datasets and learners. Precisely, we first present heatmaps where we - compared to the main paper - distinguish between binary and multi-class datasets, cf. Figure~\ref{fig:heatmaps_binary} and~\ref{fig:heatmaps_multi}, respectively. The results for the binary setting look quite similar to the ones in the main paper, where we consider all datasets. However, for the multi-class datasets, there are some differences: Specifically, for large datasets, \ac{xgb} is dominating TabPFN and Catboost.

\begin{figure}[h]
    \centering
    \begin{subfigure}[b]{0.24\textwidth}
        \centering
        \includegraphics[width=\textwidth]{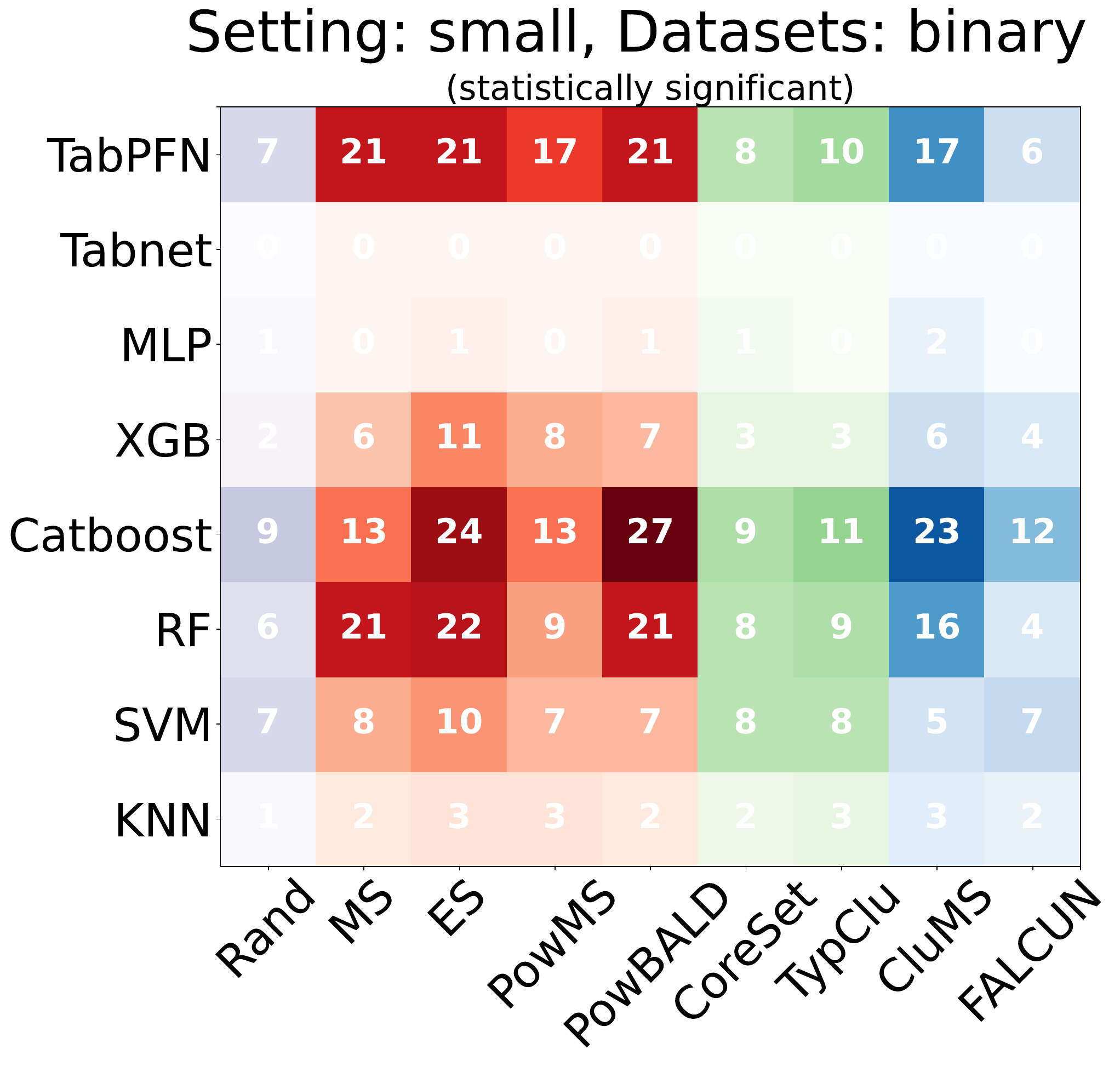}
    \end{subfigure}
    \begin{subfigure}[b]{0.24\textwidth}
        \centering
        \includegraphics[width=\textwidth]{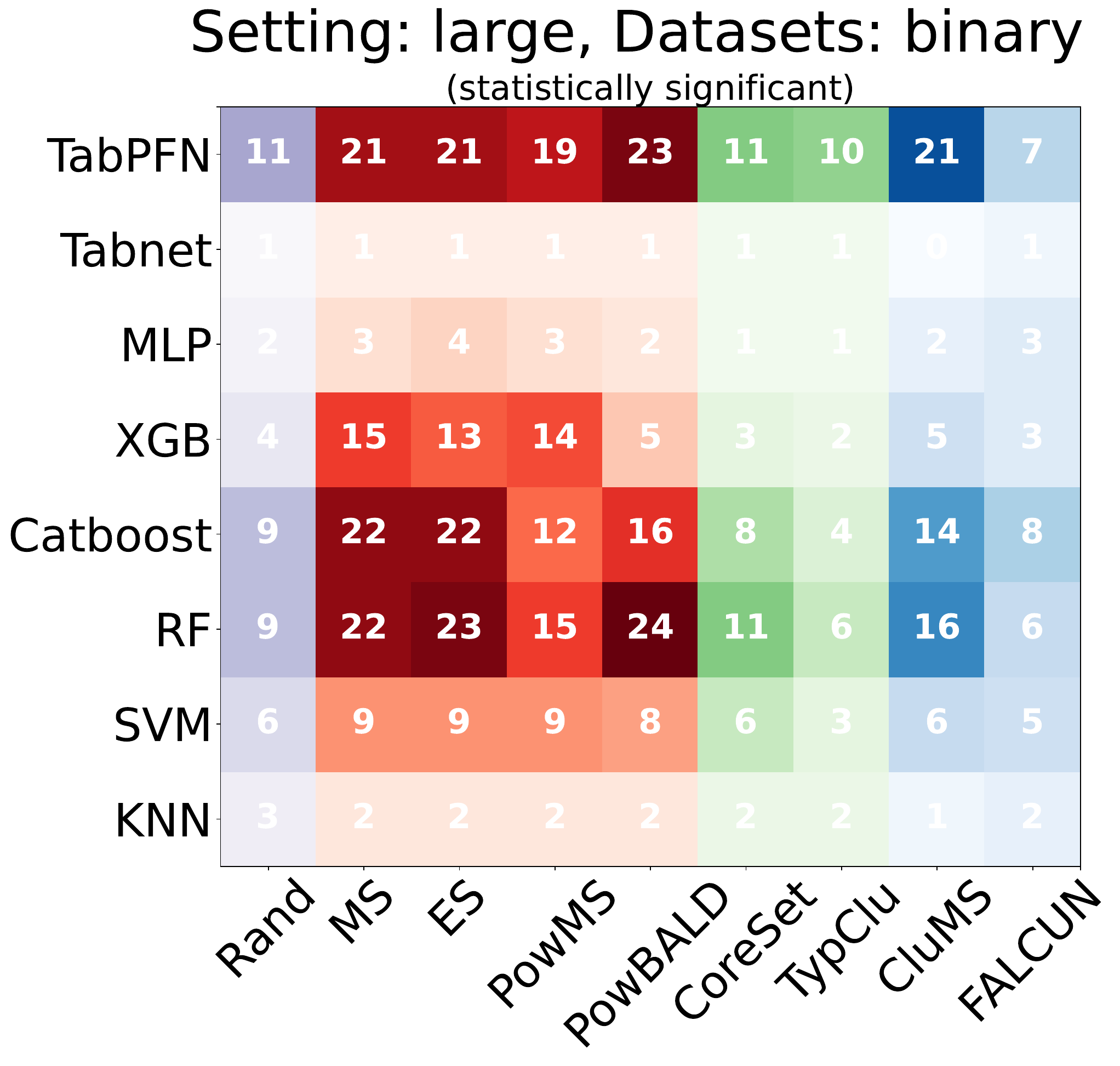}
    \end{subfigure}
    \begin{subfigure}[b]{0.24\textwidth}
        \centering
        \includegraphics[width=\textwidth]{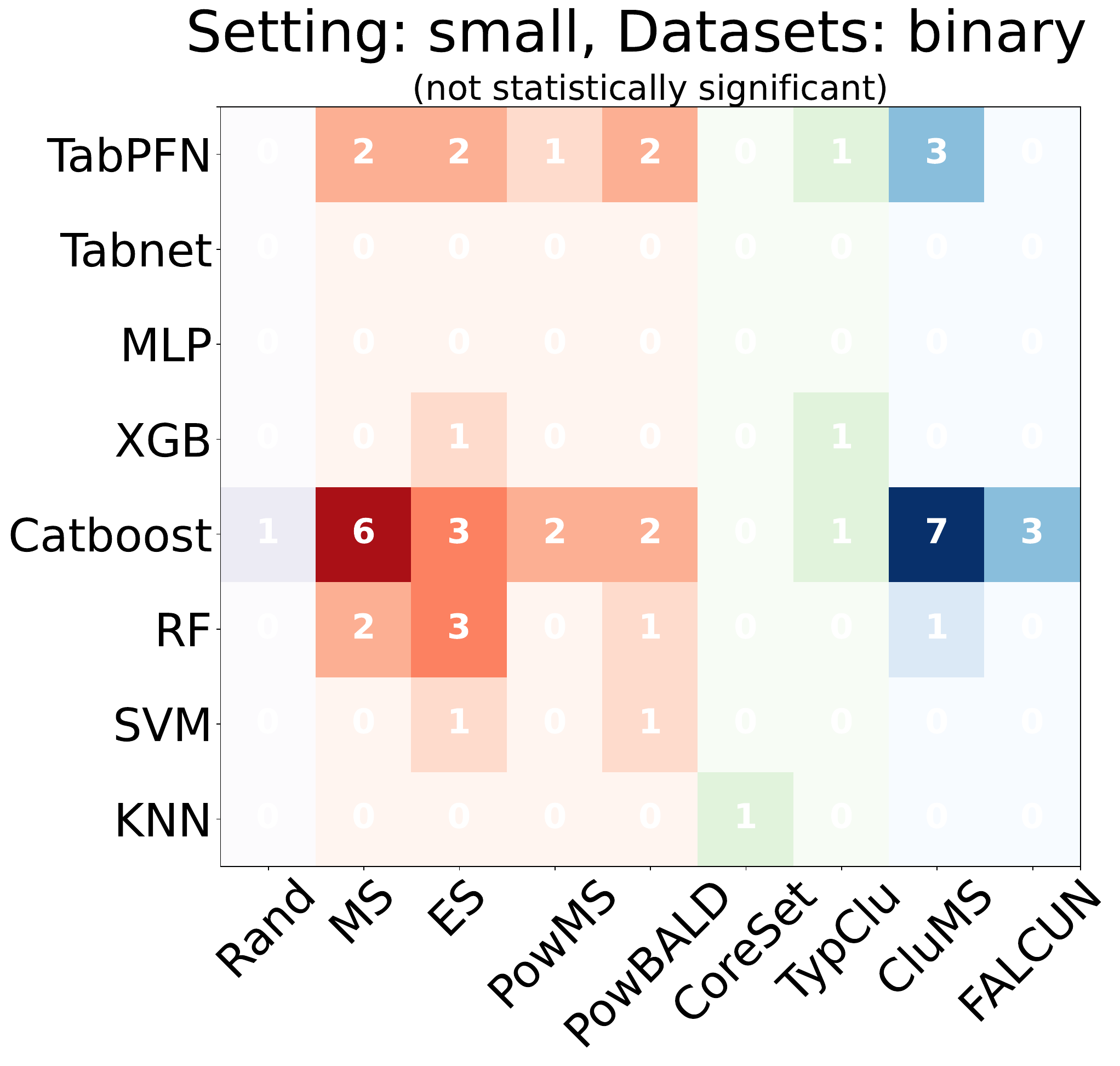}
    \end{subfigure}
    \begin{subfigure}[b]{0.24\textwidth}
        \centering
        \includegraphics[width=\textwidth]{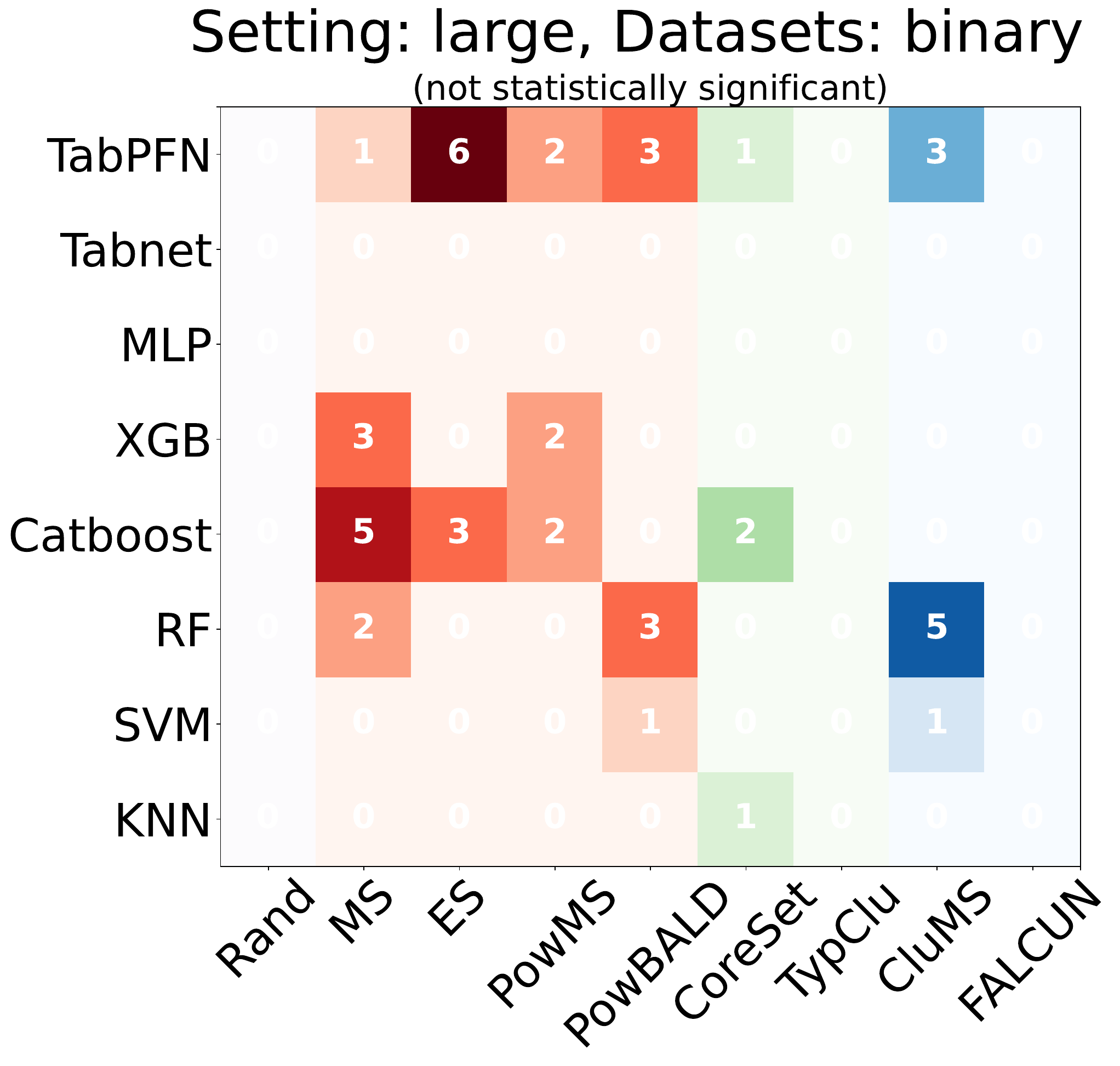}
    \end{subfigure}
    \caption{Heatmaps for all \acp{alp} within our evaluation study \textbf{with} statistical significance (first and second subfigure) and \textbf{without} (third and fourth) for \textbf{binary} datasets. Info.-based, repr.-based, and hybr. \acp{qs} are colored in red, green, and blue, respectively, and random sampling in purple.}
    \label{fig:heatmaps_binary}
\end{figure}

\begin{figure}[h]
    \centering
    \begin{subfigure}[b]{0.24\textwidth}
        \centering
        \includegraphics[width=\textwidth]{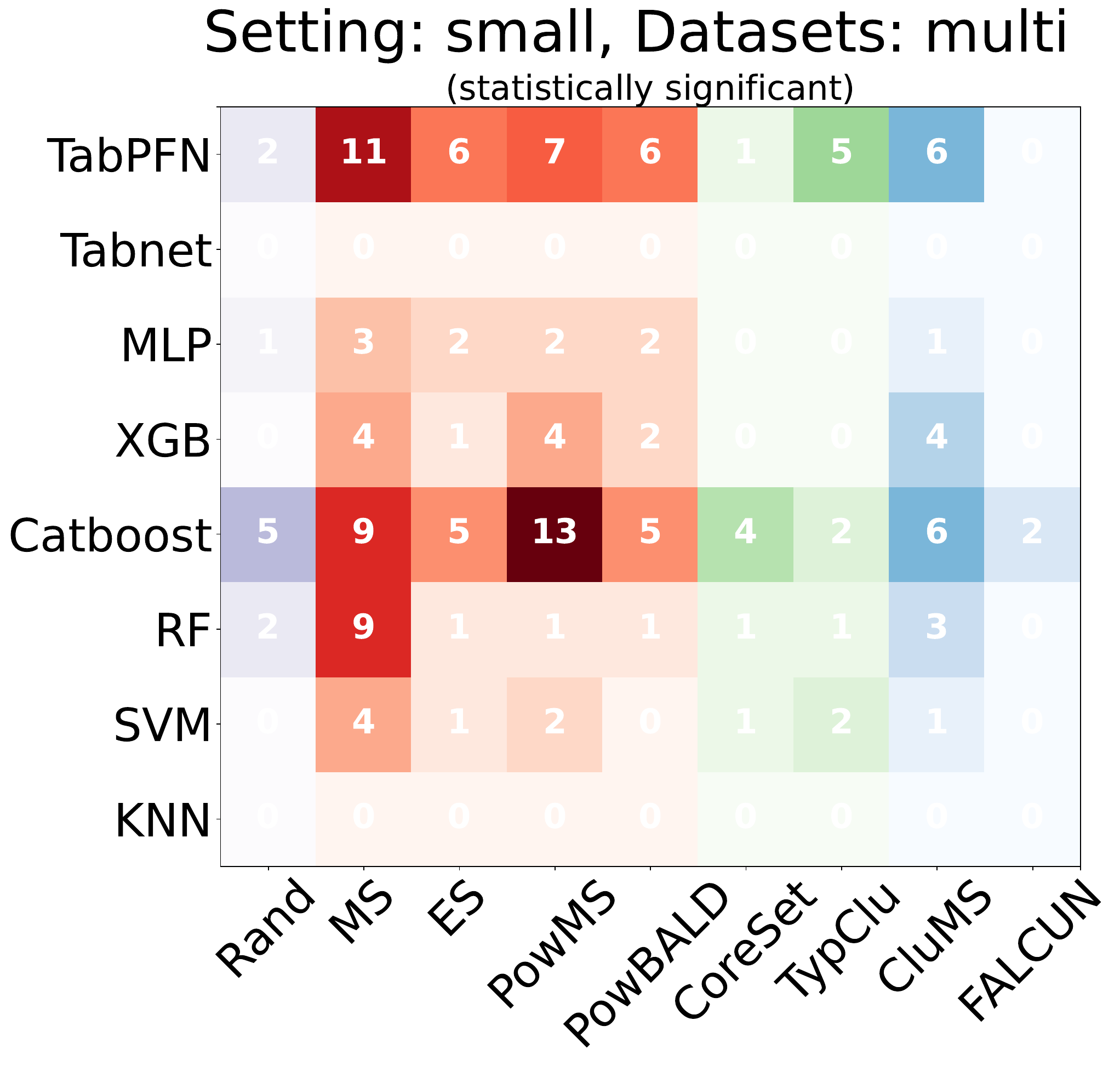}
    \end{subfigure}
    \begin{subfigure}[b]{0.24\textwidth}
        \centering
        \includegraphics[width=\textwidth]{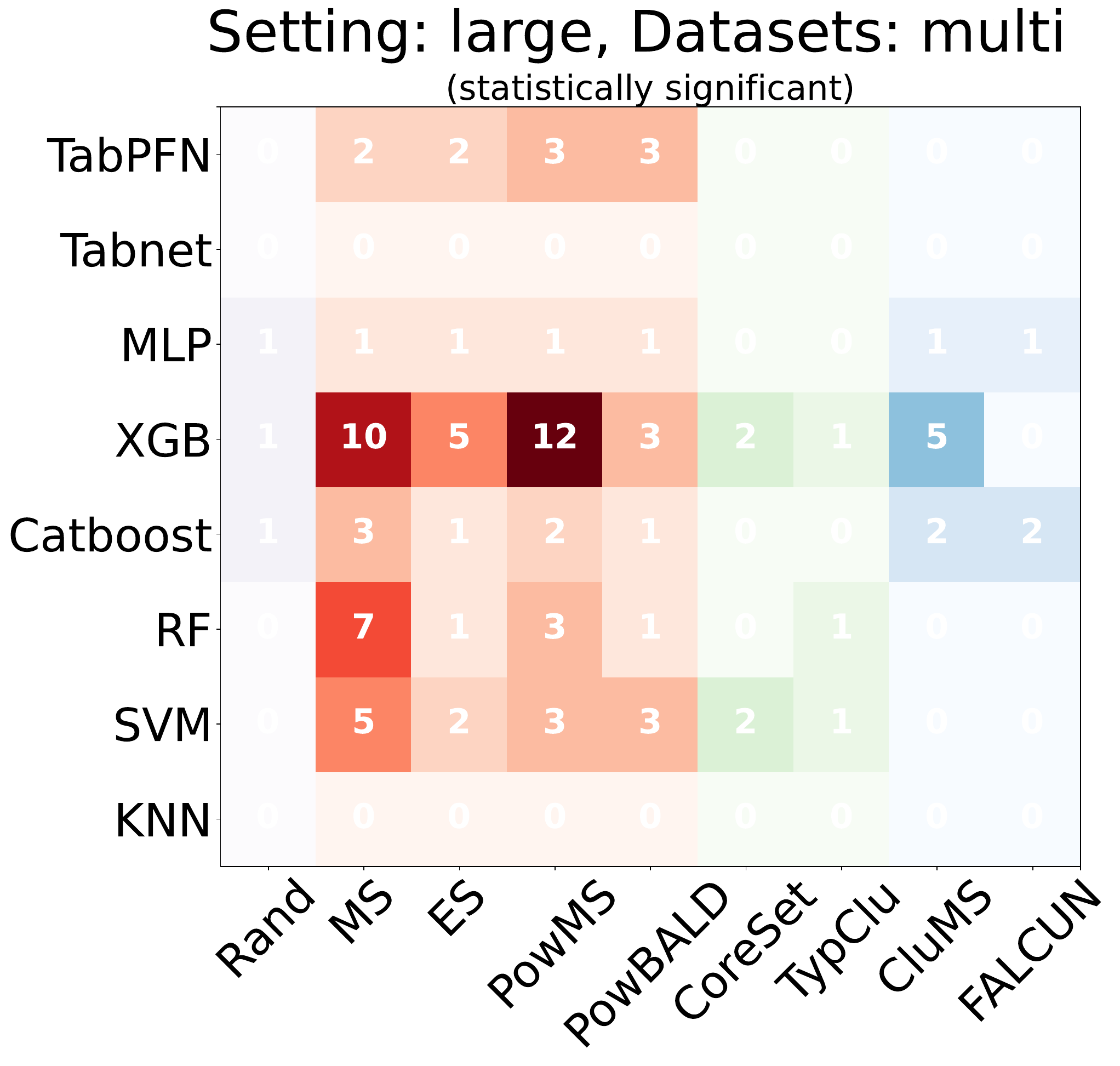}
    \end{subfigure}
    \begin{subfigure}[b]{0.24\textwidth}
        \centering
        \includegraphics[width=\textwidth]{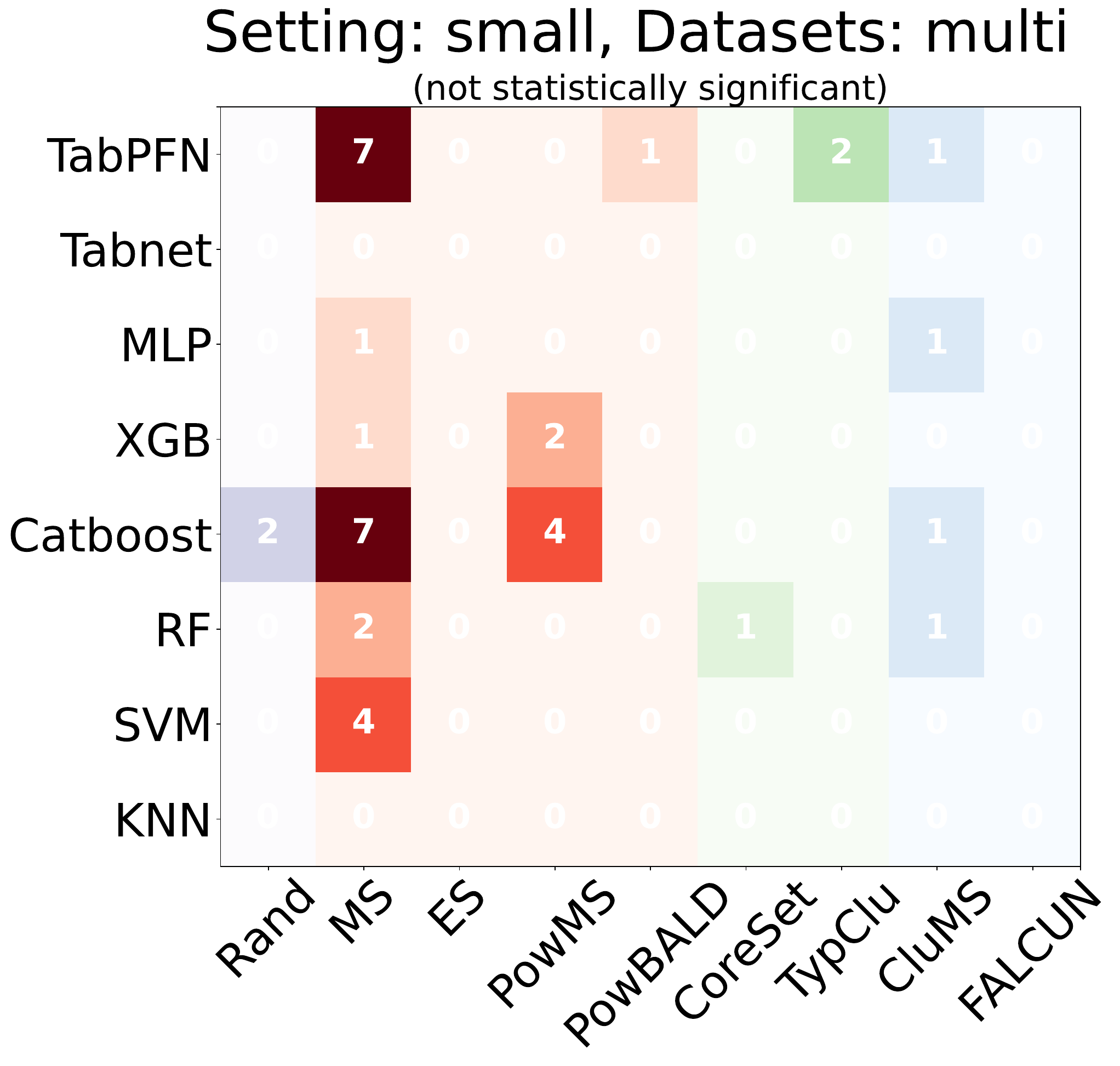}
    \end{subfigure}
    \begin{subfigure}[b]{0.24\textwidth}
        \centering
        \includegraphics[width=\textwidth]{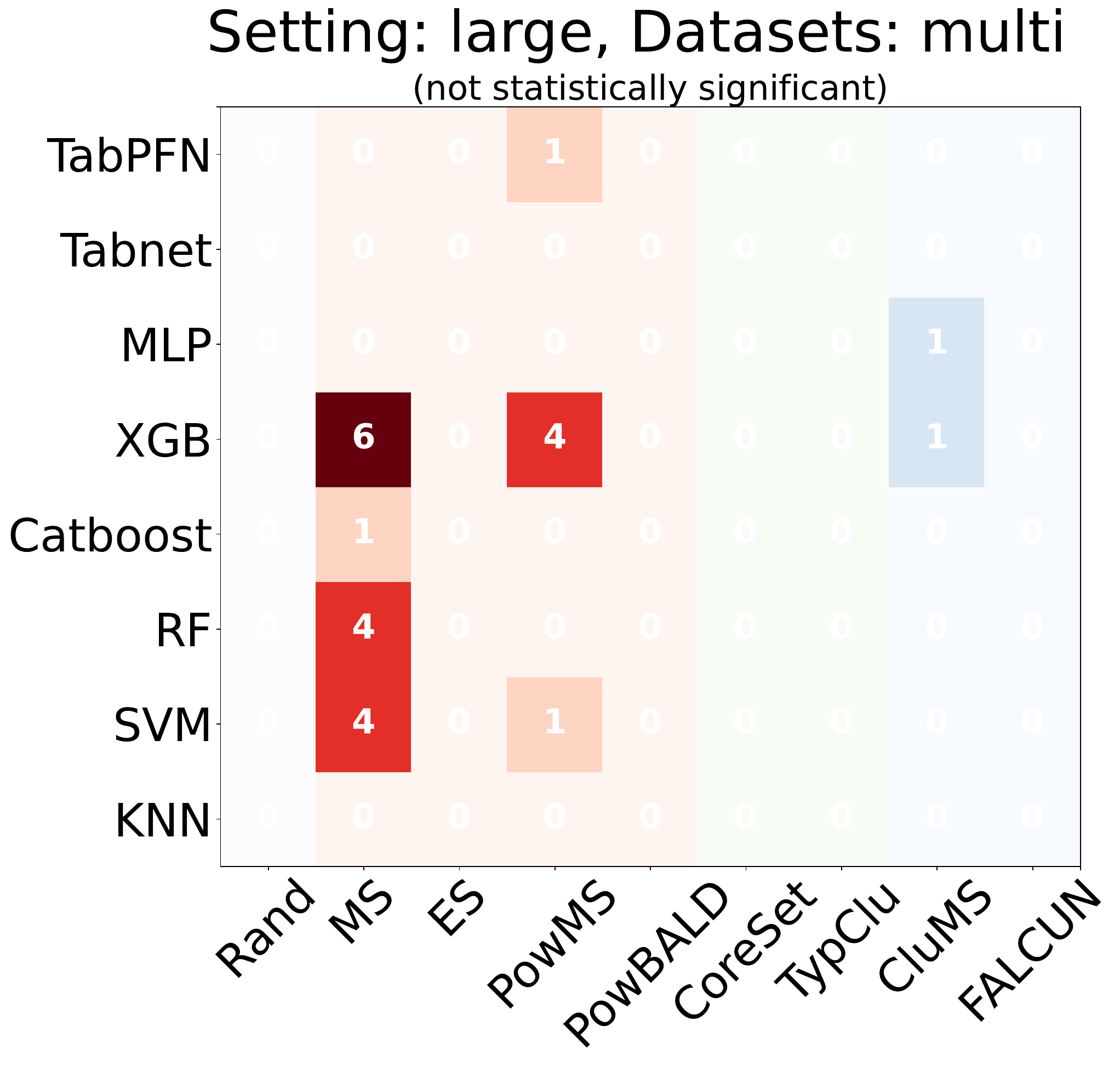}
    \end{subfigure}
    \caption{Heatmaps for all \acp{alp} within our evaluation study \textbf{with} statistical significance (first and second subfigure) and \textbf{without} (third and fourth) for \textbf{multi-class} datasets. Info.-based, repr.-based, and hybr. \acp{qs} are colored in red, green, and blue, respectively, and random sampling in purple.}
    \label{fig:heatmaps_multi}
\end{figure}

In Figure~\ref{fig:win_matrices_binary} and~\ref{fig:win_matrices_multi}, we present Win-Matrices for \ac{knn}, \ac{mlp} and \ac{xgb} considering the small setting on binary and multi-class datasets, respectively. Notably, FALCUN shows better performance with \ac{knn} and \ac{mlp} on binary datasets compared to multi-class datasets. Additionally, \acs{es} demonstrates superior performance on binary datasets when paired with \ac{xgb} as the learning algorithm.

\begin{figure}[h]
    \centering
    \begin{subfigure}[b]{0.32\textwidth}
        \centering
        \includegraphics[width=\textwidth]{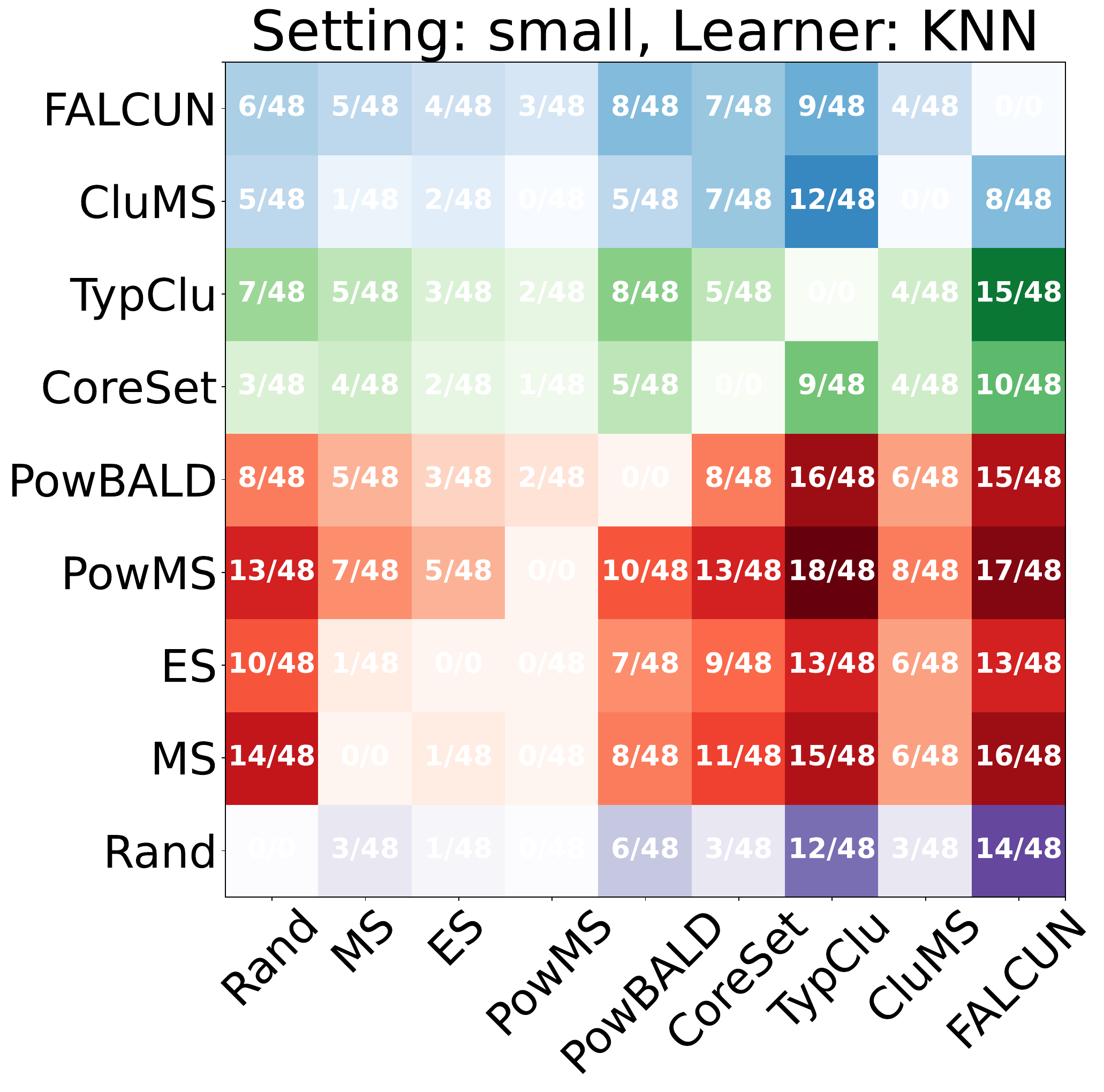}
    \end{subfigure}
    \begin{subfigure}[b]{0.32\textwidth}
        \centering
        \includegraphics[width=\textwidth]{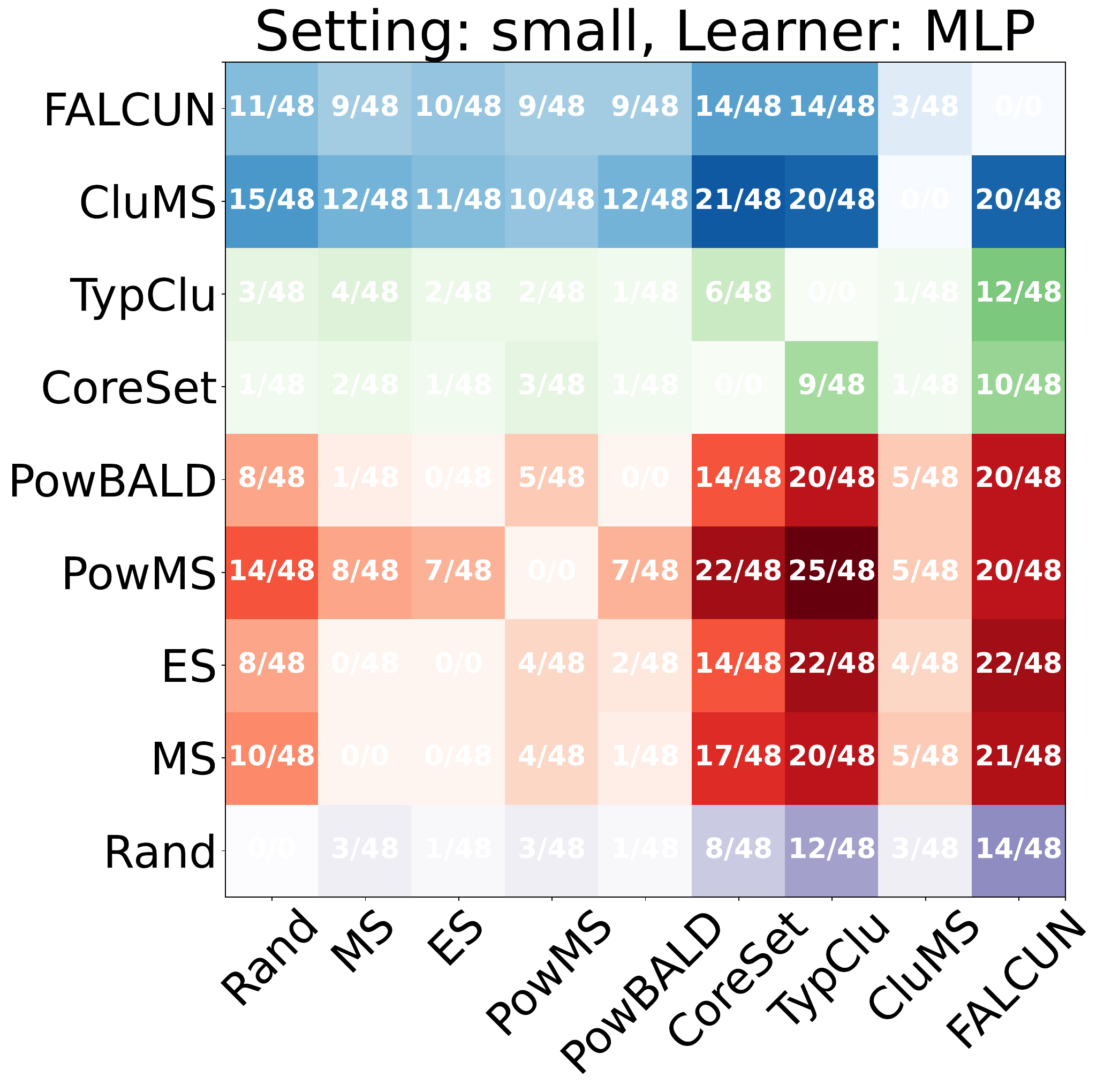}
    \end{subfigure}
        \begin{subfigure}[b]{0.32\textwidth}
        \centering
        \includegraphics[width=\textwidth]{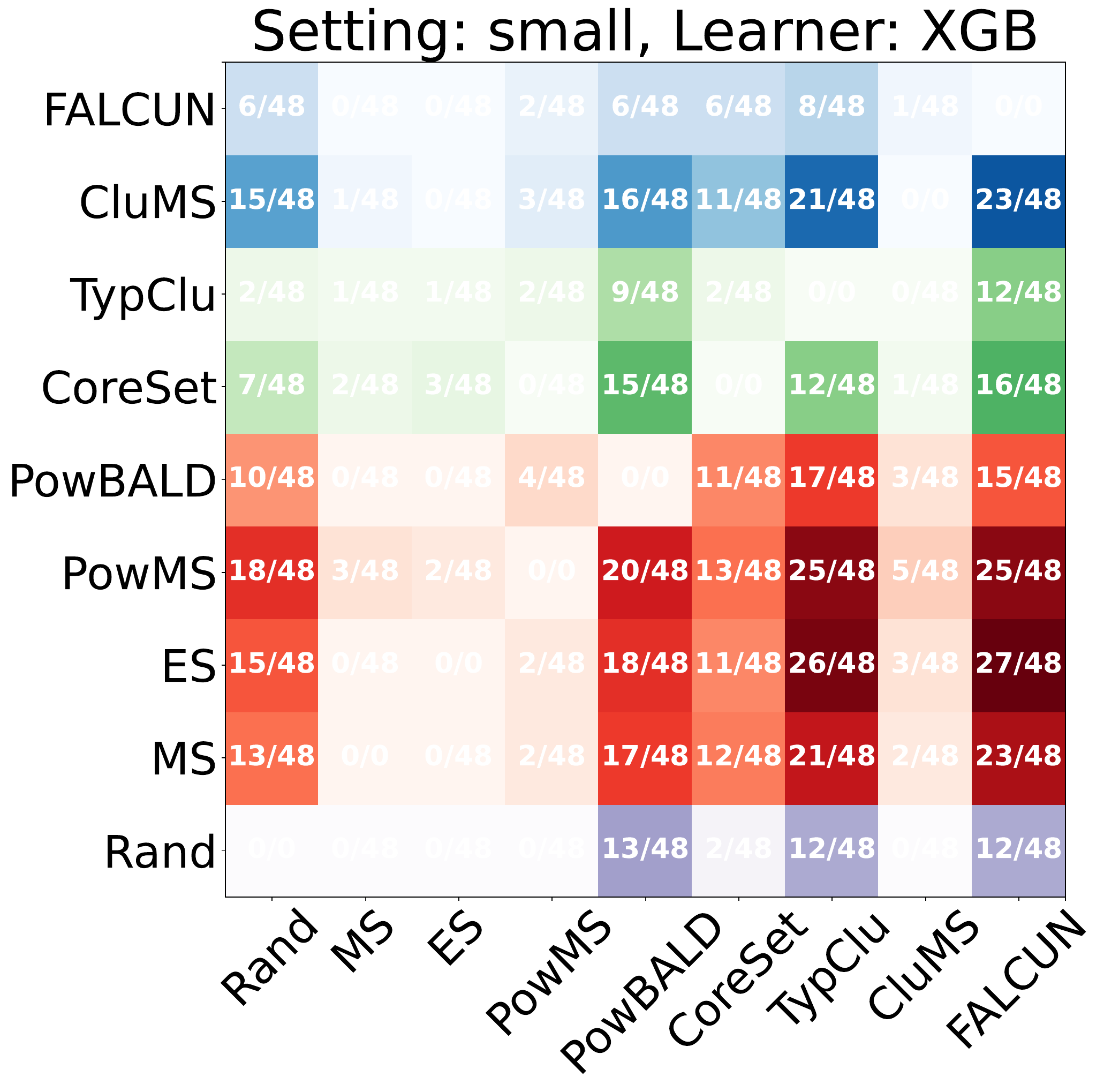}
    \end{subfigure}
    \caption{Win-Matrices for \ac{knn}, \ac{mlp} and \ac{xgb} considering \textbf{binary} dataset and the \textbf{small} setting.}
    \label{fig:win_matrices_binary}
\end{figure}

\begin{figure}[h]
    \centering
    \begin{subfigure}[b]{0.32\textwidth}
        \centering
        \includegraphics[width=\textwidth]{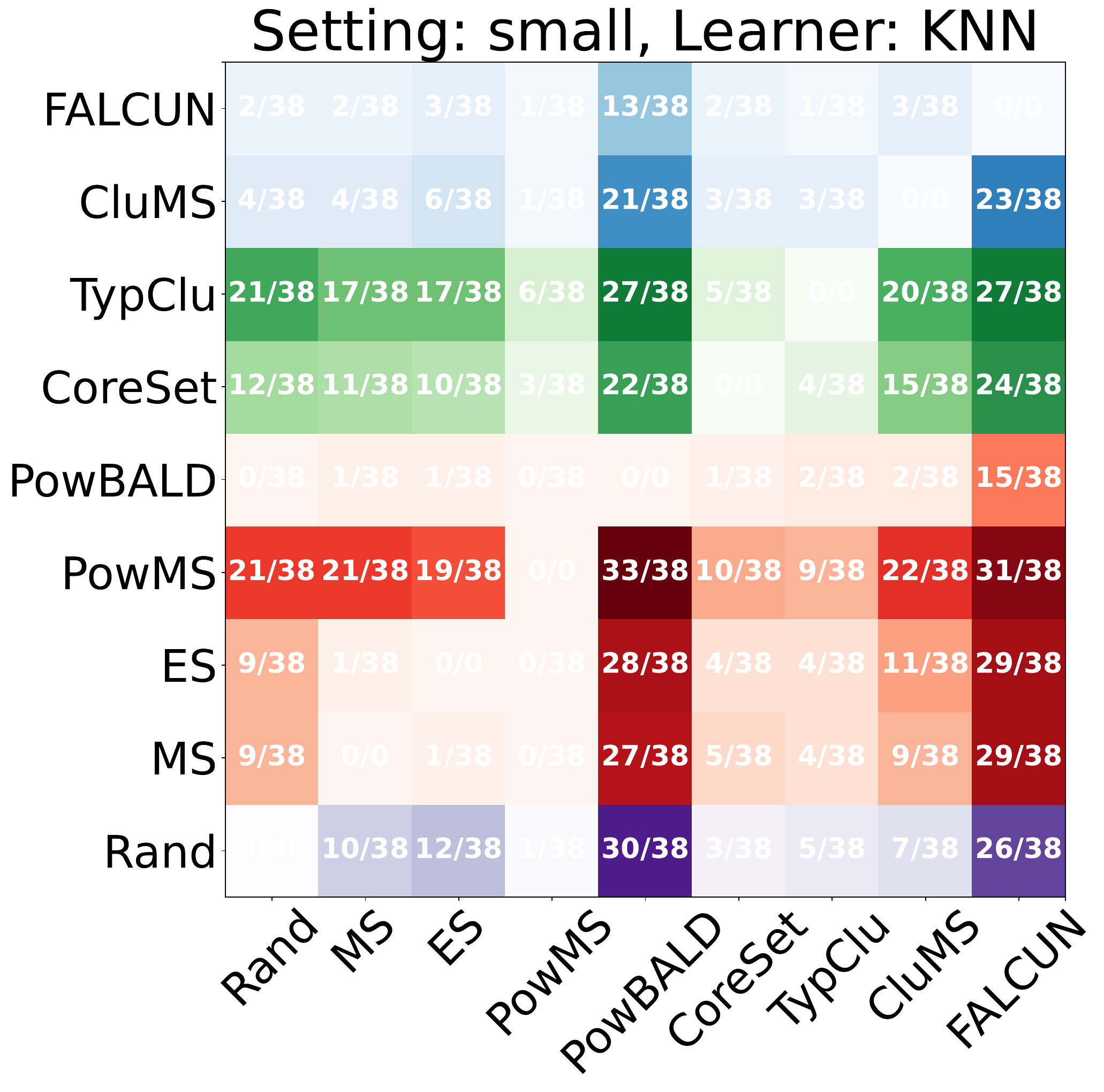}
    \end{subfigure}
    \begin{subfigure}[b]{0.32\textwidth}
        \centering
        \includegraphics[width=\textwidth]{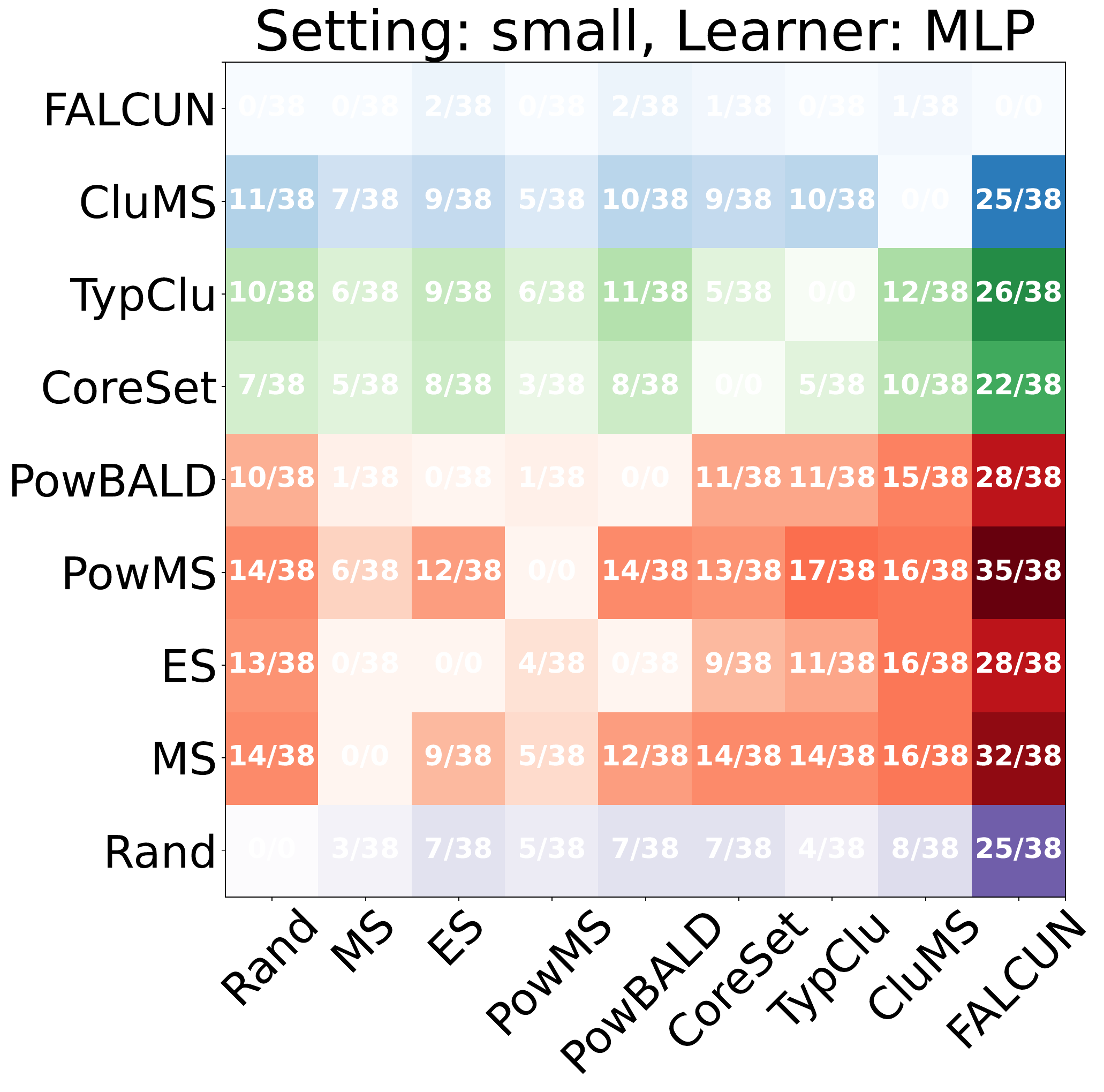}
    \end{subfigure}
        \begin{subfigure}[b]{0.32\textwidth}
        \centering
        \includegraphics[width=\textwidth]{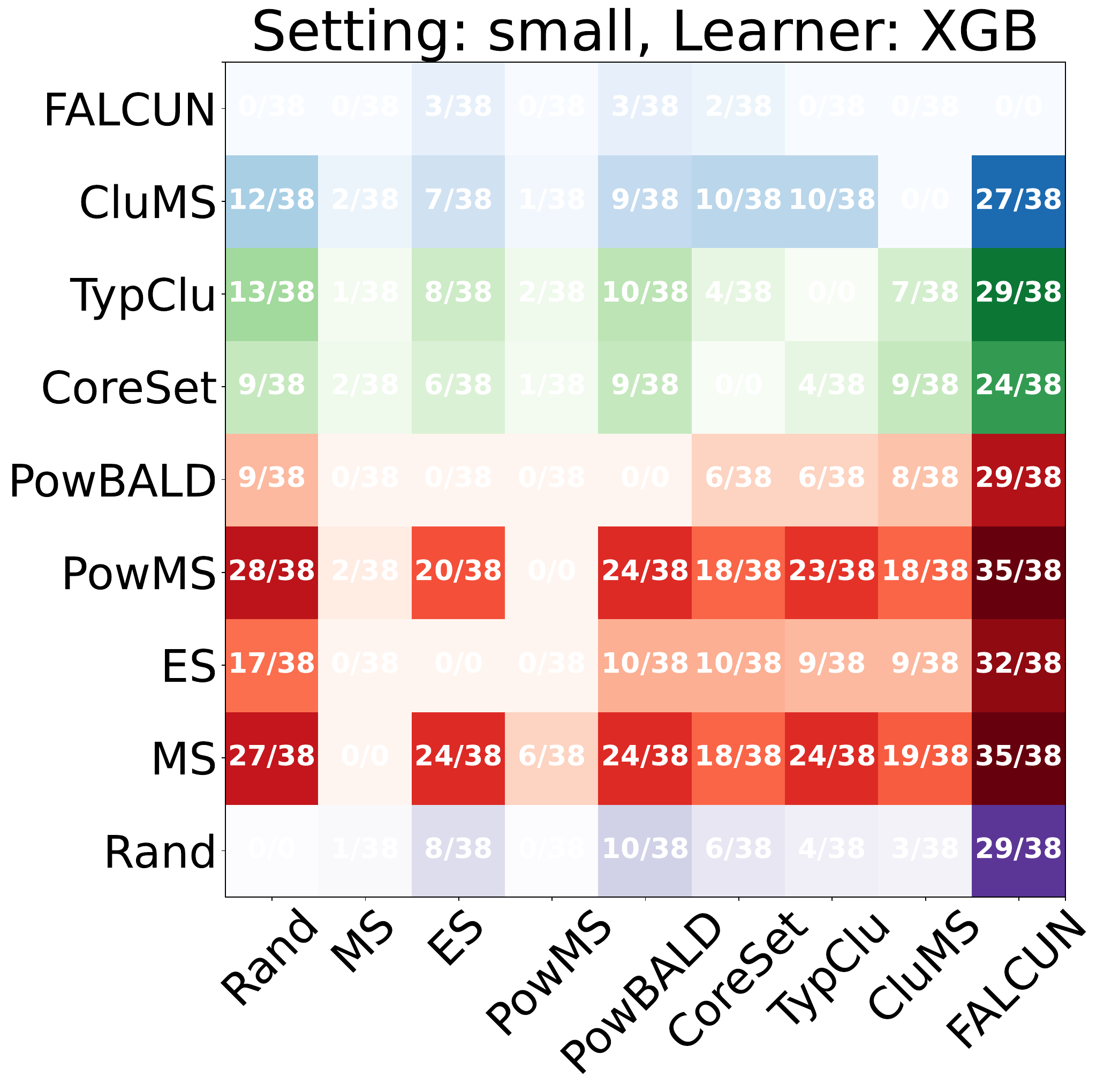}
    \end{subfigure}
    \caption{Win-Matrices for \ac{knn}, \ac{mlp} and \ac{xgb} considering \textbf{multi-class} dataset and the \textbf{small} setting.}
    \label{fig:win_matrices_multi}
\end{figure}

We present budget curves for \ac{svm}, \ac{rf}, \ac{xgb} and TabPFN in Figures~\ref{fig:1464} to~\ref{fig:25} and for all learners excluding TabPPFN in Figures~\ref{fig:46} and~\ref{fig:182}. As illustrated in Figures~\ref{fig:1464} to~\ref{fig:25}, the choice of the learning algorithm is crucial to the overall performance. For example, on the dataset with OpenML ID 334, all pipelines using TabPFN as the learner achieve performance at least as good as the best pipeline using \ac{xgb}. Another notable observation is the performance of TabNet, as shown in the budget curves in Figures~\ref{fig:46} and~\ref{fig:182}. The budget curve resembles a step function, suggesting that TabNet requires a minimum amount of labeled data to achieve reasonable performance.

\begin{figure}[t]
    \centering
    \begin{subfigure}[b]{0.49\textwidth}
        \centering
        \includegraphics[width=\textwidth]{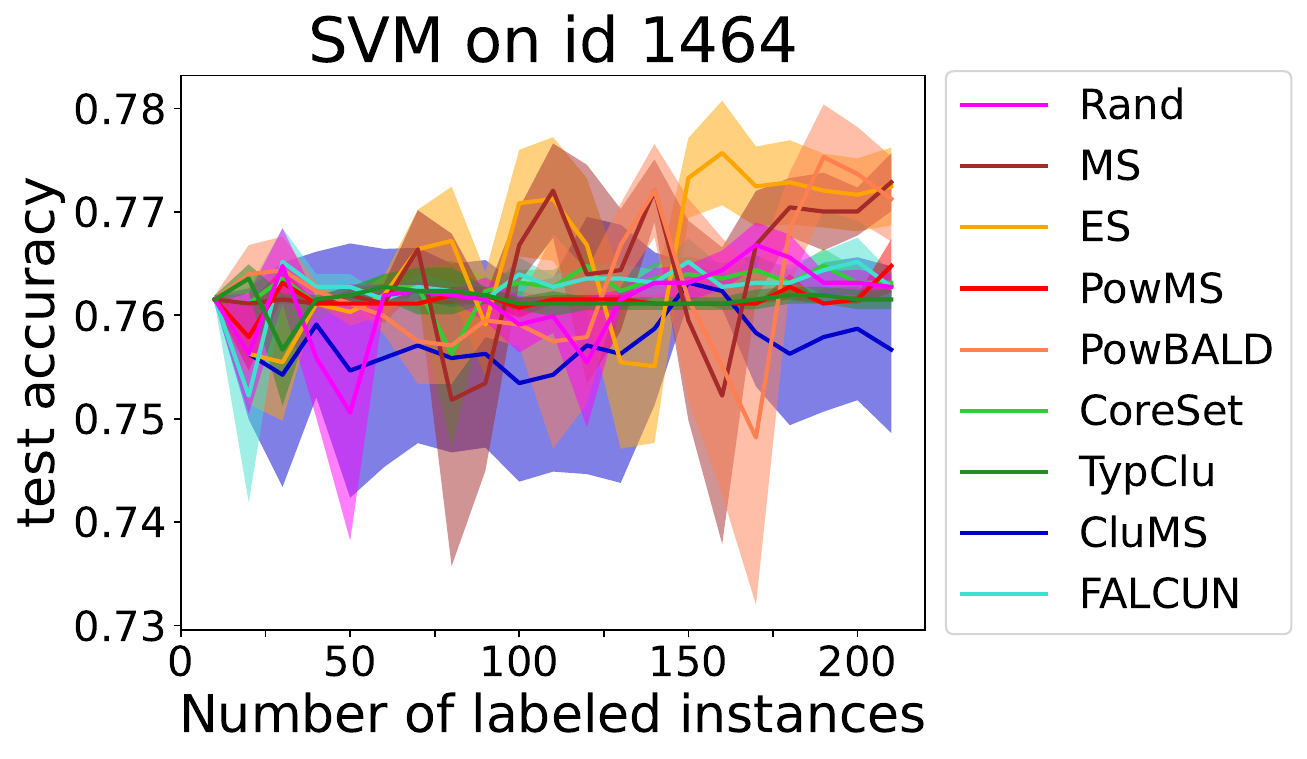}
    \end{subfigure}
    \begin{subfigure}[b]{0.49\textwidth}
        \centering
        \includegraphics[width=\textwidth]{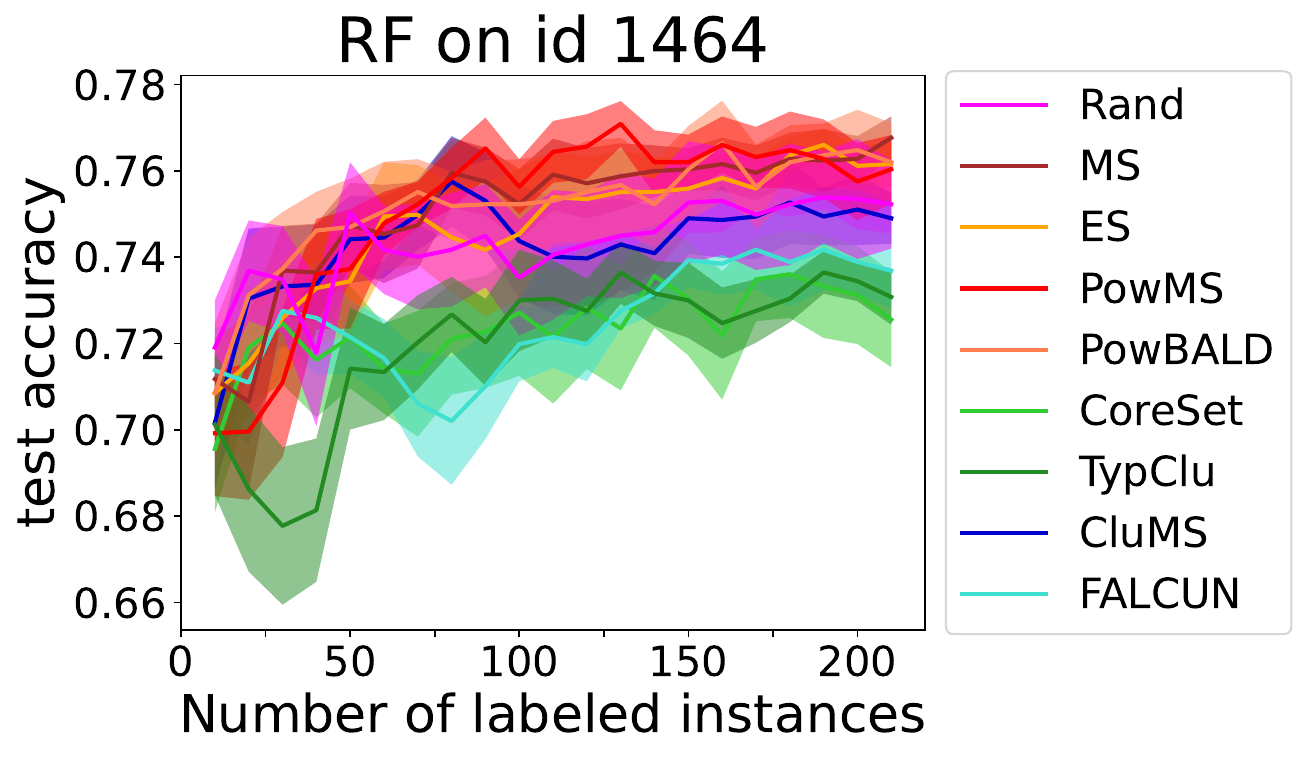}
    \end{subfigure}
        \begin{subfigure}[b]{0.49\textwidth}
        \centering
        \includegraphics[width=\textwidth]{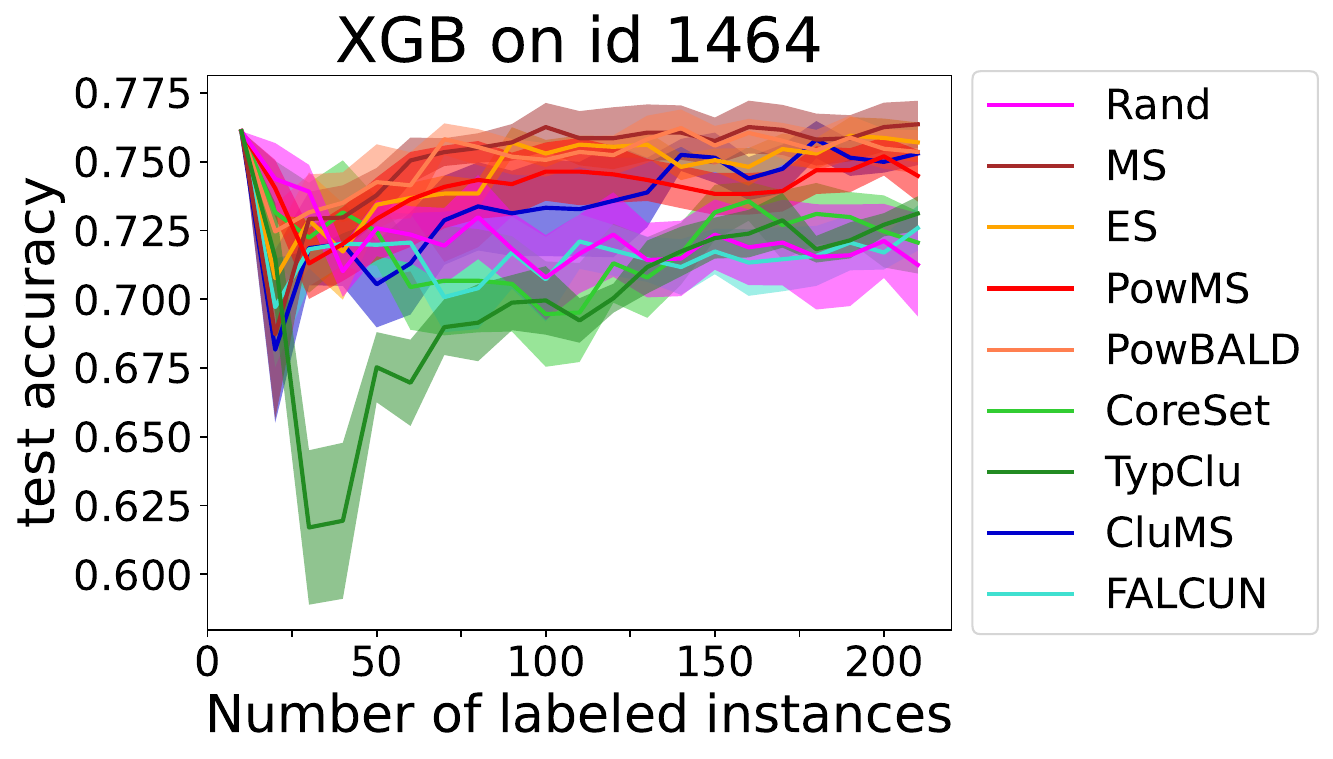}
    \end{subfigure}
        \begin{subfigure}[b]{0.49\textwidth}
        \centering
        \includegraphics[width=\textwidth]{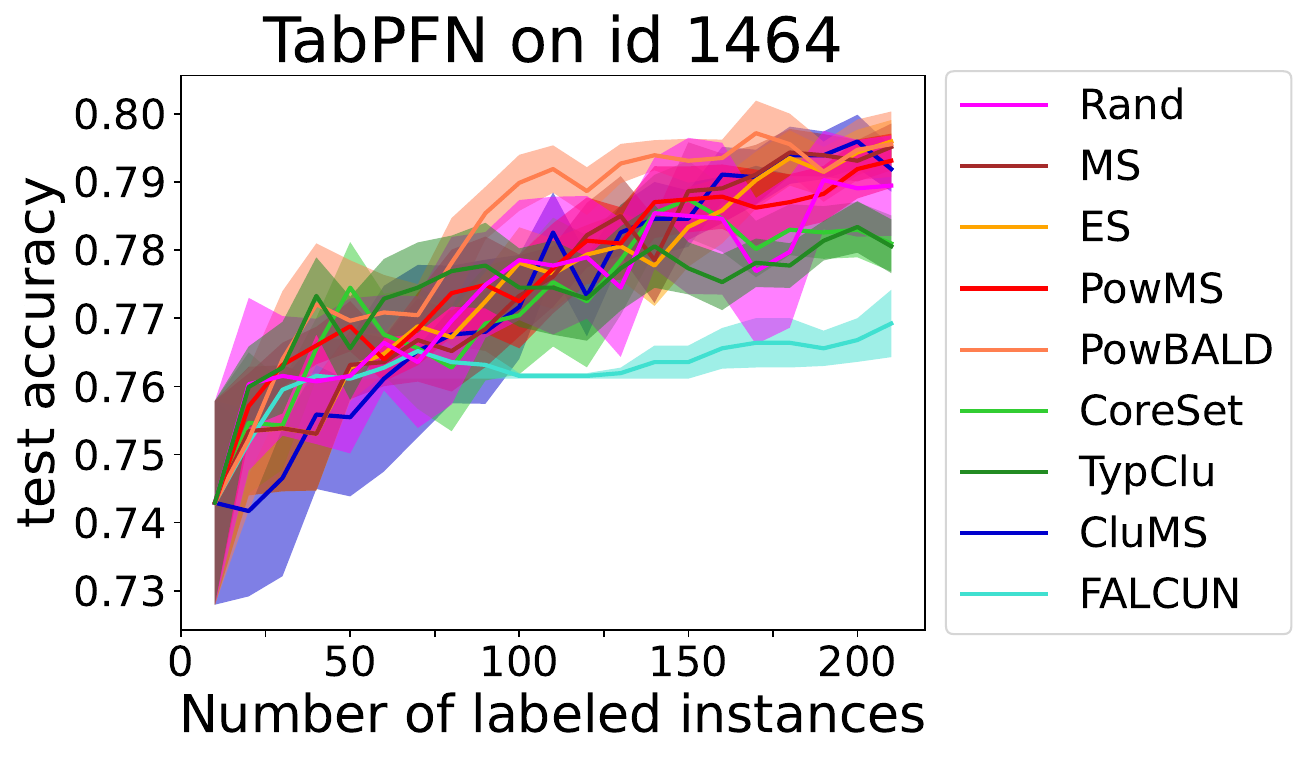}
    \end{subfigure}
    \caption{Budget curves for different \acp{alp} on the dataset with OpenML ID 1464, considering the \textbf{small} setting.}
    \label{fig:1464}
\end{figure}

\begin{figure}[t]
    \centering
    \begin{subfigure}[b]{0.49\textwidth}
        \centering
        \includegraphics[width=\textwidth]{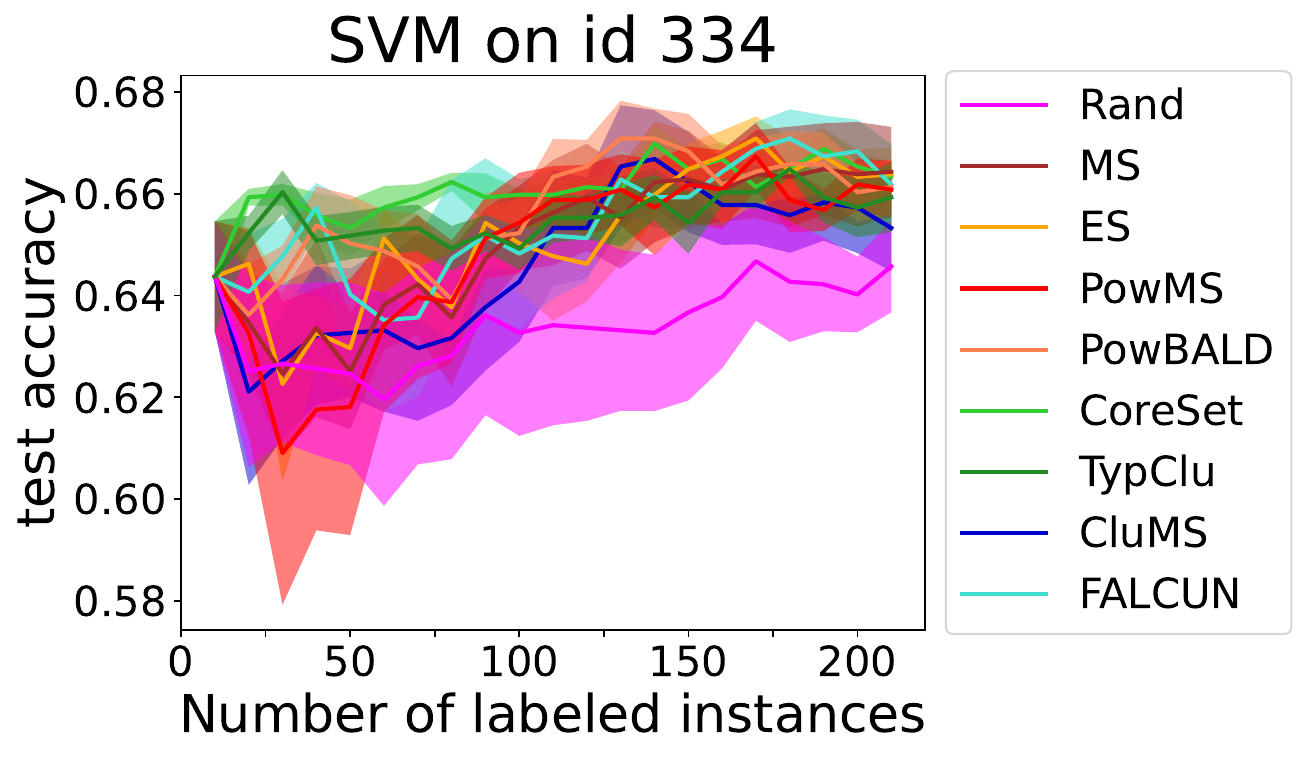}
    \end{subfigure}
    \begin{subfigure}[b]{0.49\textwidth}
        \centering
        \includegraphics[width=\textwidth]{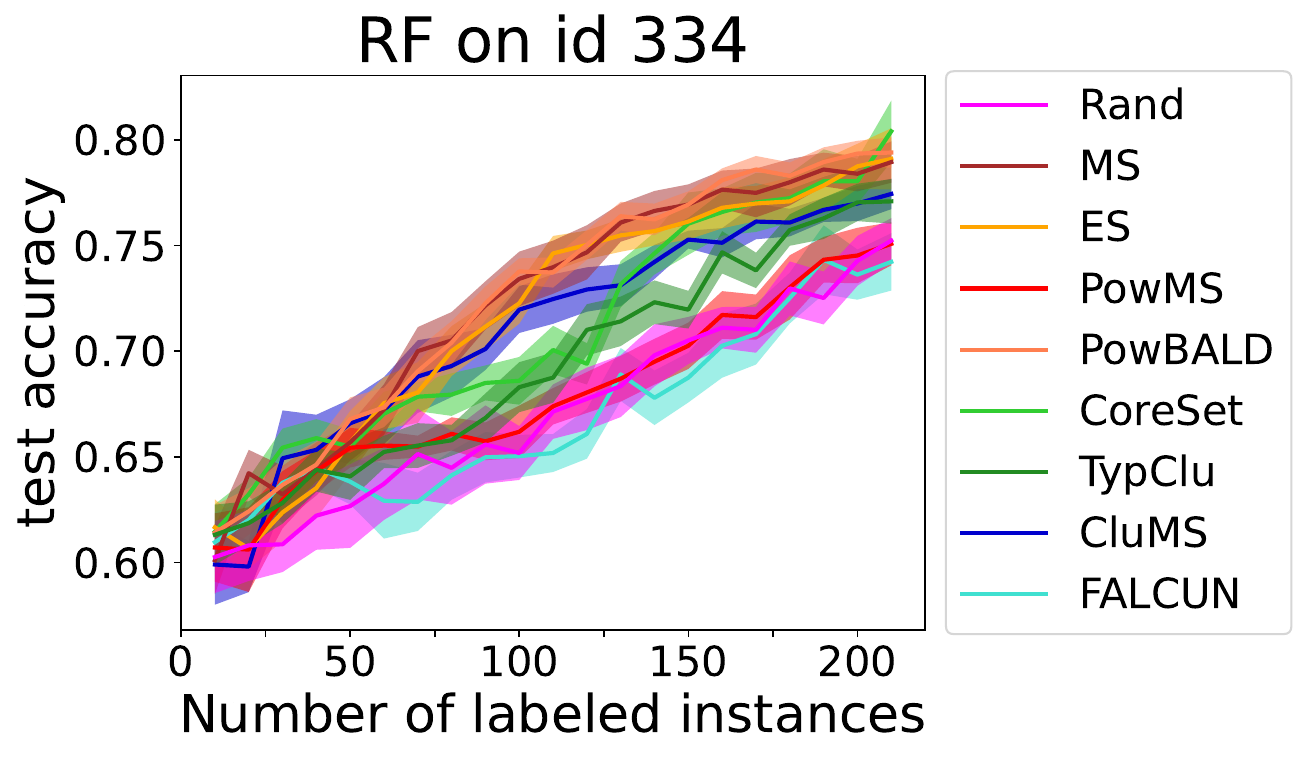}
    \end{subfigure}
        \begin{subfigure}[b]{0.49\textwidth}
        \centering
        \includegraphics[width=\textwidth]{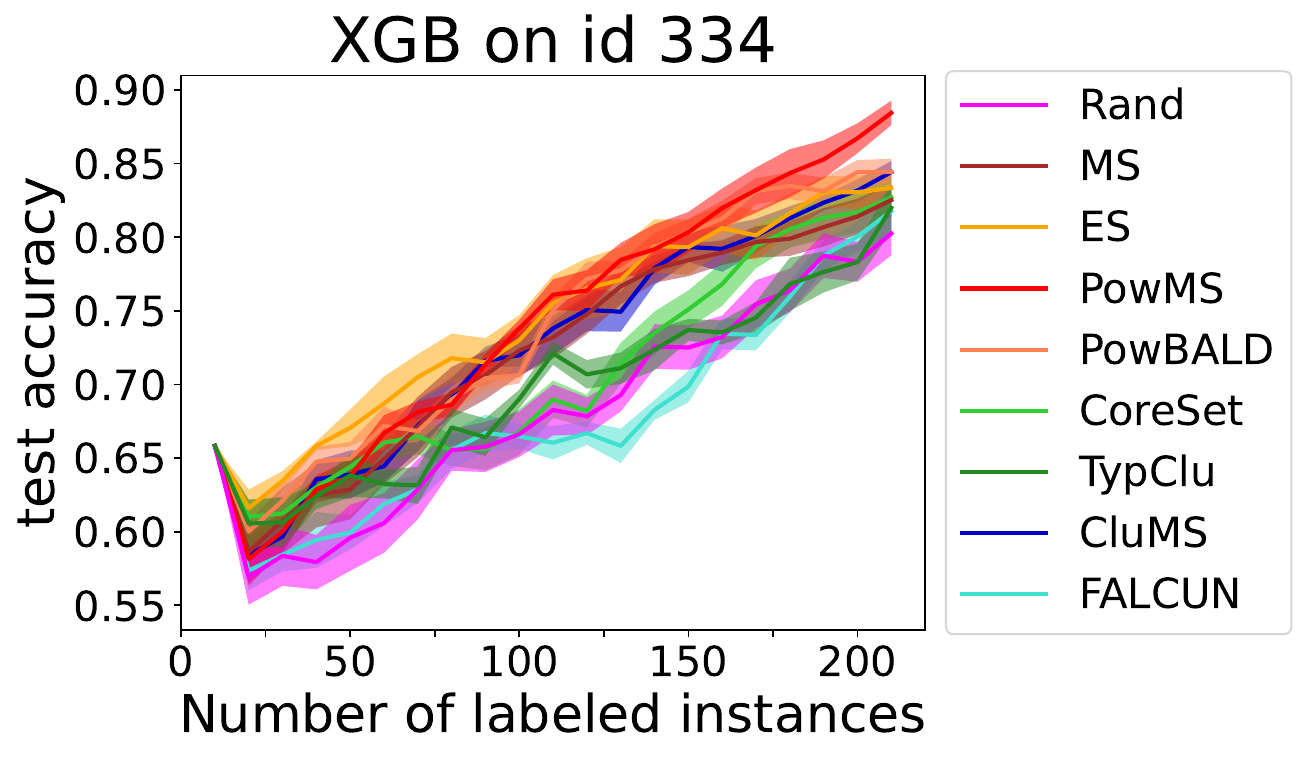}
    \end{subfigure}
        \begin{subfigure}[b]{0.49\textwidth}
        \centering
        \includegraphics[width=\textwidth]{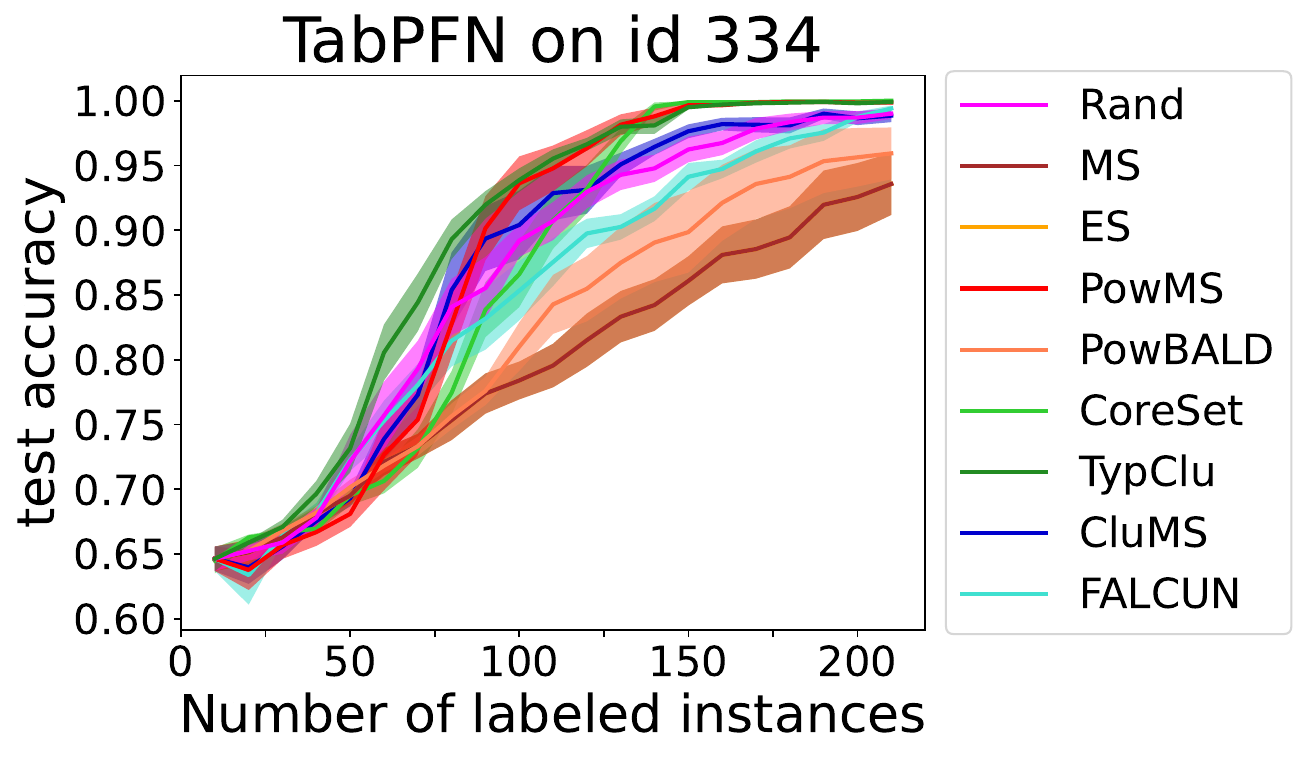}
    \end{subfigure}
    \caption{Budget curves for different \acp{alp} on dataset with OpenML ID 334, considering the \textbf{small} setting.}
    \label{fig:334}
\end{figure}

\begin{figure}[t]
    \centering
    \begin{subfigure}[b]{0.49\textwidth}
        \centering
        \includegraphics[width=\textwidth]{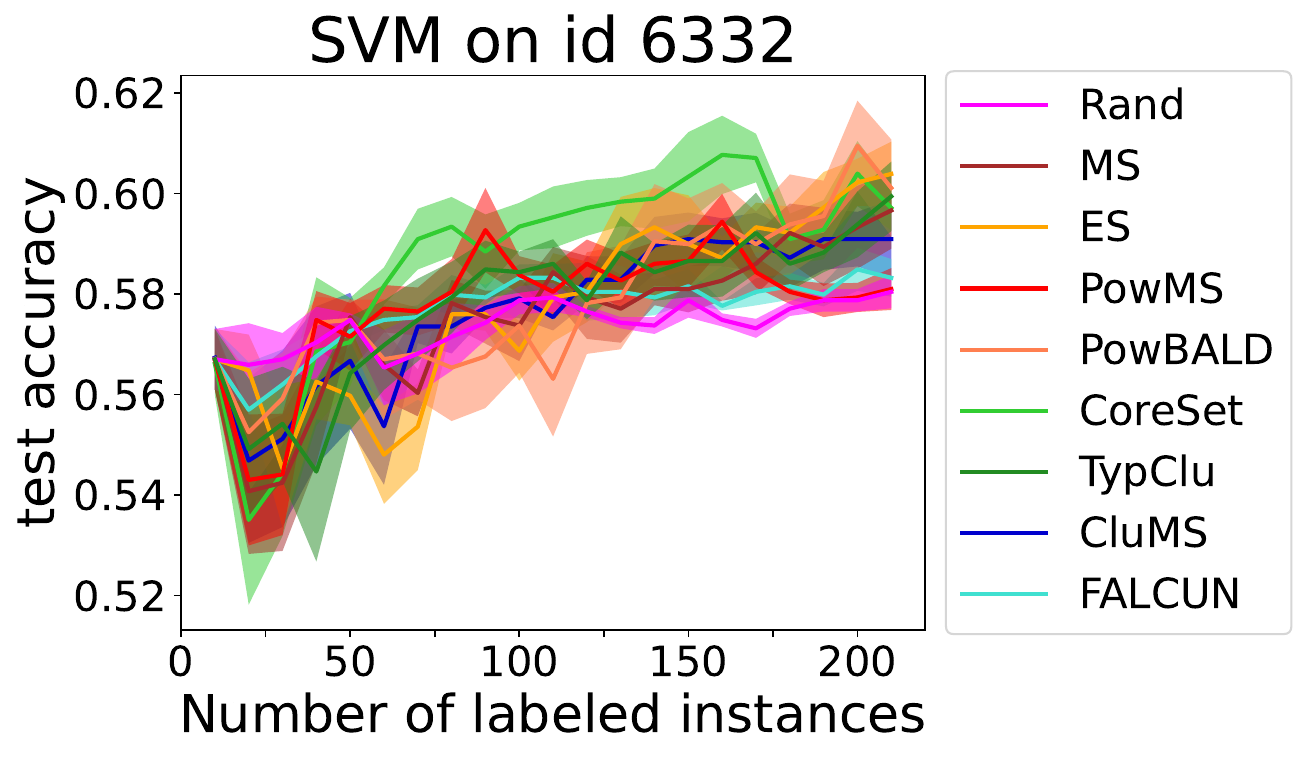}
    \end{subfigure}
    \begin{subfigure}[b]{0.49\textwidth}
        \centering
        \includegraphics[width=\textwidth]{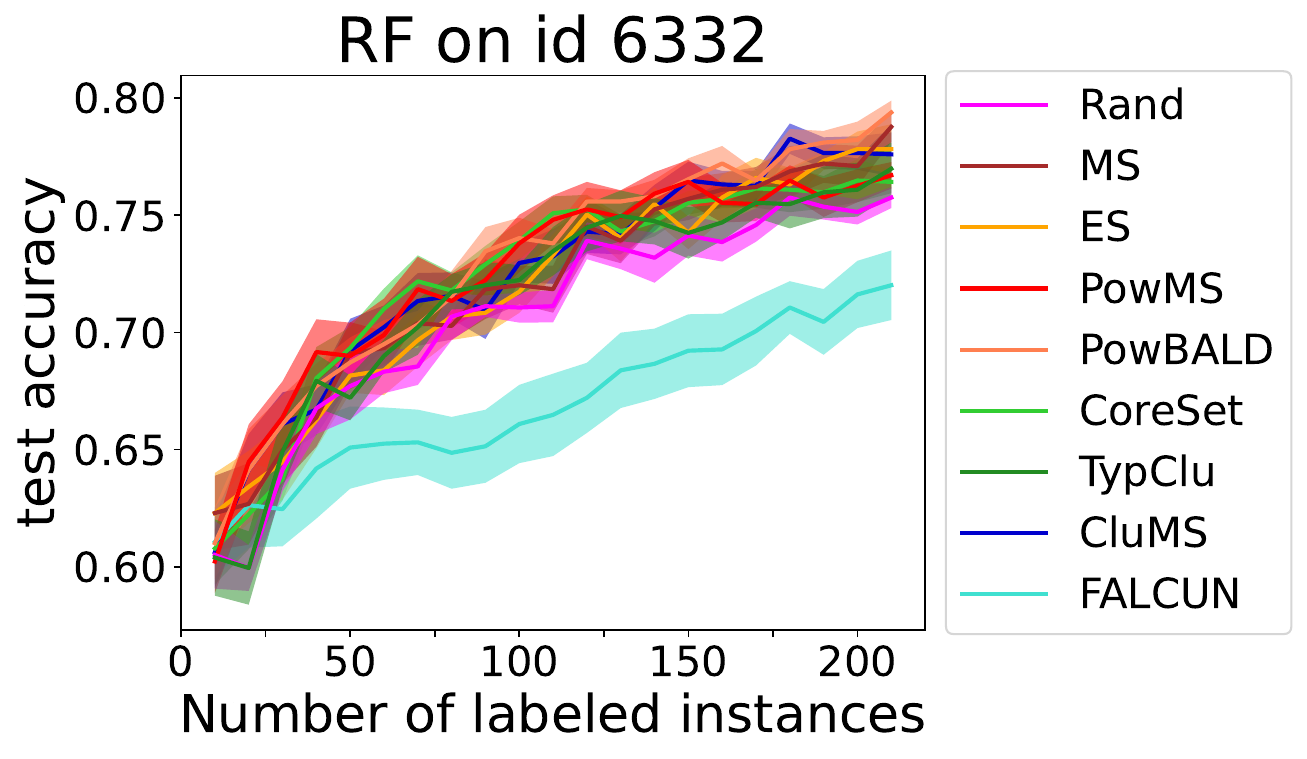}
    \end{subfigure}
        \begin{subfigure}[b]{0.49\textwidth}
        \centering
        \includegraphics[width=\textwidth]{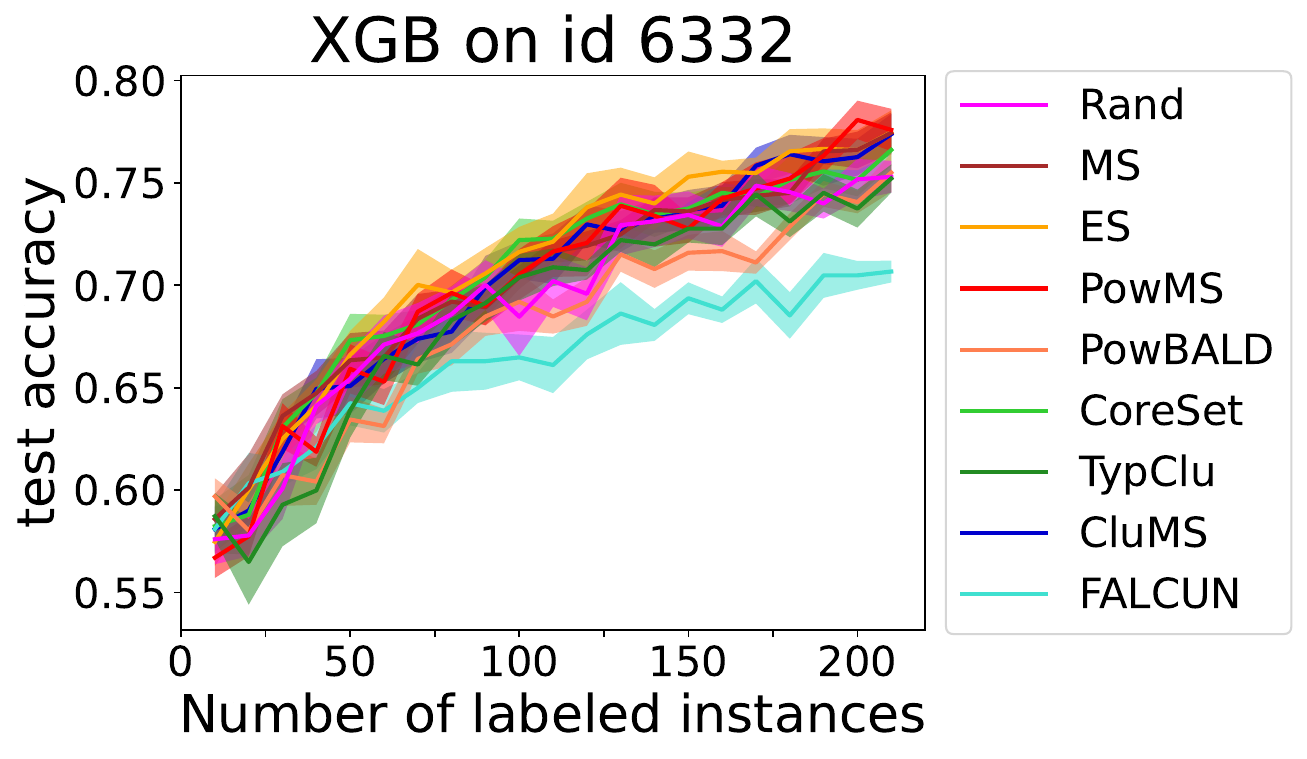}
    \end{subfigure}
        \begin{subfigure}[b]{0.49\textwidth}
        \centering
        \includegraphics[width=\textwidth]{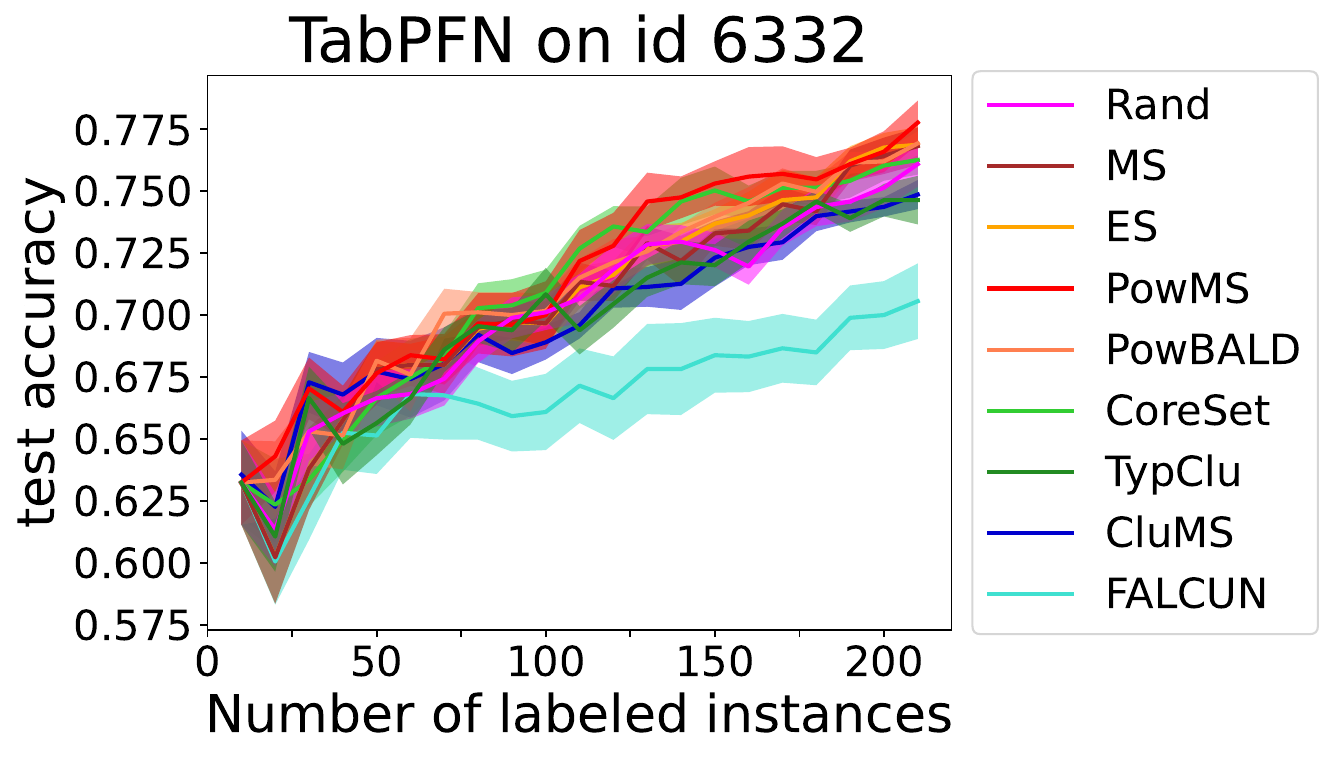}
    \end{subfigure}
    \caption{Budget curves for different \acp{alp} on the dataset with OpenML ID 6332, considering the \textbf{small} setting.}
    \label{fig:6332}
\end{figure}

\begin{figure}[t]
    \centering
    \begin{subfigure}[b]{0.49\textwidth}
        \centering
        \includegraphics[width=\textwidth]{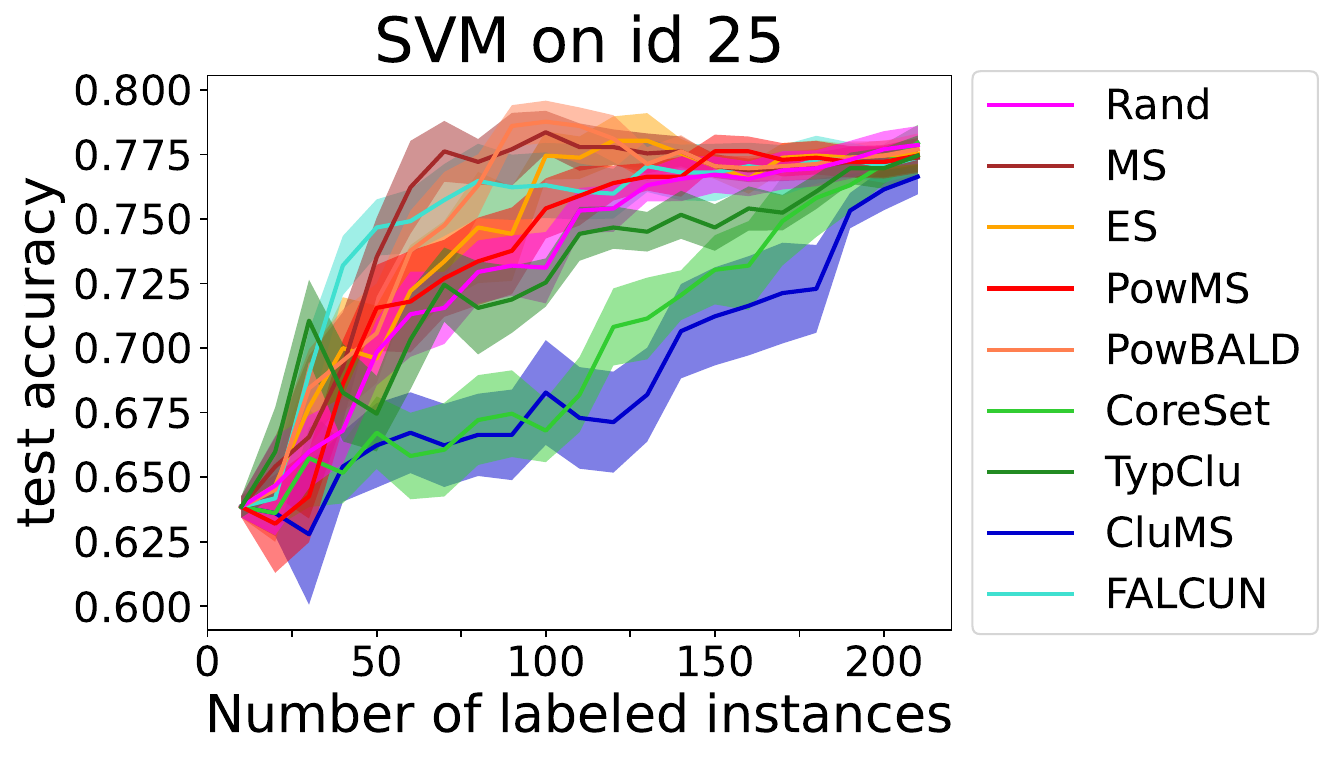}
    \end{subfigure}
    \begin{subfigure}[b]{0.49\textwidth}
        \centering
        \includegraphics[width=\textwidth]{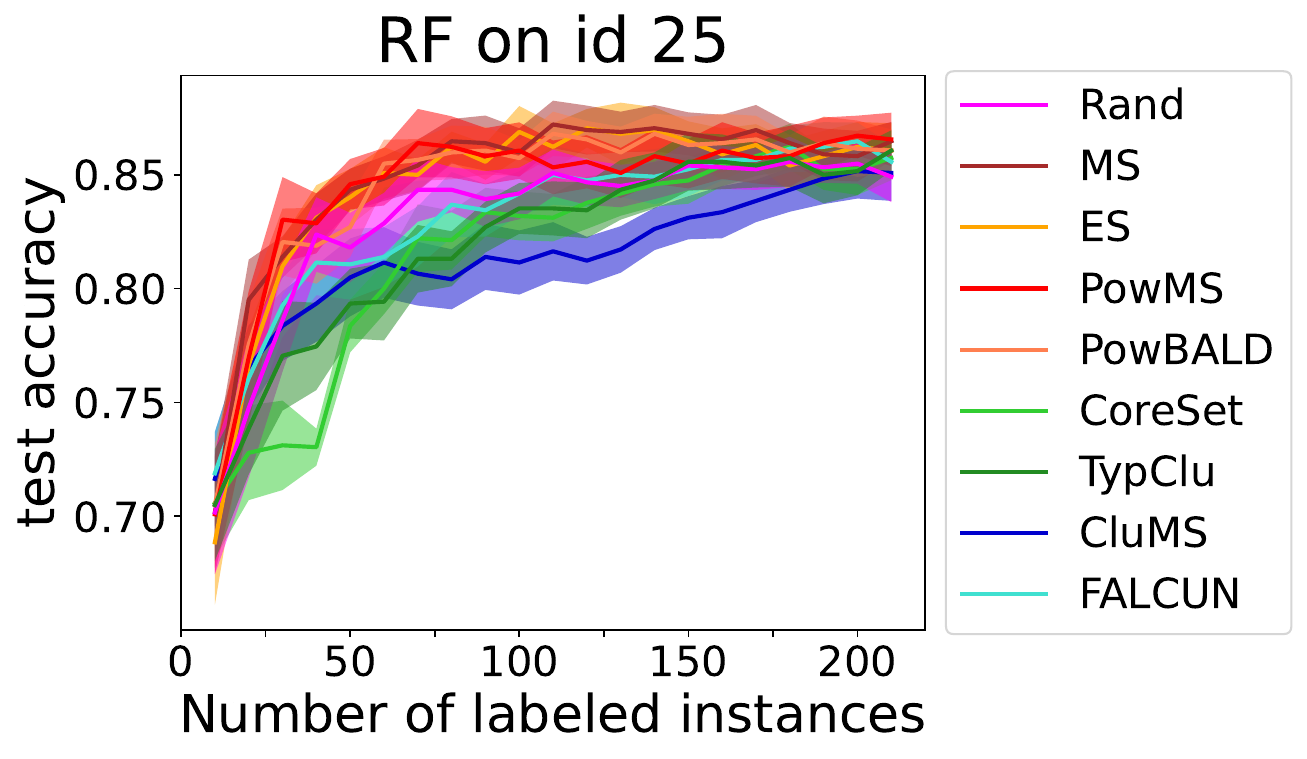}
    \end{subfigure}
        \begin{subfigure}[b]{0.49\textwidth}
        \centering
        \includegraphics[width=\textwidth]{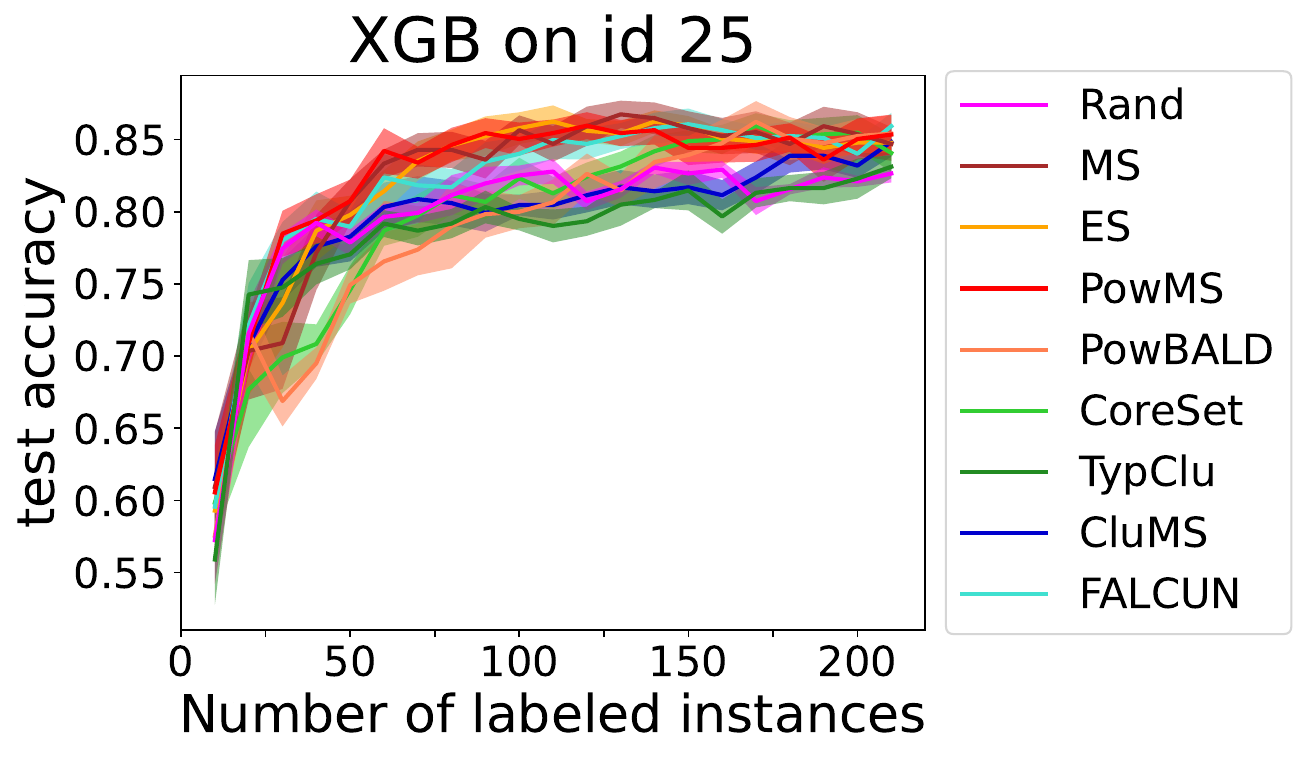}
    \end{subfigure}
        \begin{subfigure}[b]{0.49\textwidth}
        \centering
        \includegraphics[width=\textwidth]{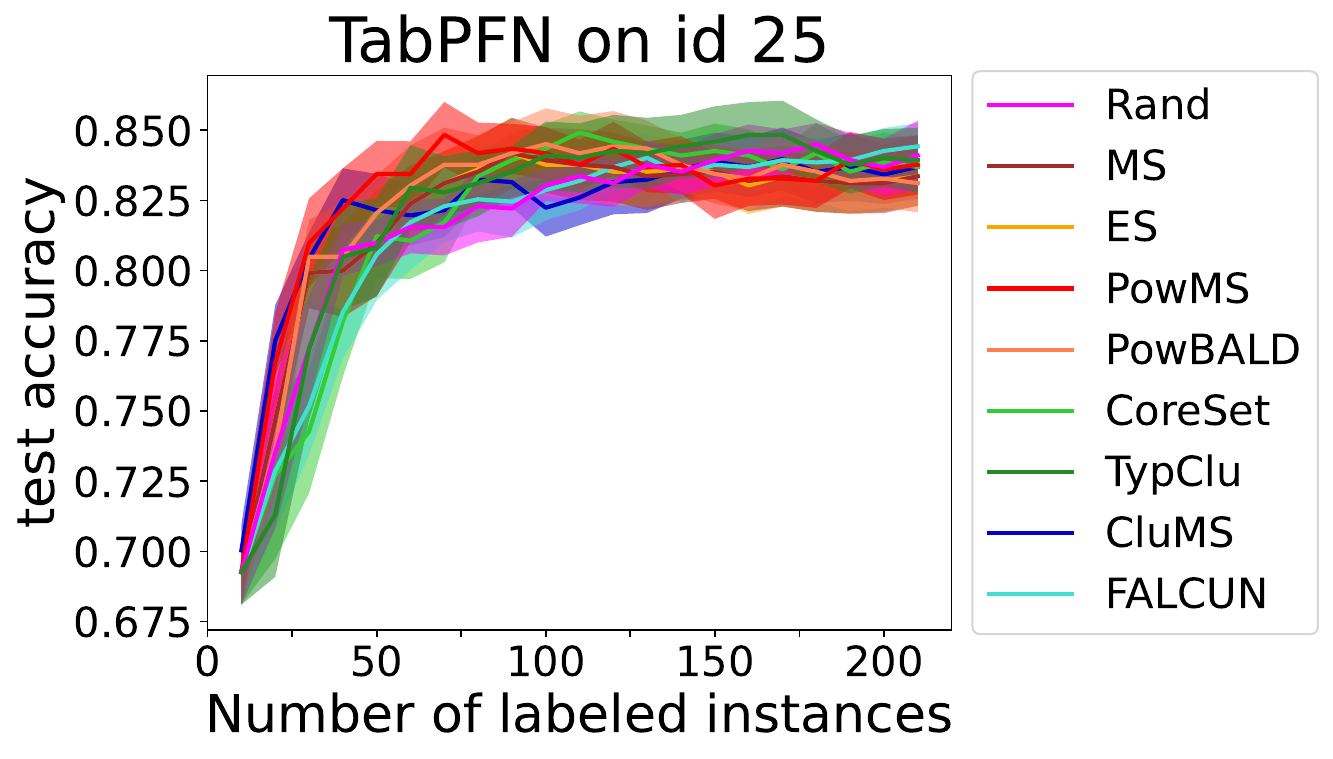}
    \end{subfigure}
    \caption{Budget curves for different \acp{alp} on the dataset with OpenML ID 25, considering the \textbf{small} setting.}
    \label{fig:25}
\end{figure}

\begin{figure}[t]
    \centering
    \begin{subfigure}[b]{0.49\textwidth}
        \centering
        \includegraphics[width=\textwidth]{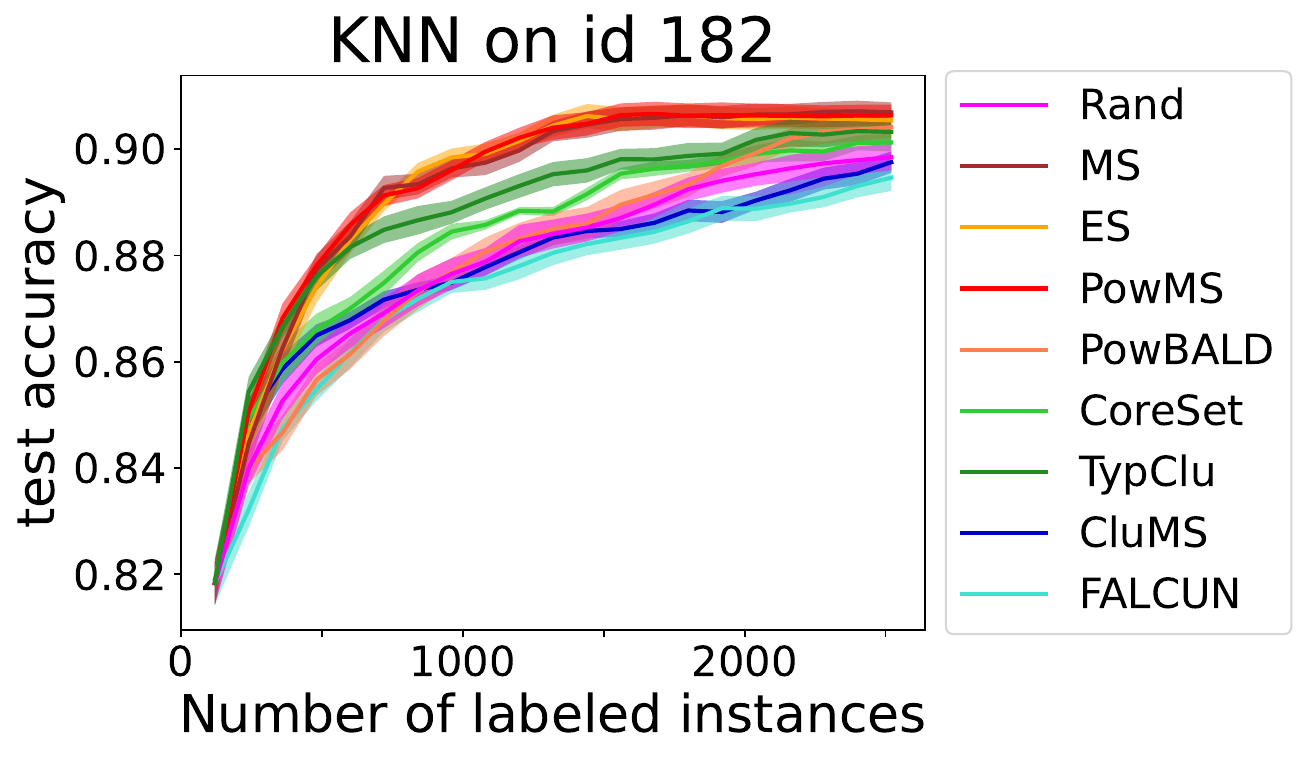}
    \end{subfigure}
    \begin{subfigure}[b]{0.49\textwidth}
        \centering
        \includegraphics[width=\textwidth]{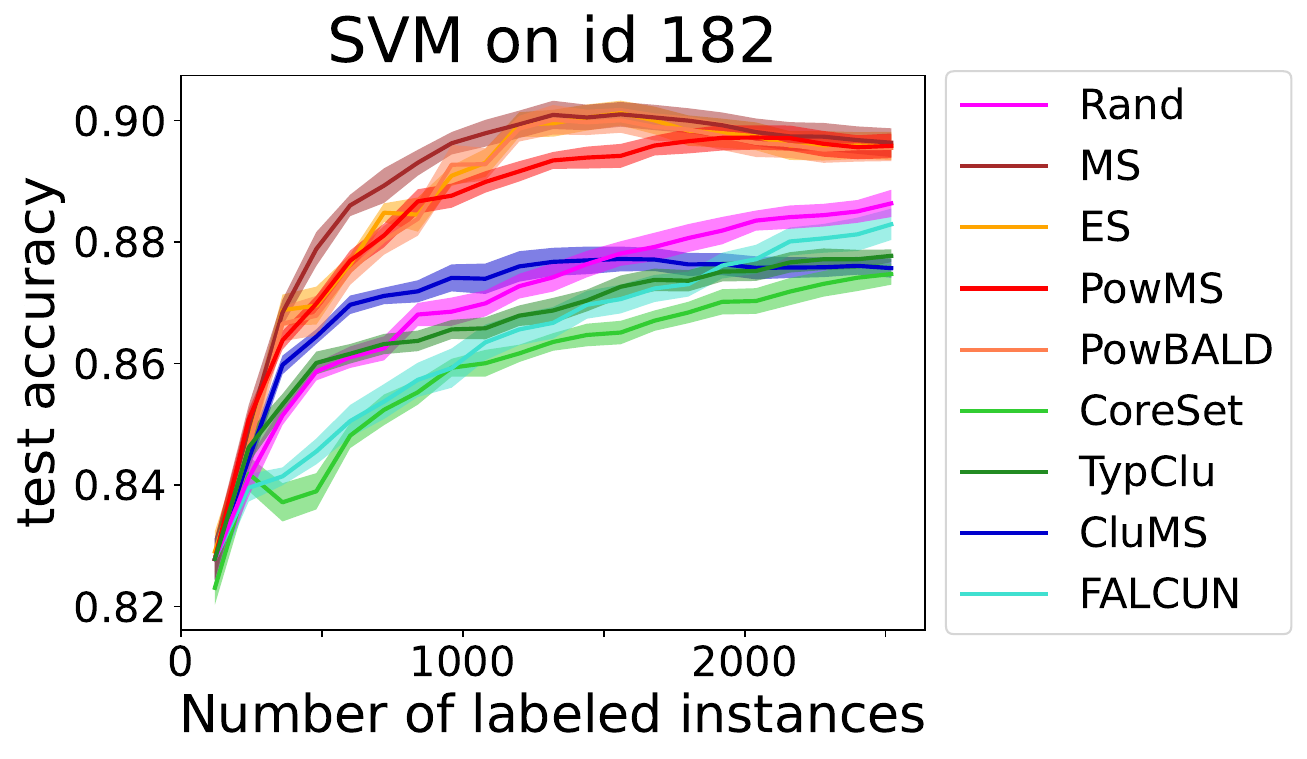}
    \end{subfigure}
        \begin{subfigure}[b]{0.49\textwidth}
        \centering
        \includegraphics[width=\textwidth]{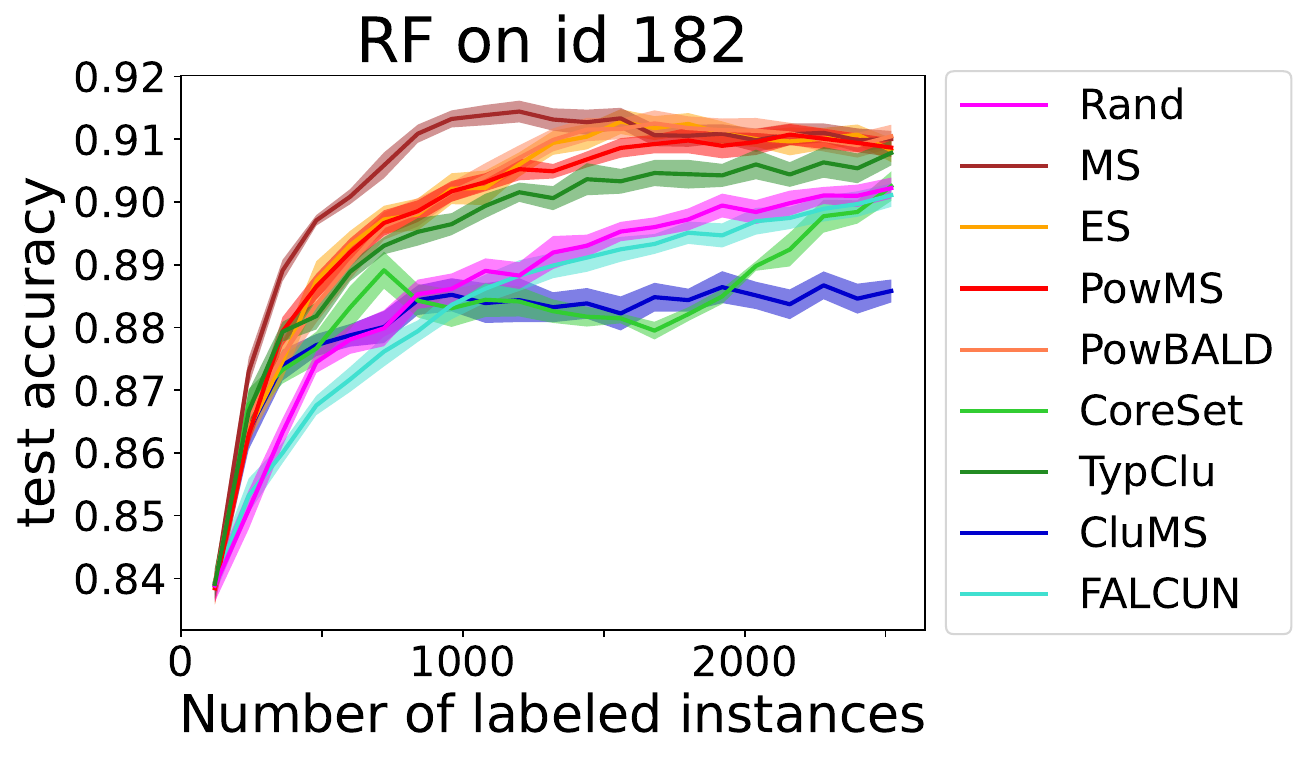}
    \end{subfigure}
        \begin{subfigure}[b]{0.49\textwidth}
        \centering
        \includegraphics[width=\textwidth]{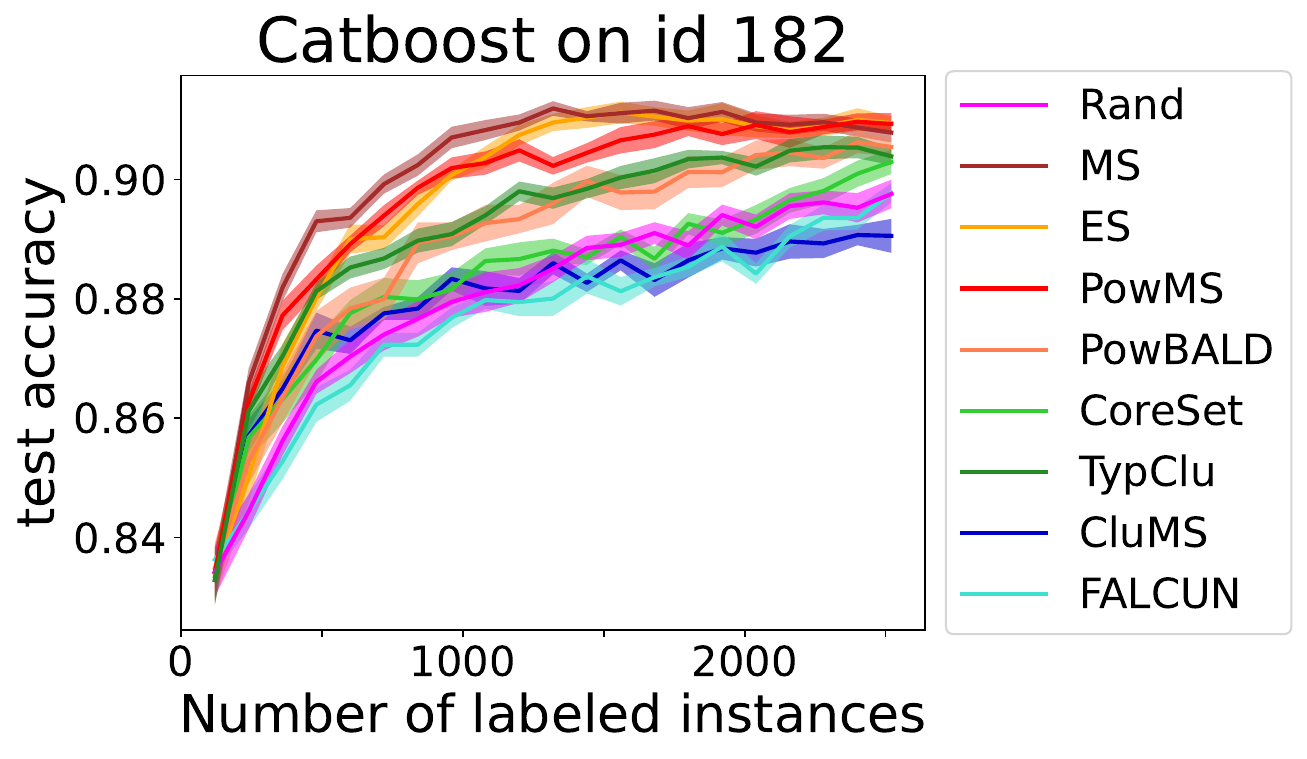}
    \end{subfigure}
        \begin{subfigure}[b]{0.49\textwidth}
        \centering
        \includegraphics[width=\textwidth]{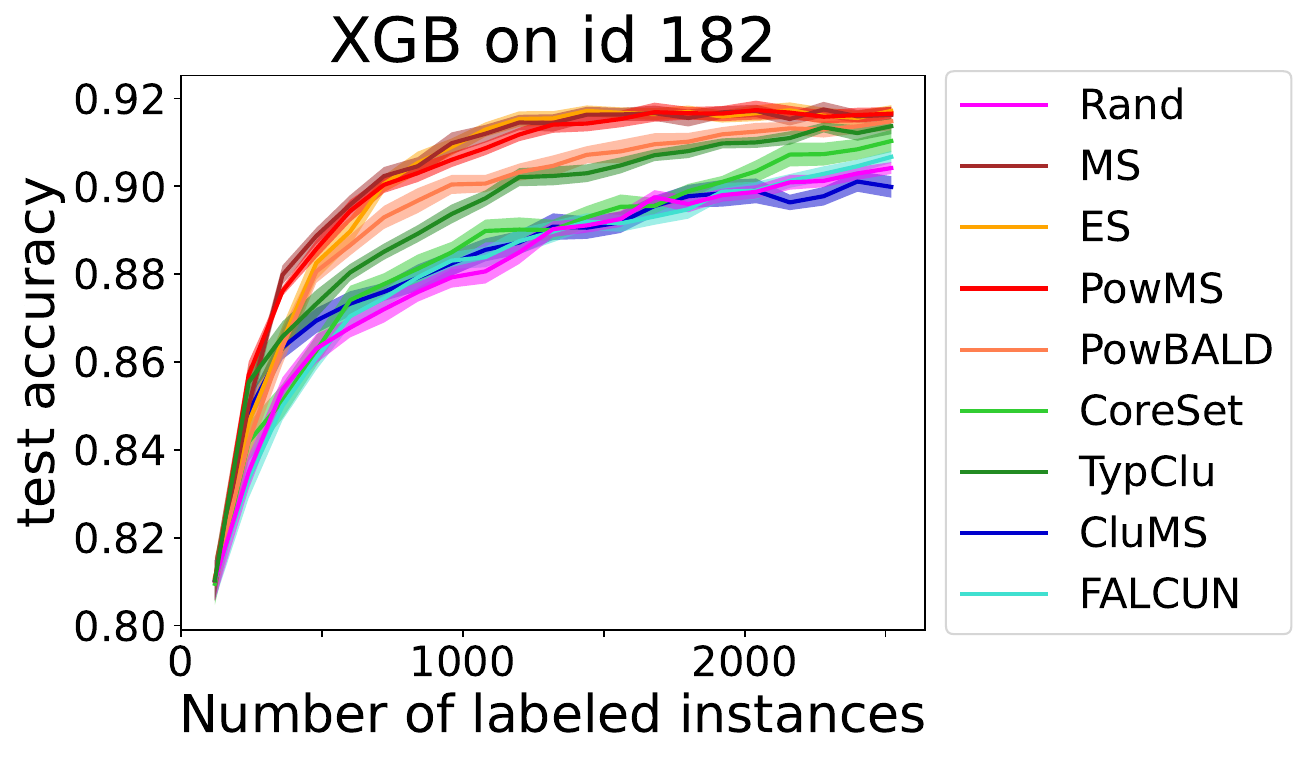}
    \end{subfigure}
        \begin{subfigure}[b]{0.49\textwidth}
        \centering
        \includegraphics[width=\textwidth]{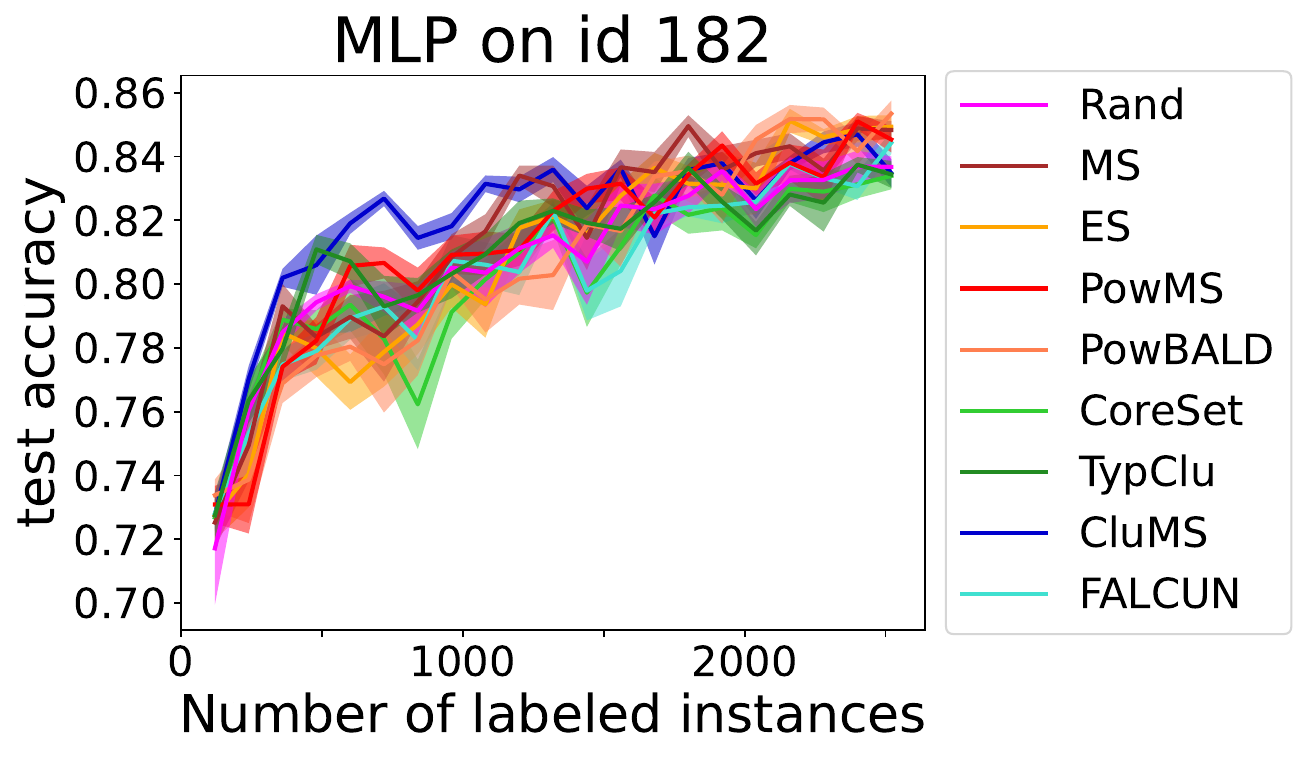}
    \end{subfigure}
        \begin{subfigure}[b]{0.49\textwidth}
        \centering
        \includegraphics[width=\textwidth]{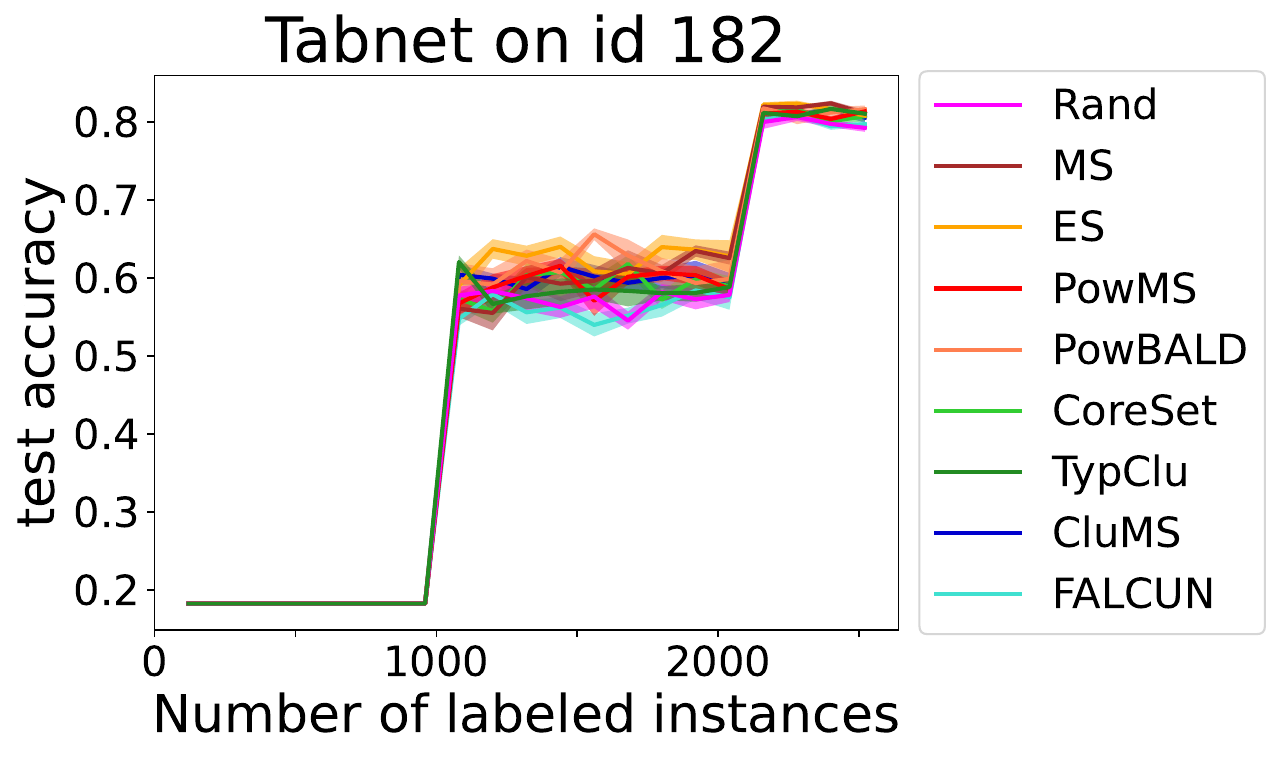}
    \end{subfigure}
    \begin{subfigure}[b]{0.49\textwidth}
         \centering
         \includegraphics[width=\textwidth]{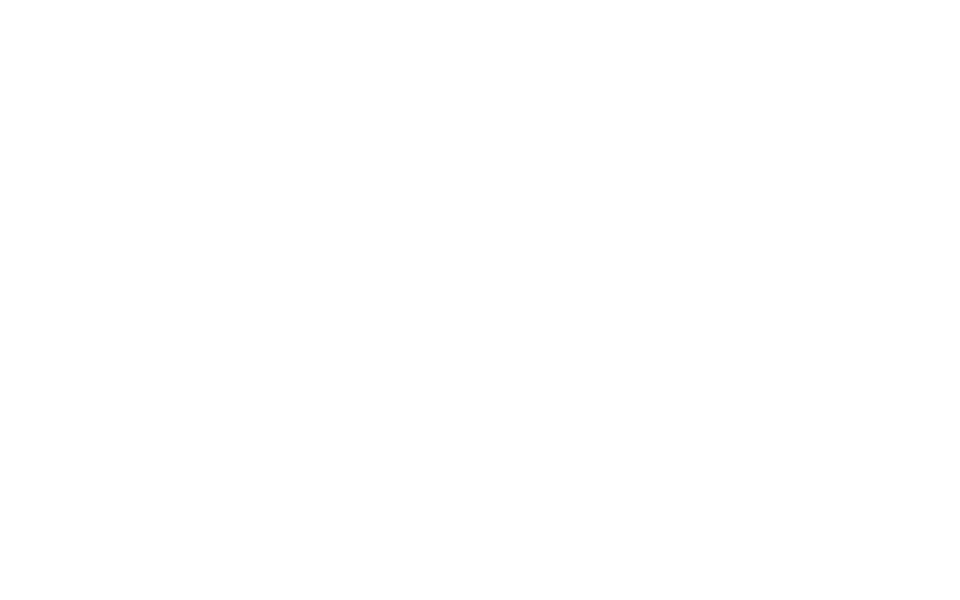}
     \end{subfigure}
    \caption{Budget curves for different \acp{alp} on the dataset with OpenML ID 182, considering the \textbf{large} setting. TabPFN is excluded since this dataset has more than 10 features.}
    \label{fig:182}
\end{figure}

\clearpage

\begin{figure}[t]
    \centering
    \begin{subfigure}[b]{0.49\textwidth}
        \centering
        \includegraphics[width=\textwidth]{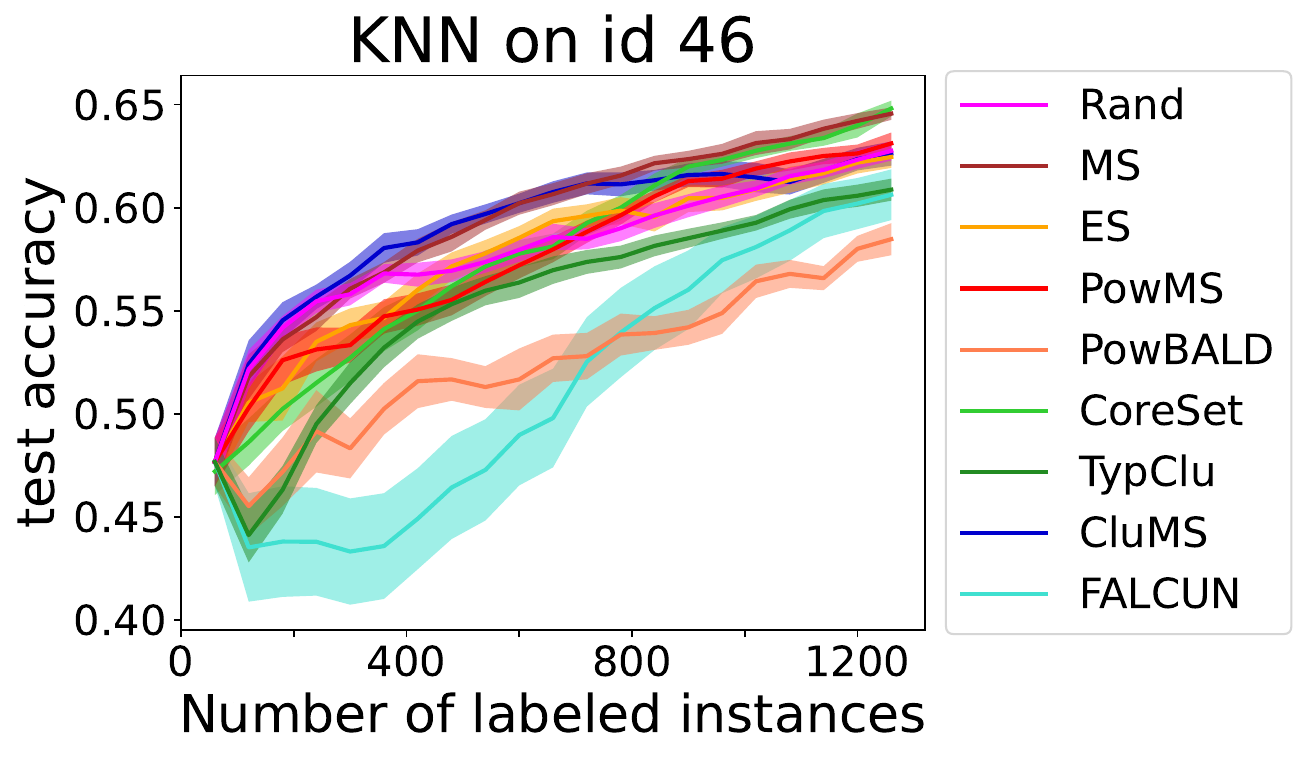}
    \end{subfigure}
    \begin{subfigure}[b]{0.49\textwidth}
        \centering
        \includegraphics[width=\textwidth]{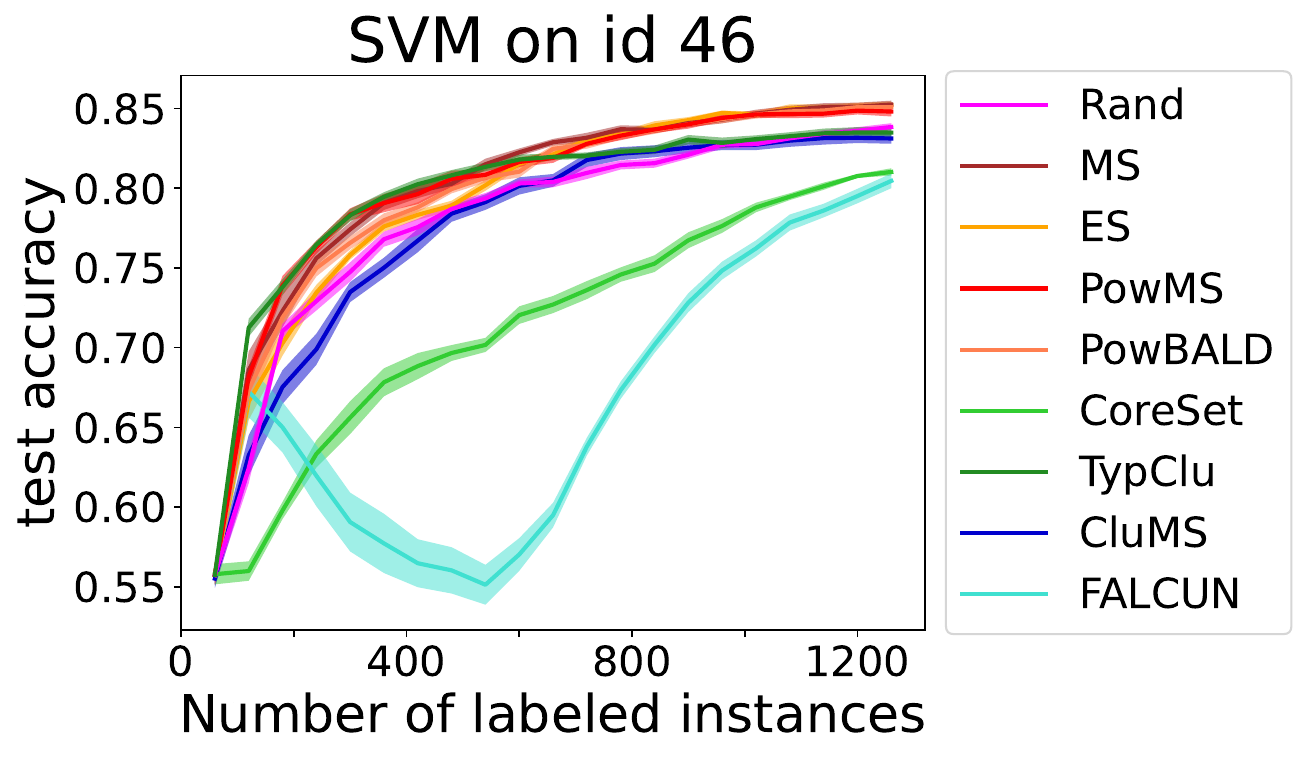}
    \end{subfigure}
        \begin{subfigure}[b]{0.49\textwidth}
        \centering
        \includegraphics[width=\textwidth]{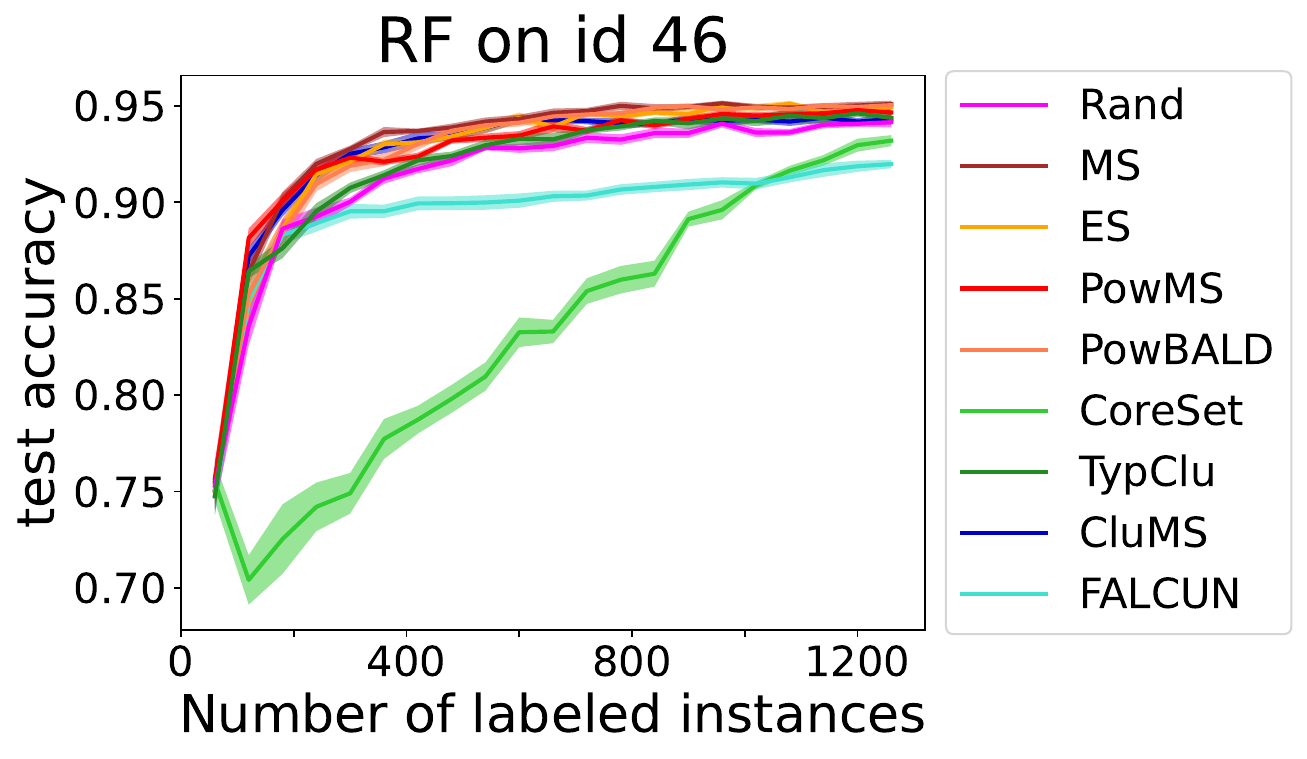}
    \end{subfigure}
        \begin{subfigure}[b]{0.49\textwidth}
        \centering
        \includegraphics[width=\textwidth]{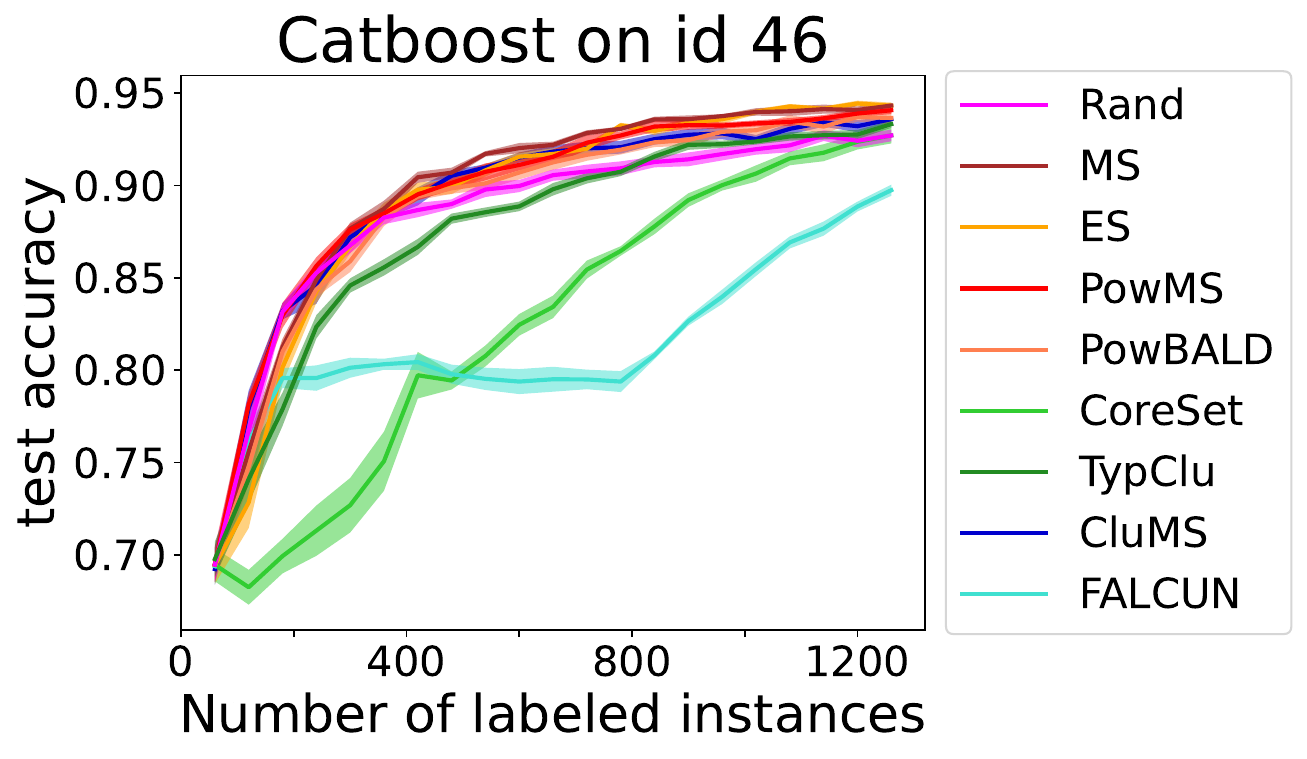}
    \end{subfigure}
        \begin{subfigure}[b]{0.49\textwidth}
        \centering
        \includegraphics[width=\textwidth]{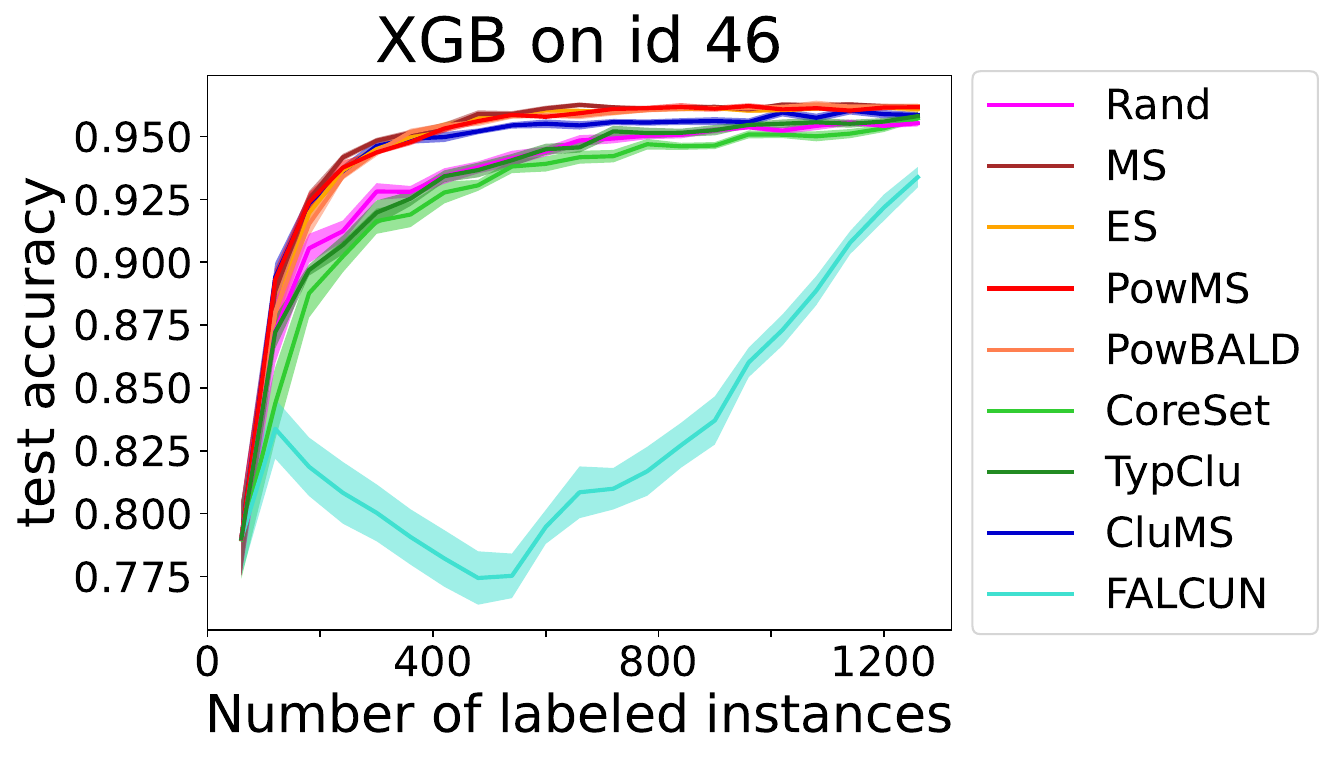}
    \end{subfigure}
        \begin{subfigure}[b]{0.49\textwidth}
        \centering
        \includegraphics[width=\textwidth]{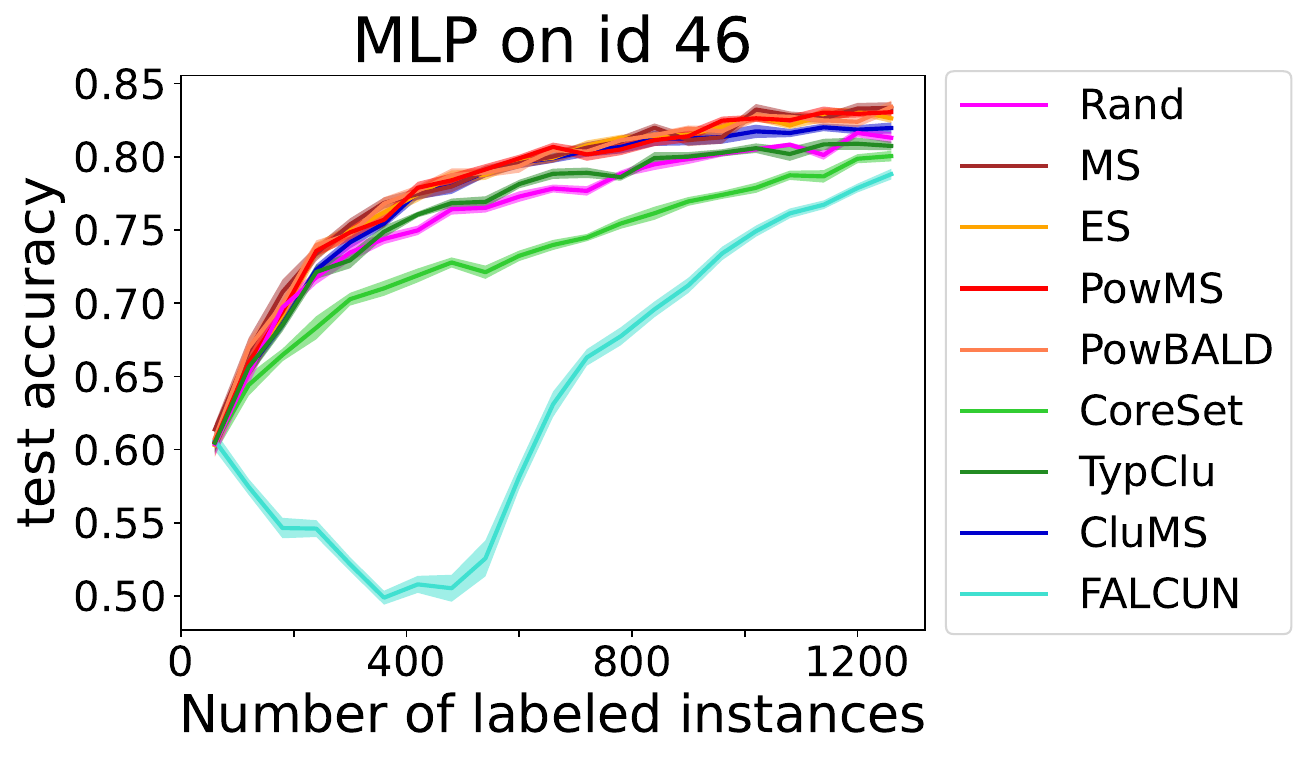}
    \end{subfigure}
        \begin{subfigure}[b]{0.49\textwidth}
        \centering
        \includegraphics[width=\textwidth]{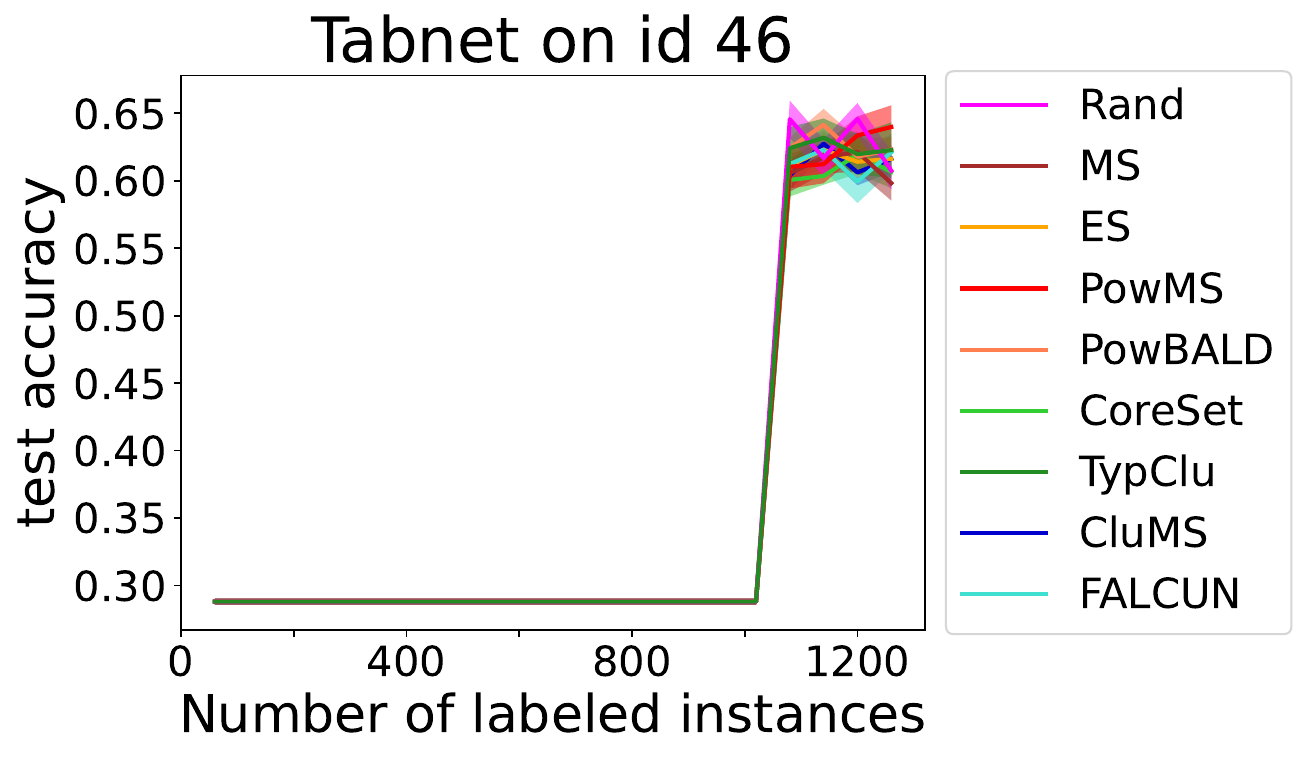}
    \end{subfigure}
    \begin{subfigure}[b]{0.49\textwidth}
         \centering
         \includegraphics[width=\textwidth]{figs/results/place.png}
     \end{subfigure}
    \caption{Budget curves for different \acp{alp} on the dataset with OpenML ID 46, considering the \textbf{large} setting. TabPFN is excluded since this dataset has more than 10 features.}
    \label{fig:46}
\end{figure}

\clearpage

\end{document}